\documentclass{article} 
\usepackage{iclr2026_conference,times}

\usepackage{amsmath,amsfonts,bm}

\def\eqref#1{equation~\ref{#1}}

\def\1{\bm{1}}

\DeclareMathAlphabet{\mathsfit}{\encodingdefault}{\sfdefault}{m}{sl}
\SetMathAlphabet{\mathsfit}{bold}{\encodingdefault}{\sfdefault}{bx}{n}

\usepackage{hyperref}
\usepackage{url}
\usepackage{graphicx}  
\usepackage{amsthm}
\usepackage{cleveref}
\crefname{equation}{Eq.}{Eqs.}
\usepackage{float}
\PassOptionsToPackage{hyphens}{url}
\usepackage{algorithm}
\usepackage{algpseudocode}
\usepackage{xurl}      

\usepackage{tikz}
\usetikzlibrary{positioning,arrows.meta,fit,calc}

\title{Analysis of Fourier Neural Operators via Effective Field Theory}
\iclrfinalcopy

\author{Taeyoung Kim  \\
School of Computational Sciences\\
Korea Institute for Advanced Study\\
Seoul 02455, South Korea \\
\texttt{taeyoungkim@kias.re.kr} \\
}

\begin{document}

\maketitle

\begin{abstract}
Fourier Neural Operators (FNOs) have emerged as leading surrogates for solver operators for various functional problems, yet their stability, generalization and frequency behavior lack a principled explanation. We present a systematic effective field theory analysis of FNOs in an infinite-dimensional function space, deriving closed recursion relations for the layer kernel and four-point vertex and then examining three practically important settings—analytic activations, scale-invariant cases and architectures with residual connections. The theory shows that nonlinear activations inevitably couple frequency inputs to high frequency modes that are otherwise discarded by spectral truncation, and experiments confirm this frequency transfer. For wide networks, we derive explicit criticality conditions on the weight initialization ensemble that ensure small input perturbations maintain a uniform scale across depth, and we confirm experimentally that the theoretically predicted ratio of kernel perturbations matches the measurements. Taken together, our results quantify how nonlinearity enables neural operators to capture non-trivial features, supply criteria for hyperparameter selection via criticality analysis, and explain why scale-invariant activations and residual connections enhance feature learning in FNOs. Finally, we translate the criticality theory into a practical criterion-matched initialization (calibration) procedure; on a standard
PDEBench Burgers benchmark, the calibrated FNO exhibits markedly more stable optimization, faster convergence, and improved test error relative to a vanilla FNO.
\end{abstract}

\vspace{0.5em}

\section{Introduction}

\subsection{Fourier Neural Operator}
In scientific machine learning, neural operators—networks that approximate solution operators so they can act directly on functional data—are attracting growing attention (\cite{Kovachki:212}, \cite{Lu:21}). In addition to the linear layers of standard fully connected networks (FCNs), neural operator architectures include layers that approximate a kernel and apply an integral transform. How the kernel is represented gives rise to many variants: graph-based constructions (\cite{Li:20}), tensor product decompositions (\cite{Kovachki:212}), hierarchical (multi-resolution) kernels (\cite{Gupta:21}), and, most notably, Fourier Neural Operators (FNOs) (\cite{Li:21}). Thanks to their computational efficiency, high accuracy, and ability to capture long-range interactions that elude CNNs and GNNs, FNOs have become the workhorse choice for surrogate modeling (\cite{Pathak:22}, \cite{Sun:23}). Prior theory has addressed their generalization error (\cite{Kim:24}, \cite{Benitze:24}), the universal approximation property (\cite{Kovachki:21}, \cite{Lee:25}), and expressivity/trainability from a mean-field perspective (\cite{Koshizuka:24}), yet a comprehensive statistical description is still lacking. Here we fill that gap by analyzing FNOs through the lens of effective field theory (EFT).

\subsection{Effective Field Theory for Neural Networks}
Because stochastic elements such as random initialization and stochastic gradient descent are intrinsic to neural networks, statistical-physics tools are natural candidates for theoretical analysis. Recent work has imported methods from field theory (\cite{Halverson21}, \cite{Banta24}), relating connected correlators in randomly initialized ensembles and deriving susceptibility based criticality conditions that predict when training remains stable (\cite{Roberts22}). Within the EFT framework one can see how the choice of activation or hyperparameters moves a model between different universality classes, revealing when signals explode, vanish, or propagate cleanly.

\subsection{Contributions}
Building on the EFT formulation for FCNs, we extend the approach to Fourier Neural Operators—an essential step, because FNOs act on infinite dimensional function spaces rather than finite-dimensional vectors. While the basic relations among connected correlators mirror those in FCNs, the architectural differences introduce qualitatively new phenomena: the observables become functions rather than scalars, and mechanisms such as frequency coupling appear. While frequency-coupling phenomena can be inferred qualitatively, our EFT framework explains the quantitative decay of coupling strength and its dependence on network width. We show that the kernel function—the infinite width approximation of the two-point correlator—admits a closed layerwise recursion, and we derive analytic formulas for kernel perturbations; we then validate these predictions experimentally. Our analysis quantifies how non-trivial activations and residual connections influence information flow in FNOs and yields explicit design criteria for stable models. Finally, leveraging these theoretical insights, we propose a practical criterion-matched initialization (calibration) algorithm; on a real benchmark dataset, it leads to more stable training dynamics, faster convergence, lower achieved loss, and improved generalization compared to the vanilla FNO baseline.

\begin{figure*}[htp]
\centering
\begin{tikzpicture}[
  font=\small,
  box/.style={draw, rounded corners=2pt, align=left, inner sep=6pt, line width=0.5pt, text width=0.92\textwidth},
  arrow/.style={-Latex, thick},
  node distance=6mm
]

\node[box] (a) {%
\textbf{Fourier Neural Operator (FNO) with spectral truncation.}\;
Given an input function $u(x)$, each Fourier layer applies
$\mathrm{FFT}\rightarrow$ truncation ($|f|\le k_{\max}$) $\rightarrow$ multiplication by $R^{(l)}(f)$ $\rightarrow \mathrm{iFFT}$,
followed by a pointwise nonlinearity $\sigma$; optionally, a residual update
$Z^{(l+1)}=\mathcal{R}^{(l+1)}(\sigma(\cdot))+\tilde{\gamma}Z^{(l)}$ is used to improve depth-wise information retention.
};

\node[box, below=of a] (b) {%
\textbf{Mechanism of frequency transfer.}\;
A pointwise nonlinearity becomes a convolutional expansion in Fourier space,
\[
 \mathcal{F}(\sigma(g))(f)=\sum_{n=1}^{\infty}\frac{\sigma_{n}}{n!}\hat{g}^{*n}(f).
\]
so nonlinear layers inevitably generate modes beyond the truncation band.
For analytic activations, the contribution of the $k$-th term is suppressed by the width as $\mathcal{O}(n^{-(k-1)})$.
};

\node[box, below=of b] (c) {%
\textbf{EFT formulation in function space.}\;
We treat random initialization as an ensemble over parameters (with fixed input functions $u,v$) and track connected correlators of the Fourier-domain pre-activations.
The principal observables are the two-point kernel $\mathcal{K}^{(l)}$ and the four-point vertex $\mathcal{V}^{(l)}$.
In the infinite-width limit, pre-activations converge to a Gaussian process and $\mathcal{V}^{(l)}=\mathcal{O}(1/n)$,
yielding closed layerwise recursions for $\mathcal{K}^{(l+1)}$ and for the parallel/perpendicular susceptibilities
$\chi_{\parallel}$ and $\chi_{\perp}$, which determine criticality conditions for stable signal propagation.
};

\node[box, below=of c] (d) {%
\textbf{Three representative regimes.}\;
(i) \emph{Analytic activations}: explicit convolution-series recursions for $\mathcal{K}^{(l)}$ and susceptibilities.
(ii) \emph{Scale-invariant (ReLU-type) activations}: kernels are governed by the position-space correlation $\rho^{(l)}$, leading to broad spectral support and slow decay.
(iii) \emph{Residual FNOs}: residual gain $\gamma$ controls the retention of post-truncation energy and enables deeper compositions while preserving truncation efficiency.
};

\node[box, below=of d] (e) {%
\textbf{Empirical validation.}\;
Across activation families (tanh/polynomial/ReLU) and with/without residual connections,
the kernel recursion accurately predicts the measured kernel evolution, including energy in post-truncation frequencies.
Measured susceptibilities $\chi_{\parallel}$ and $\chi_{\perp}$ agree with theory within one-standard-deviation bands over repeated random initializations.
Importantly, applying the proposed calibration at initialization translates to improved downstream training performance,
yielding more stable optimization and consistently better final metrics (e.g., faster convergence and/or lower test error)
than uncalibrated baselines.
};

\draw[arrow] (a.south) -- (b.north);
\draw[arrow] (b.south) -- (c.north);
\draw[arrow] (c.south) -- (d.north);
\draw[arrow] (d.south) -- (e.north);

\end{tikzpicture}
\caption{Schematic overview of the paper: (top) the truncated FNO architecture; (middle) nonlinearity-induced frequency transfer and the EFT description via the kernel/vertex hierarchy; (bottom) closed-form results in three regimes and their experimental validation and the effectiveness of calibration based on our theory.}
\label{fig:schematic_overview}
\end{figure*}

\section{Preliminaries}
In this section, we examine the definition of neural operators designed for processing functional data, and explore effective field theory along with its application to the statistical analysis of basic neural networks (fully connected networks). Henceforth, the Einstein summation convention will be adopted for repeated indices.

\subsection{Effective Field Theory}
Suppose that the distribution $p(X_{1},\dots,X_{n})\sim \exp(-S(X_{1},\dots,X_{n}))$ is given. For an analytic function $f(x_{1},\dots,x_{n})=\sum a_{i_{1}\dots i_{n}}x_{1}^{i_{1}}\dots x_{n}^{i_{n}}$ the expectation of $f$ with respect to $p$ is expressed as
\[
\mathbb{E}[f(X_{1},\dots,X_{n})]=\sum a_{i_{1}\dots i_{n}}\mathbb{E}[X_{1}^{i_{1}}\dots X_{n}^{i_{n}}]
\]
Thus, in the analytic observable case, the collection of terms $\mathbb{E}[X_{1}^{i_{1}}\dots X_{n}^{i_{n}}]$ contains all the information about the distribution. These terms are referred to as the $(i_{1}+\dots+i_{n})$-point correlators (or $(i_{1}+\dots+i_{n})$-th moments). For a Gaussian distribution, where $S(X_{1},\dots,X_{n})$ consists solely of quadratic terms, the following well-known result holds:

\noindent
{\bf Proposition 1 (Wick contraction)}{ Suppose $(X_{1},\dots,X_{n})$ is a zero-mean multivariate normal random vector. Then, all odd order correlators vanish, and the even order correlators are given by
\[
\mathbb{E}[X_{j_{1}}^{i_{1}}\dots X_{j_{n}}^{i_{n}}]=\sum_{\text{(all possible pairings)}}\prod_{\text{(pairings)}}\mathbb{E}[X_{p_{1}}X_{p_{2}}]
\]
}
Motivated by this proposition, we define the connected correlator (or cumulant) for an arbitrary distribution as follows:

\noindent
{\bf Definition 1}{ The $k$-point connected correlator (or $k$-th cumulant, $k=i_{1}+\dots+i_{n}$) is defined by
\begin{equation*}
    \begin{split}
        &\mathbb{E}|_{\text{conn}}[X_{j_{1}}^{i_{1}}\dots X_{j_{n}}^{i_{n}}]:=\mathbb{E}[X_{j_{1}}^{i_{1}}\dots X_{j_{n}}^{i_{n}}] \\-&\sum_{\text{all subdivisions of }(1,\dots,k)}\mathbb{E}[X_{j_{\mu_{1},1}}\dots X_{j_{\mu_{1},t_{1}}}]|_{\text{conn}}\dots\mathbb{E}[X_{j_{\mu_{\nu},1}}\dots X_{j_{\mu_{\nu},t_{\nu}}}]|_{\text{conn}}
    \end{split}
\end{equation*}

}
Since the connected correlators vanish for Gaussian distributions, they serve as an indicator of the deviation of a distribution from Gaussianity. A review of the effective theory analysis for fully connected networks (FCNs) is provided in Appendix~\ref{app:fcn}.

\subsection{Neural Operator}
The architecture of the neural operator we consider consists of the following two types of layers: \cref{linearlayer} describes a linear layer that linearly transforms the channel dimension of functional data, and \cref{kernellayer} represents a kernel integration layer performing convolution operations with a kernel function. \cref{activationlayer} expresses the kernel integration layer combined with a non-linear activation function. In general, the kernel function does not need to be translationally invariant; however, in our setting, we focus on the translationally invariant case as explained in assumption (A1).
\begin{equation}
    \Big(\mathcal{L}(u)(x)\Big)_{i}:=\sum_{j}W_{ij}u_{j}(x), \quad i=1,\dots,n. \label{linearlayer}
\end{equation}
\begin{equation}
    \Big(\mathcal{R}(u)(x)\Big)_{i}:=\int_{\mathbb{R}^{d}}k_{ij}(x,x')u_{j}(x')dx',\quad i=1,\dots,n.\label{kernellayer}
\end{equation}
\begin{equation}
    \mathfrak{S}(x):=\sigma(\mathcal{R}(u)(x)). \label{activationlayer}
\end{equation}

\paragraph{Modeling assumptions and scope.}
For clarity, we list the assumptions under which the main EFT recursions and closed-form results in Sections~3--4 are derived.

\noindent\textbf{(A1) Translation-invariant kernel layers.}
Each kernel integration layer is a translation-invariant convolution operator $k(x-x')$, so that in Fourier space the layer acts by mode-wise multiplication.

\noindent\textbf{(A2) Channel-i.i.d. Gaussian initialization.}
For each Fourier mode, the complex-valued parameters $\{R^{(l)}_{ij}(f)\}_{i,j}$ are initialized as mean-zero Gaussian variables, independent across channel indices $(i,j)$ and across layers, with variance scaling $1/n^{(l-1)}$.

\noindent\textbf{(A3) Frequency-diagonal spectral covariance (baseline setting).}
In the main development we assume frequency-wise independence,
$\mathbb{E}[R^{(l)}_{ij}(f)\overline{R^{(l)}_{i'j'}(f')}]\propto \delta(f-f')$,
which yields the diagonal spectral profile $C_{R^{(l)}}(f)\delta(f-f')$ in \cref{initen}.
Truncation is encoded by choosing $C_{R^{(l)}}(f)=0$ for $|f|>k_{\max}$.

\medskip
\noindent\textbf{Covered architectures.}
Assumptions (A1)--(A3) include the standard FNO layer (FFT $\rightarrow$ truncation $\rightarrow$ multiplication by a learnable tensor $\rightarrow$ iFFT $\rightarrow \sigma$), arbitrary fixed spectral profiles $C_{R}(f)$ (step, decaying, etc.), and ResNet-style variants as in \cref{resfno}.
Linear lift/projection layers initialized as in the FCN baseline are also covered.
Moreover, the CNN subnetwork commonly used in standard FNO implementations can be interpreted in our framework as a Fourier layer whose effective kernel is spatially localized (equivalently, whose spectral response is broadly supported rather than concentrated on low modes); hence, its contribution can be analyzed within the same operator-kernel formalism.

\noindent\textbf{Outside the present scope.}
Architectures that break translation invariance at the kernel level (e.g., explicitly $x$-dependent kernels, windowed/local operators that are not diagonalizable by a global Fourier basis), structured parameterizations that couple channels and frequencies beyond the Gaussian ensemble above, or non-Gaussian initializations may require modified correlator calculations.
We discuss a direct generalization to correlated Gaussian spectral weights below.

\noindent
{\bf Definition 2 }{ By composing the linear layers and kernel integration layers introduced above, we build the Neural Operator architecture as follows:
\begin{equation}
    \begin{split}
        \mathcal{Z}^{(1)}&:=\mathcal{L}_{\text{lift}}(u), \\
        \mathcal{Z}^{(l+1)}(u)&:=\mathcal{R}^{(l+1)}\Big(\mathfrak{S}^{(l)}(u)\Big), \quad l=1,\dots,L-1,\\
        \mathcal{Y}(u)&:=\mathcal{L}_{\text{proj}}\Big(\mathcal{Z}^{(L)}(u)\Big).
    \end{split} \label{defno}
\end{equation}
Here, $\mathcal{L}_{\text{lift}}$ and $\mathcal{L}_{\text{proj}}$ are linear layers. $\mathcal{L}_{\text{lift}}$ lifts the input function’s pointwise values into a higher dimensional feature space,  $\mathcal{L}_{\text{proj}}$ projects the pre-activations—after they have passed through the sequence of kernel integration layers—back down to the dimensionality of the target function. $\mathcal{R}^{(k)}$ denotes a kernel integration layer. In the Fourier Neural Operator of \cite{Li:21}, each kernel integration layer is implemented by taking the Fourier transform of its input, multiplying by a learnable tensor, and then applying the inverse Fourier transform. Concretely, one can write: 
\[
(u_{i}^{(k)})\mapsto \sigma\Big(\mathcal{F}^{-1}\Big(\sum_{j}R_{ij}^{(k+1)}(f)\hat{u}^{(k)}_{j}(f)\Big)\Big)
\]
where $R_{ij}^{(k+1)}(f)$ represents the parameterized complex-valued tensor. We adopt the convention of using $\mathcal{F}$ and the hat notation for the Fourier transform, and $\mathcal{F}^{-1}$ for the inverse transform.  As with fully‐connected networks, we initialize all parameters so that they follow the statistical distributions:
\begin{equation}
    \begin{split}
        \mathbb{E}[R_{ij}^{(k+1)}(f)]&=0, \\
        \mathbb{E}[\text{Re}(R_{i_{1}j_{1}}^{(k+1)}(f))\text{Re}(R_{i_{2}j_{2}}^{(k+1)}(f'))]&=\frac{C_{R^{(k+1)}}(f)}{2n^{(k)}}\delta_{i_{1}i_{2}}\delta_{j_{1}j_{2}}\delta(f-f'), \\
        \mathbb{E}[\text{Im}(R_{i_{1}j_{1}}^{(k+1)}(f))\text{Im}(R_{i_{2}j_{2}}^{(k+1)}(f'))]&=\frac{C_{R^{(k+1)}}(f)}{2n^{(k)}}\delta_{i_{1}i_{2}}\delta_{j_{1}j_{2}}\delta(f-f'), \\
        \mathbb{E}[\text{Re}(R_{i_{1}j_{1}}^{(k+1)}(f))\text{Im}(R_{i_{2}j_{2}}^{(k+1)}(f'))]&=0.
    \end{split} \label{initen}
\end{equation}
Therefore, the initialization distribution of the kernel integration layers can be viewed as white noise, and when implemented in practice via discretization, it matches the setup in \cite{Li:21}. Our neural-operator architecture assumes real-valued functions. To guarantee that multiplying by the parameter tensor $R_{ij}^{(k)}$ and then applying the inverse Fourier transform still yields a real function, we theoretically impose the symmetry on the sampled parameters: $R_{ij}^{(k)}(f)=\overline{R_{ij}^{(k)}(-f)}$. During training the values of $R_{ij}^{(k)}$ evolve, yet the symmetry is preserved in practice because the implementation processes data with a real discrete Fourier transform, which enforces the required conjugate symmetry automatically. For the linear layers, we define their initialization distributions in exactly the same way as FCN in Appendix~\ref{app:fcn}.

\paragraph{Remark (Lemma~2 and the frequency-diagonal assumption).}
Lemma~2 relies critically on the frequency-diagonal covariance in Assumption~(A3),
\(
\mathbb{E}[R(f)\overline{R(f')}]=C(f)\delta(f-f').
\)
This $\delta$-structure forces the Wick contractions to collapse onto a single frequency variable,
yielding the closed one-dimensional convolution form \(\delta(f-f')\,H^{*n}(f)\).
If we instead allow correlated Gaussian spectral weights with a non-diagonal covariance
\(
\mathbb{E}[R(f)\overline{R(f')}]=\Sigma_R(f,f'),
\)
then the contractions no longer produce \(\delta(f-f')\), and the conclusion of Lemma~2
does not hold in this simple form: the resulting moments remain genuinely two-frequency objects
(with sums over pairings/permutations), and the recursion formulas in Theorem~1 must be written
in terms of the full kernel \(\mathcal{K}^{(l)}(f,f')\) rather than its diagonal.

 Most practical FNO implementations parameterize Fourier modes independently (up to the conjugate-symmetry constraint for real outputs)
and apply spectral truncation mode-wise, which corresponds precisely to Assumption~(A3) with
a diagonal profile \(C_R(f)\delta(f-f')\).
Our experiments follow this standard implementation choice; it preserves the FFT-based efficiency
and yields a frequency-diagonal kernel that admits the explicit convolution recursions used throughout the paper.
Allowing general \(\Sigma_R(f,f')\) is a meaningful extension, but it typically requires tracking dense
frequency--frequency correlations and leads to substantially more involved analytic expressions.

\section{Effective Field Theory for Neural Operators}

Following the effective field theory (EFT) framework for neural networks set out in Appendix~\ref{app:fcn}, the functional degrees of freedom $u$ are assumed to follow the Boltzmann type distribution \[
P(u) \propto \exp({-S(u)}).
\]

where the action $S(u)$ fully determines the statistical weight of each configuration. The expectation value of $u$ at position $x$ is therefore obtained from the path integral
\[
\mathbb{E}[u(x)]=\frac{\int u(x) P(u)\mathcal{D}u }{\int P(u)\mathcal{D}u}.
\]
Because both differentiation and ordinary integration act linearly on the functional measure, connected correlators furnish a systematic means of computing expectation values that involve products, derivatives, or integrals of multiple fields. In what follows, closed form expressions for these correlators are derived, and—paralleling the analysis in Appendix~\ref{app:fcn}—recursive relations are established for the two-point kernel and the four-point vertex. Finally, explicit formulas are provided for the susceptibility of the kernel to perturbations parallel and perpendicular to a chosen reference trajectory in function space, thereby quantifying anisotropic responses to input variations. In the analysis, $\mathbf{u}$ and $\mathbf{v}$ are fixed input functions; no assumption on an input distribution is needed. Expectations are taken over the random initialization of the network parameters, while in experiments we use a Gaussian random field for sampling input functions only as a convenient testbed.

\subsection{Correlators for Neural Operators}

Henceforth, boldface letters indicate vector-valued functions. The notation $\hat{u}$ also means $\mathcal{F}(u)$. Because the explicit evaluation of correlators in layers that include convolution is prohibitively cumbersome, we instead work in Fourier space and compute the correlators of the Fourier transformed pre-activations. In particular, \cref{conditwo} gives the two-point correlator of the 
$l$-th layer conditioned on the $(l-1)$-th layer’s outputs $\mathbf{u}$ and $\mathbf{v}$. We write $\Big(\mathcal{Z}^{(l)}|_{\mathbf{u}}\Big)_{i}$ for the $i$-th component of $\mathcal{Z}^{(l)}|_{\mathbf{u}}$, that is, the $i$-th pre-activation in layer $l$ when the preceding layer is fixed to the function $\mathbf{u}$. Likewise, we write $\Big(\mathcal{Z}^{(l)}\{\mathbf{u}\}\Big)_{i}$ for the corresponding output component of layer $l$ when the input function is $\mathbf{u}$.

\begin{equation}
\begin{split}
    &\mathbb{E}\bigg[\mathcal{F}\Big(\mathcal{Z}^{(l)}|_{\mathbf{u}}\Big)_{i}(f)\overline{\mathcal{F}\Big(\mathcal{Z}^{(l)}|_{\mathbf{v}}\Big)}_{i'}(f')\bigg|u,v\bigg] \\
    &=\mathbb{E}\bigg[\sum_{jj'}R^{(l)}_{ij}(f)\overline{R^{(l)}_{i'j'}(f')}\hat{u}_{j}(f)\overline{\hat{v}_{j'}}(f')\bigg|u,v\bigg] \\
   &=\sum_{jj'}\frac{C_{R^{(l)}}}{n^{(l-1)}}\delta(f-f')\delta_{ii'}\delta_{jj'}\hat{u}_{j}(f)\overline{\hat{v}_{j'}}(f') \\
   &=\frac{C_{R^{(l)}}}{n^{(l-1)}}\delta(f-f')\delta_{ii'}\sum_{j}\hat{u}_{j}(f)\overline{\hat{v}_{j}}(f').
\end{split} \label{conditwo}
\end{equation}

Although we currently compute all statistical objects for the Fourier transforms of the pre-activation functions, the corresponding quantities in the spatial domain are recovered via \cref{domain}. Specifically,
\begin{equation}
    \begin{split}
        &\mathbb{E}\bigg[\Big(\mathcal{Z}^{(l)}|_{\mathbf{u}}(x)\Big)_{i}\overline{\Big(\mathcal{Z}^{(l)}|_{\mathbf{v}}(y)\Big)}_{i'}\bigg|\mathbf{u},\mathbf{v}\bigg] \\        &=\int_{\mathbb{R}^{d}}\int_{\mathbb{R}^{d}}\mathbb{E}\bigg[\mathcal{F}\Big(\mathcal{Z}^{(l)}|_{\mathbf{u}}\Big)_{i}(f)\overline{\mathcal{F}\Big(\mathcal{Z}^{(l)}|_{\mathbf{v}}\Big)}_{i'}(f')\bigg|\mathbf{u},\mathbf{v}\bigg]e^{ifx}e^{-if'y}dfdf' \\
        &=\frac{C_{R^{(l)}}}{n^{(l-1)}}\delta_{ii'}\mathcal{F}^{-1}\Big(\langle \mathbf{\hat{u}},\mathbf{\hat{v}}\rangle\Big).
    \end{split} \label{domain}
\end{equation}

Now we consider the two-point correlator for the neural operator. Since the parameters in $l$-th layer are independent of the variables up to $(l-1)$-th layers, the two-point correlator factorizes as in \cref{2point}. 
\begin{equation}
    \begin{split}
        &\mathbb{E}\bigg[\mathcal{F}\Big(\mathcal{Z}^{(l)}\{\mathbf{u}\}\Big)_{i}(f)\overline{\mathcal{F}\Big(\mathcal{Z}^{(l)}\{\mathbf{v}\}\Big)_{i'}}(f')\bigg] \\
        &=\frac{C_{R^{(l)}}}{n^{(l-1)}}\delta_{ii'}\delta(f-f')\mathbb{E}\bigg[\Big(\mathcal{F}\Big(\mathfrak{S}^{(l-1)}\{\mathbf{u}\}\Big)(f)\cdot\overline{\mathcal{F}\Big(\mathfrak{S}^{(l-1)}\{\mathbf{v}\}\Big)}(f')\bigg].  \label{2point}
    \end{split}
\end{equation}

We define the $l$-th layer mean metric $\mathcal{G}^{(l)}$ as the non-trivial part of \cref{2point}, i.e., \cref{meanmetric}, and define stochastic metric $\widetilde{\mathcal{G}^{(l)}}$ as in \cref{stometric}, where the expectation is taken only over the parameters of $l$-th layer:
\begin{equation}
    \mathcal{G}^{(l)}\{\mathbf{u},\mathbf{v}\}(f,f'):=\delta(f-f')\frac{C_{R^{(l)}}}{n^{(l-1)}}\mathbb{E}\bigg[\mathcal{F}\Big(\mathfrak{S}^{(l-1)}\{\mathbf{u}\}\Big)(f)\cdot\overline{\mathcal{F}\Big(\mathfrak{S}^{(l-1)}\{\mathbf{v}\}\Big)}(f')\bigg]. \label{meanmetric}
\end{equation}
\begin{equation}
    \widetilde{\mathcal{G}^{(l)}}\{\mathbf{u},\mathbf{v}\}(f,f'):=\delta(f-f')\frac{C_{R^{(l)}}}{n^{(l-1)}}\sum_{j}\mathcal{F}\Big(\mathfrak{S}^{(l-1)}\{\mathbf{u}\}\Big)_{j}(f)\overline{\mathcal{F}\Big(\mathfrak{S}^{(l-1)}\{\mathbf{v}\}\Big)_{j}}(f'). \label{stometric}
\end{equation}

The fluctuation of this metric is governed by the four-point connected correlator; see the Appendix~\ref{app:sec3} for a detailed derivation.

\subsection{Running of Couplings}

In this subsection we derive recursion relations for the correlation functions—what, in field theory language, is the running of couplings. As the layer width tends to infinity, the pre-activation ensemble becomes Gaussian and the layer metric converges to a deterministic kernel $\mathcal{K}^{(l)}$ (see Appendix~\ref{app:sec3}). In this vanishing fluctuation regime, the next layer kernel can be expressed as a Gaussian expectation with covariance $\mathcal{K}^{(l)}$ inherited from the previous layer; equivalently, $\mathcal{K}^{(l+1)}$ is obtained by integrating over the Gaussian process defined by $\mathcal{K}^{(l)}$:

\begin{equation*}
    \begin{split}
        \mathcal{K}^{(l+1)}\{\mathbf{u},\mathbf{v}\}(f,f'):=\delta(f-f')C_{R^{(l+1)}}(f)\langle \mathcal{F}(\mathfrak{S}^{(l)}\{\mathbf{u}\}),\mathcal{F}(\mathfrak{S}^{(l)}\{\mathbf{v}\})\rangle_{\mathcal{K}^{(l)}}(f,f').
    \end{split}
\end{equation*}

Here, notation $\langle A,B\rangle_{\mathcal{K}}(f,f')$ means the expectation value of $A(f)\overline{B(f')}$ where each $A$ and $B$ are Gaussian random fields with kernel $\mathcal{K}$. We will denote functions $\mathcal{K}\{\mathbf{u},\mathbf{u}\}$ simply as $\mathcal{K}\{\mathbf{u}\}$ and $\langle A,A\rangle_{\mathcal{K}}$ as $\|A\|_{\mathcal{K}}$. $\mathcal{K}\{\mathbf{u}\}(f)$ will mean the diagonal values $\mathcal{K}\{\mathbf{u}\}(f,f)$. Specifically, the first kernel is defined as in \cref{firstkernel} and deeper layer kernels are calculated recursively as in \cref{recurker}.
\begin{equation}
    \begin{split}
        \mathcal{K}^{(1)}\{\mathbf{u},\mathbf{v}\}(f,f'):=\frac{C_{W}}{n^{(0)}}\sum_{j}\hat{u_{j}}(f)\hat{v_{j}}(f').
    \end{split} \label{firstkernel}
\end{equation}
\begin{equation}
    \begin{split}
        \mathcal{K}^{(l+1)}\{\mathbf{u}\}(f,f'):=\delta(f-f')\frac{C_{R^{(l+1)}}}{n^{(l)}}g(\mathcal{K}^{(l)}), \\
        g(\mathcal{K}):= \langle\mathcal{F}(\mathfrak{S}^{(l)}\{\mathbf{u}\}),\mathcal{F}(\mathfrak{S}^{(l)}\{\mathbf{u}\})\rangle_{\mathcal{K}^{(l)}}(f,f').
    \end{split} \label{recurker}
\end{equation}

We analyze the layerwise amplification factor (susceptibility) of the kernel—i.e., the ratio of the kernel perturbation at layer $l\!+\!1$ to that at layer $l$—when an infinitesimal perturbation is injected into the previous layer. We consider both the parallel ($\chi_{\parallel}$) and the perpendicular ($\chi_{\perp}$) susceptibilities; formal definitions and derivations appear in Appendix~\ref{app:susc}.
Note that parallel and perpendicular susceptibilities are defined at different perturbative orders.
Because the perturbation is orthogonal to the input in the perpendicular case, the leading non-vanishing contribution appears at second order in the small perturbation $\delta\boldsymbol{\eta}$. By contrast, a perturbation parallel to the input produces a non-zero contribution already at first order along parameter $\epsilon$. Each susceptibility is therefore defined in terms of the fluctuation that arises at its respective lowest non-trivial order. Accordingly, for local fluctuations to remain stable under perturbations perpendicular to the reference trajectory, the following condition must be satisfied:
\begin{equation}
    \delta(f-f')C_{R^{(l+1)}}\Big(\Big\|\mathcal{F}\Big(\mathfrak{S}'^{(l)}\{u_{0}\}\Big)*\delta\boldsymbol{\eta}\Big\|_{\mathcal{K}^{(l)}}\Big)(f,f')=\mathcal{K}_{\eta}(f,f'). \label{localperp}
\end{equation}
Imposing that relation \cref{localperp} exactly would make $\|\mathcal{F}(\mathfrak{S}'^{(l)}\{\mathbf{u_0}\})\|_{\mathcal{K}^{(l)}\{u_{0}\}}$ behave like a Dirac delta, which is too restrictive for most practical networks. Instead, we adopt a softer requirement: the integrated (layer-wise) sum of the kernel remains constant across layers, so the overall spectral energy is preserved while still allowing a non-trivial, spatially extended kernel shape.
\begin{equation*}
\begin{split}
        &\int \delta(f-f')C_{R^{(l+1)}}\|\mathcal{F}(\mathfrak{S}'^{(l)}\{\mathbf{u_0}\})*\widehat{\delta\boldsymbol{\eta}}\|_{\mathcal{K}^{(l)}\{u_{0}\}}dfdf'=\int\mathcal{K}_{\eta}(f,f')dfdf' \\
        &\Rightarrow \int \delta(f-f')C_{R^{(l+1)}}\|\mathcal{F}(\mathfrak{S}'^{(l)}\{\mathbf{u_0}\})\|_{\mathcal{K}^{(l)}\{u_{0}\}}(f,f')dfdf'=1 \\
        &\Rightarrow \int C_{R^{(l+1)}}(f)\|\mathcal{F}(\mathfrak{S}'^{(l)}\{\mathbf{u_0}\})\|_{\mathcal{K}^{(l)}\{u_{0}\}}(f)df=1.
\end{split}
\end{equation*}
For convenience we introduce the reduced perpendicular susceptibility
\[
\tilde{\chi}_{\perp}(f,f'):=\|\mathcal{F}(\mathfrak{S}'^{(l)}\{\mathbf{u_0}\})\|_{\mathcal{K}^{(l)}\{u_{0}\}}(f,f')
\]
With this notation, the analysis of correlators and their recursion relations yields two depth-wise critical conditions:
\begin{equation*}
    \begin{split}
        \chi_{\parallel}&\equiv1, \quad \text{ local condition (for parallel perturbation)}.\\
        \int C(f)\tilde{\chi}_{\perp}&(f)df \equiv 1\quad \text{ global condition (for perpendicular perturbation)}.
    \end{split}
\end{equation*}
Here, $\tilde{\chi}_{\perp}$ is calculated at layer $l$ and $C(f)$ is $C_{R^{(l+1)}}(f)$.

\section{Results: Three Representative Cases}
In this section we put the general framework of Section 3 to work by carrying out the calculation in three representative settings—(i) analytic activations, (ii) scale-invariant architectures, and (iii) networks equipped with residual connections.

\subsection{Analytic Activations}

First, we consider the case where the activation is an analytic function so that we can expand it and kernels perturbatively. We will frequently use the notation $\mathcal{H}^{(l)}(f):=\mathcal{K}^{(l)}(f)C_{R^{(l)}}(f)$; in the discrete case, $\mathcal{H}^{(l)}(n):=\mathcal{K}^{(l)}(n)C_{R^{(l)}}(n)$. Suppose the activation passes through the origin then it can be written as \cref{analy}.
\begin{equation}
    \begin{split}
        \sigma(x):=\sum_{n=1}\frac{\sigma_{n}}{n!}x^{n} 
    \end{split} \label{analy}
\end{equation}

Following the analysis in Appendix~\ref{app:thm1}, we obtain the following expressions for the kernel recursion and the parallel/perpendicular susceptibilities:

\noindent
{\bf Theorem 1} { With the analytic, origin passing activation specified in \cref{analy} and a Fourier Neural Operator defined by \cref{defno} under the initialization ensemble \cref{initen}, the kernel and the susceptibilities are given by the following recursion relations:
\begingroup
\small
\begin{equation*}
\begin{split}
&\mathcal{K}^{(l+1)}(f,f')=\delta(f-f')C_{R^{(l+1)}}(f)\sum_{k=1}^{\infty}\frac{\sigma_{k}^{2}}{(n^{(l)})^{k-1}} \\
&\sum_{n_{k,1}+\dots+n_{k,l}=k}\frac{(n_{k,1})!\dots(n_{k,l})!}{k!}(\mathcal{H}^{(l)})^{*n_{k,1}}\cdots(\mathcal{H}^{(l)})^{*n_{k,l}}(f), \\
        &\chi_{\parallel}(f,f')=\delta(f-f')C_{R^{(l+1)}}(f)\sum_{k=1}^{\infty} \frac{\sigma_{k}^{2}}{(n^{(l)})^{k-1}} \\
        &\sum_{n_{k,1}+\dots+n_{k,l}=k}\frac{(n_{k,1})!\dots(n_{k,l})!}{(k-1)!}\frac{(\mathcal{H}^{(l)})^{*n_{k,1}}\cdots(\mathcal{H}^{(l)})^{*n_{k,l}}(f)}{\mathcal{K}^{(l)}\{\mathbf{u_0}\}(f)}, \\
    &\tilde{\chi}_{\perp}(f,f')=\delta(f-f')\sum_{k=1}^{\infty}\frac{\sigma_{k}^{2}}{(n^{(l)})^{k-1}} \\
    &\sum_{n_{k,1}+\dots+n_{k,l}=k-1}\frac{(n_{k,1})!\dots(n_{k,l})!}{(k-1)!}(\mathcal{H}^{(l)})^{*n_{k,1}}\cdots(\mathcal{H}^{(l)})^{*n_{k,l}}(f).
\end{split}
\end{equation*}
\endgroup
}

\subsection{scale-invariant Activations}
The scale-invariant class of activations is defined as follows:
\begin{equation}
    \sigma(z):=\begin{cases}
        \alpha z \quad \text{for }z\geq0 \\
        \beta z \quad \text{for }z<0.
    \end{cases} \label{scaleac}
\end{equation}

Following the analysis in Appendix~\ref{app:thm2}, we obtain the following expressions for the kernel recursion and the parallel/perpendicular susceptibilities:

\noindent
{\bf Theorem 2} { With the scale-invariant activation specified in \cref{scaleac} and a Fourier Neural Operator defined by \cref{defno} under the initialization ensemble \cref{initen}, the kernel and the susceptibilities are given by the following recursion relations:
\begingroup
\small
\begin{equation*}
\begin{split}
\mathcal{K}^{(l+1)}(f,f')&=(\alpha-\beta)^{2}\delta(f-f')C_{R^{(l+1)}}(f) \\
&\int\int \frac{V^{(l)}}{4\pi}\Big(2\sqrt{1-(\rho^{(l)})^{2}}+\rho^{(l)}(\pi+2\arcsin{\rho^{(l)}})\Big)e^{-ifx+if'x'}dxdx'\\&+\alpha\beta \delta(f-f')C_{R^{(l+1)}}(f)\mathcal{H}^{(l)}(f), \\
\int\mathcal{K}^{(l+1)}(f,f')dfdf'&=\frac{\alpha^{2}+\beta^{2}}{2}\int \mathcal{H}^{(l)}(f)df \\
        \chi_{\parallel}(f,f')&=\frac{\mathcal{K}^{(l+1)}(f,f')}{\mathcal{K}^{(l)}(f,f')},\\
    \tilde{\chi}_{\perp}(f,f')&=\mathcal{F}_{x}\mathcal{F}_{x'}\Big(\frac{(\alpha-\beta)^{2}}{4}\Big(1+\frac{2}{\pi}\arcsin{\rho^{(l)}}\Big)+\alpha\beta\Big).
\end{split}
\end{equation*}
\endgroup
where $V^{(l)}=\int \mathcal{H}^{(l)}(f)df$, $V^{(l)}\rho^{(l)}(x,x')=\int \mathcal{H}^{(l)}(f)\cos(f(x-x'))df$. And each $\mathcal{F}_{x},\mathcal{F}_{x'}$ denote the Fourier transform along each axis.

} 
\subsection{Architecture with Residual Connection}

In practice, an FNO keeps only a subset of Fourier modes: after the FFT, all modes above a cutoff are truncated. This truncation can exacerbate spectral bias and hinder the learning of fine-scale features; multi-resolution wavelet-based constructions alleviate this by enriching the representation across scales (\cite{Han2026WaveletMR}), and complementary strategies explicitly promote high-frequency learning in neural operators (\cite{Khodakarami2026HFS}). With no activation (purely linear layers), both pre-activations and outputs stay strictly within this band. Once a nonlinearity is added, however, low frequencies couple to high frequency kernels—as shown in Sections 4.1–4.2 and confirmed empirically in Section 5.1. For inputs of modest extent and large channel width, Theorem 1 implies that this coupling is weak. To retain the efficiency of truncation while still accumulating nonlinear interactions, we adopt a ResNet style update that adds each layer’s pre-activation to its activation output. This helps preserve high frequency information with depth and improves fine scale processing. The rest of this section develops the corresponding theory for residual FNOs. We begin by defining a weighted residual connection for the kernel integration layer, modifying \cref{defno} accordingly.

\begin{equation}
    \mathcal{Z}^{(l+1)}(u):=\mathcal{R}^{(l+1)}(\mathfrak{S}^{(l)}(u))+\tilde{\gamma}\mathcal{Z}^{(l)}(u). \label{resfno}
\end{equation}

where $\tilde{\gamma}$ is a hyperparameter to be set. By extending the arguments of Theorem 1 to the ResNet architecture in Appendix~\ref{app:thm3}, we obtain the following result:

\noindent
{\bf Theorem 3} { With the analytic, origin passing activation specified in \cref{analy} and a Fourier Neural Operator defined based on \cref{defno} and modification of Fourier layers by \cref{resfno} under the initialization ensemble \cref{initen}, the kernel and the susceptibilities are given by the following recursion relations. Here $\gamma=\tilde{\gamma}^{2}$:
\begingroup
\small
\begin{equation*}
\begin{split}
&\mathcal{K}^{(l+1)}\{\mathbf{u}\}(f,f')=\delta(f-f')\Big((\sigma_{1}^{2}C_{R^{(l+1)}}(f)+\gamma)\mathcal{K}^{(l)}\{\mathbf{u}\}(f)\\&+C_{R^{(l+1)}}(f)\sum_{k=2}^{\infty}\frac{\sigma_{k}^{2}}{(n^{(l)})^{k-1}}\sum_{n_{k,1}+\dots+n_{k,l}=k}\frac{(n_{k,1})!\dots(n_{k,l})!}{k!}(\mathcal{H}^{(l)})^{*n_{k,1}}\cdots(\mathcal{H}^{(l)})^{*n_{k,l}}(f)\Big),\\
        &\chi_{\parallel}(f,f')=\delta(f-f')\Big(\gamma+C_{R^{(l+1)}}(f)\sum_{k=1}^{\infty} \frac{\sigma_{k}^{2}}{(n^{(l)})^{k-1}} \\
        &\sum_{n_{k,1}+\dots+n_{k,l}=k}\frac{k(n_{k,1})!\dots(n_{k,l})!}{(k-1)!}\frac{(\mathcal{H}^{(l)})^{*n_{k,1}}\cdots(\mathcal{H}^{(l)})^{*n_{k,l}}(f)}{\mathcal{K}^{(l)}\{\mathbf{u_0}\}(f)}\Big),\\
    &\tilde{\chi}_{\perp}(f,f')=\delta(f-f')\Big(\sum_{k=1}^{\infty}\frac{\sigma_{k}^{2}}{(n^{(l)})^{k-1}} \\ &\sum_{n_{k,1}+\dots+n_{k,l}=k-1}\frac{(n_{k,1})!\dots(n_{k,l})!}{(k-1)!}(\mathcal{H}^{(l)})^{*n_{k,1}}\cdots(\mathcal{H}^{(l)})^{*n_{k,l}}(f)\Big)+(\gamma+2\tilde{\gamma}\sigma_{1})\delta(f)\delta(f').
\end{split}
\end{equation*}
\endgroup
Results for the compact (periodic) domain are provided in Appendix~\ref{app:compact}.
}

\section{Experimental Validations}
In this section we empirically validate the theoretical findings from Section 4 and illustrate how they can be exploited in practice. We run three complementary sets of experiments. First, we verify qualitatively that a non-linear activation couples low and high frequencies, and then test the theory quantitatively by predicting the next layer kernel from the current one via \cref{kerre,kerrelu,kerres} and comparing that prediction with measurements. Next, under fixed hyperparameters, we experimentally confirm that the susceptibilities in the Fourier layers behave as our theory predicts.

\subsection{Experimental Setup and Data Generation}\label{sec:exp_setup}
\paragraph{Input functions.}
Our theory conditions on fixed input functions $\mathbf{u}$ and $\mathbf{v}$ and takes expectations only over random parameter initialization.
To instantiate $\mathbf{u}$ and $\mathbf{v}$ in experiments, we use a synthetic mean-zero Gaussian random field (GRF) as a convenient testbed:
we draw $\mathbf{u}$ (and $\mathbf{v}$ when needed) once from a stationary covariance
$C(x,y)=\exp(-\|x-y\|^{2}/\ell^{2})$ (unless stated otherwise),
and keep the sampled inputs fixed while averaging over $N=100$ independent random initializations.
This separation mirrors the theoretical ensemble and isolates the frequency-transfer effect induced by the network nonlinearity rather than by input variability.

\paragraph{Domain and discretization.}
We work on a periodic domain of length $L$ discretized on an equispaced grid with $N_x$ points.
GRF samples are generated in Fourier space with the corresponding power spectrum and transformed back via FFT, enforcing conjugate symmetry to obtain real-valued signals.
Unless otherwise stated, inputs are normalized to have unit $L^2$-norm before being fed to the network.

\paragraph{Reproducibility.}
Our implementation and experiment scripts are available at \url{https://github.com/xx257xx/Analysis-of-Fourier-Neural-Operators-via-Effective-Field-Theory.git}.

\subsection{Kernel Evolution Prediction}

\paragraph{Overview.}
We study how the layer kernel evolves under three activation families—quadratic, $\tanh$, and ReLU—across varying depths and widths, and then examine a ResNet-style FNO with a $\tanh$ nonlinearity. Beyond qualitative inspection of frequency transfer, we also test the theory quantitatively: given an empirical kernel at layer $l$, we use the recursion derived in Section~4 and Appendix~\ref{app:compact} to predict the kernel at layer $l\!+\!1$, and compare this prediction against measurements under independent random initializations.
\paragraph{Reduced kernel and post-truncation evaluation.}
To isolate the mechanism of frequency coupling beyond the truncation wavenumber $k_{\max}$, we work with the reduced kernel
\begin{equation*}
\tilde{\mathcal{K}}^{(l+1)}\{\mathbf{u},\mathbf{v}\}(f,f')
:=\big\langle \mathcal{F}(\mathfrak{S}^{(l)}{\{\mathbf{u}\}}),
\mathcal{F}(\mathfrak{S}^{(l)}{\{\mathbf{v}}\})\big\rangle_{\mathcal{K}^{(l)}}(f,f'),
\end{equation*}
i.e., the Gaussian expectation term that appears in the kernel update prior to multiplication by the spectral profile $C_R$.
We place particular emphasis on the post-truncation band (e.g., $f\in[17,63]$ for $k_{\max}=16$), where purely linear truncated Fourier layers would yield identically zero energy.

\paragraph{Qualitative frequency transfer: linear vs.\ nonlinear layers.}
Figs.~\ref{fig2} to~\ref{fig5} compare networks with and without nonlinear activation under a step profile $C_R(f)=\mathbf{1}_{\{|f|\le k_{\max}\}}$.
We consider two width--depth pairs, $n\in\{4,64\}$ and $L\in\{2,4\}$, with inputs sampled from a mean-zero Gaussian random field with covariance $C(x,y)\propto \exp(-\|x-y\|^{2}/l^{2})$ and truncation $k_{\max}=16$.
For each configuration, we initialize $N=100$ independent models and plot (in log scale) the kernel magnitude as a function of frequency: dots denote individual runs, while the thick curve shows the ensemble mean. For the quadratic activation $\sigma(x)=x^{2}+x$ (Fig.~\ref{fig2}), the kernel drops to numerical zero immediately after $k_{\max}$ when the activation is removed, reflecting strict band limitation in the linear truncated operator. With the nonlinearity present, nonzero energy persists up to approximately $2k_{\max}$, consistent with the leading double-convolution term in the analytic recursion.
For $\tanh$ (Fig.~\ref{fig3}), the Taylor-series contributions populate substantially higher wavenumbers, so frequency transfer remains visible well beyond $2k_{\max}$.
For ReLU (Fig.~\ref{fig4}), the decay beyond $k_{\max}$ is markedly slower even without residual connections, matching the scale-invariant theory: the update is governed by the spatial correlation $\rho^{(l)}(x,x')$, whose sharp structure in position space induces broad spectral support.

\paragraph{Residual connections and depth-wise retention.}
Figure~\ref{fig5} further shows the effect of residual updates with $\tanh$.
To prevent kernel blow-up, the weight variance is scaled by $1-\gamma$.
Small $\gamma$ yields rapid post-truncation decay with depth, whereas large $\gamma$ preserves substantial high-frequency energy, in line with the residual criticality analysis in Section~4.3.

\paragraph{Width scaling and finite-width fluctuations.}
Two consistent trends emerge across Figs.~\ref{fig2} to~\ref{fig5}.
First, increasing width concentrates the empirical kernels around the ensemble mean, consistent with the theoretical $O(1/n)$ suppression of the four-point vertex (and hence of kernel fluctuations).
Second, larger widths also reduce the absolute magnitude of post-truncation energy, as predicted by the width-suppressed higher-order terms in the analytic recursion.

\paragraph{Quantitative agreement with the kernel recursion.}
We quantitatively test the kernel recursion by a one-step prediction protocol: given the empirical kernel at layer $l$, we apply the theoretical update (Section~4 and Appendix~\ref{app:compact}) to predict the kernel at layer $l\!+\!1$ and compare it against the measured kernel under independent random initializations.
In the post-truncation band $f\in[17,63]$ for $k_{\max}=16$, the measured reduced kernels closely follow the theoretical curves across activation families (tanh, cubic, ReLU) and with/without residual connections.
Fixing the width to $n=32$, we report depths $l\in\{0,2\}$ for tanh/cubic and $l\in\{0,1,2,3\}$ for ReLU/residual; in almost all results the theoretical curve lies within the empirical mean $\pm 1$ std\ band over $N=100$ runs (Figs.~\ref{fig6} to~\ref{fig9}).
Ablations that remove individual activation terms from the theoretical update systematically break this agreement, indicating that all relevant contributions must be retained.
The same level of agreement persists for non-constant truncated spectra, e.g.\ $C_R(f)=2.5/|f|$, both with and without residual connections (Fig.~\ref{fig9}).

\begin{figure}[htp]

     \begin{center}

        \includegraphics[height = 6.9cm]{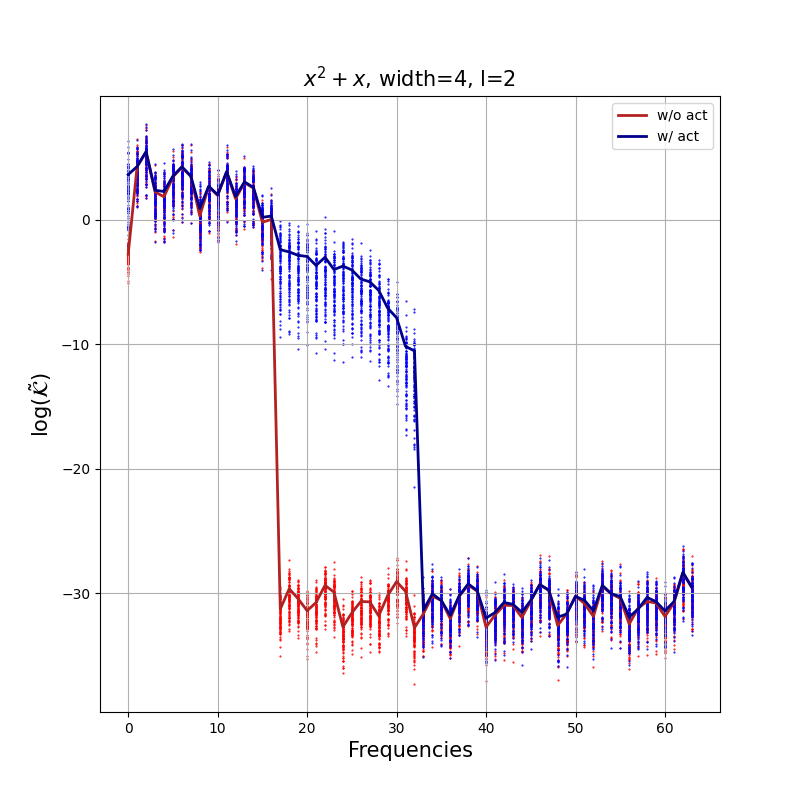}
        \includegraphics[height = 6.9cm]{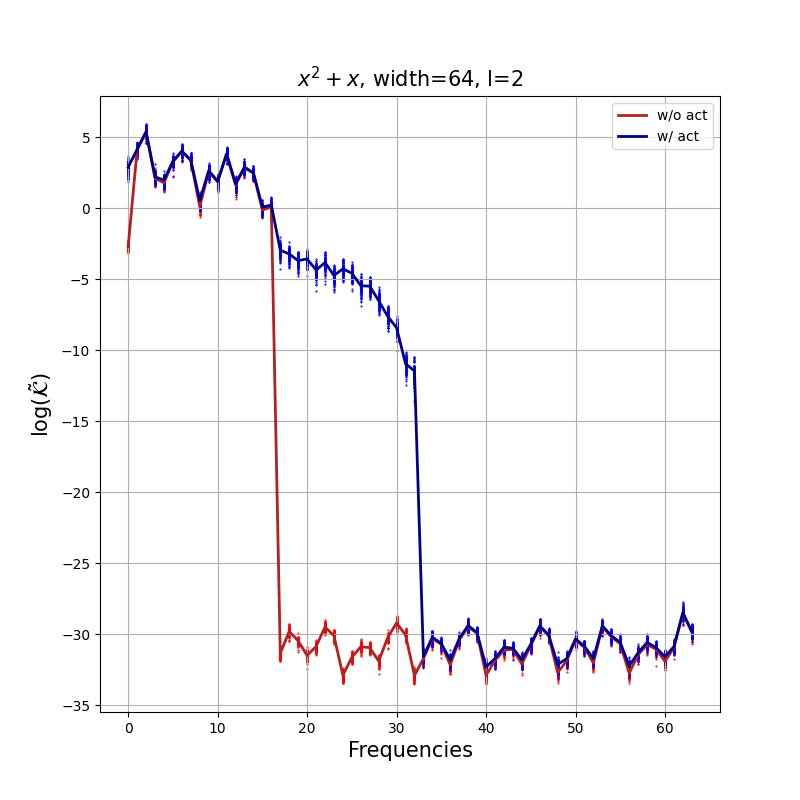}
        \includegraphics[height = 6.9cm]{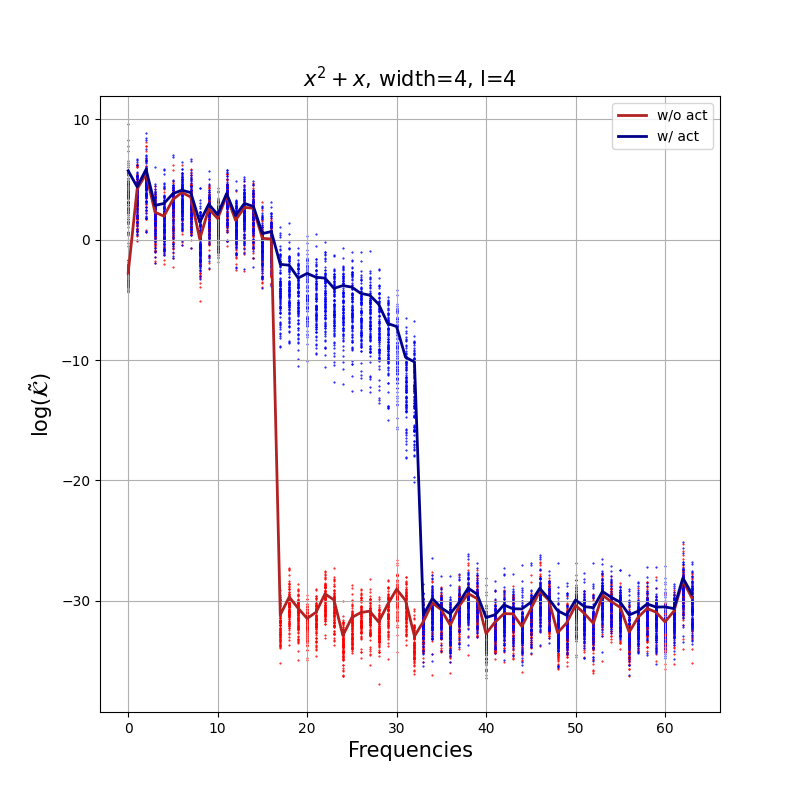}
        \includegraphics[height = 6.9cm]{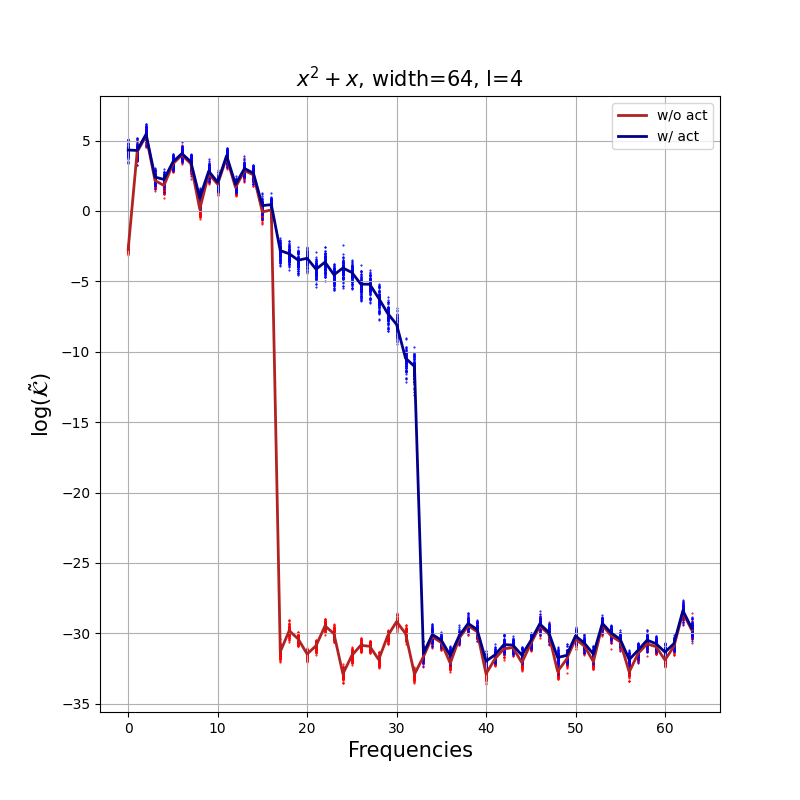}

   \end{center}\caption{Reduced kernel (log scale) quadratic, no residuals, widths $n\in\{4,64\}$ at depths $l=2$ (top) and $l=4$ (bottom).
}\label{fig2}
\end{figure}

\begin{figure}[htp]

     \begin{center}

        \includegraphics[height = 6.9cm]{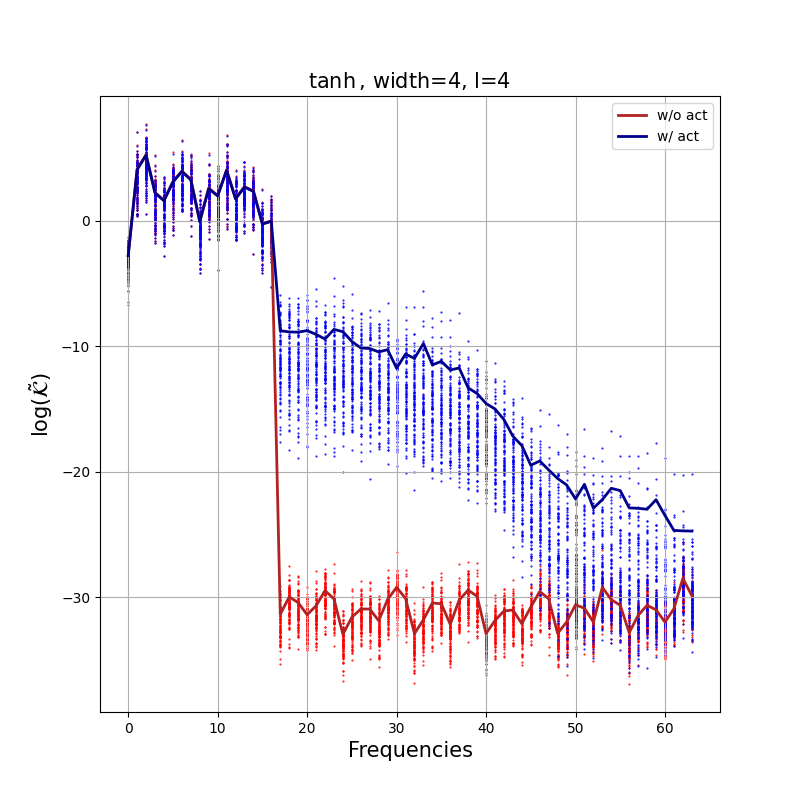}
        \includegraphics[height = 6.9cm]{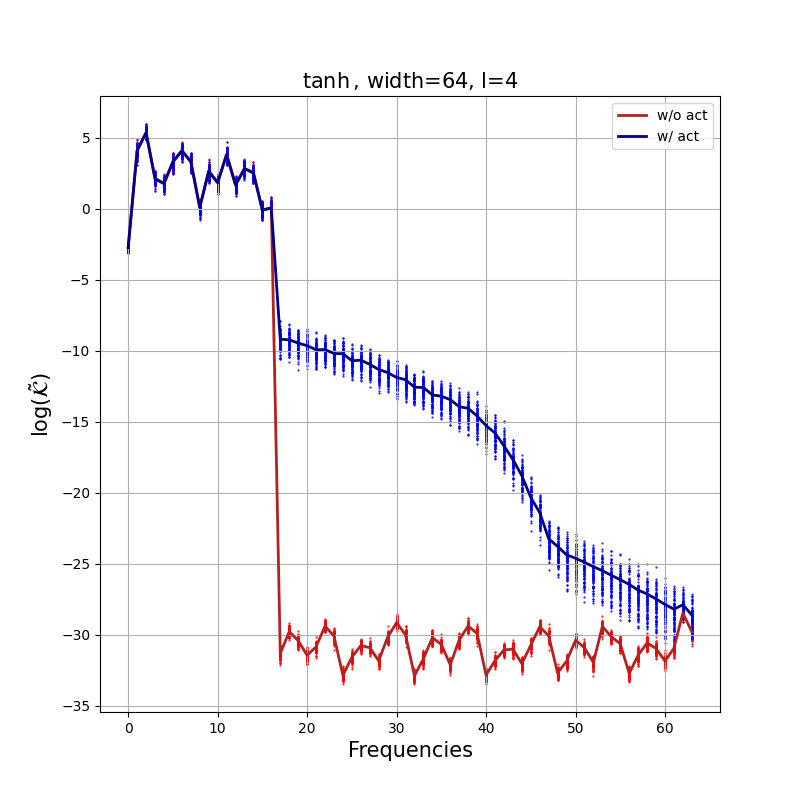}
        \includegraphics[height = 6.9cm]{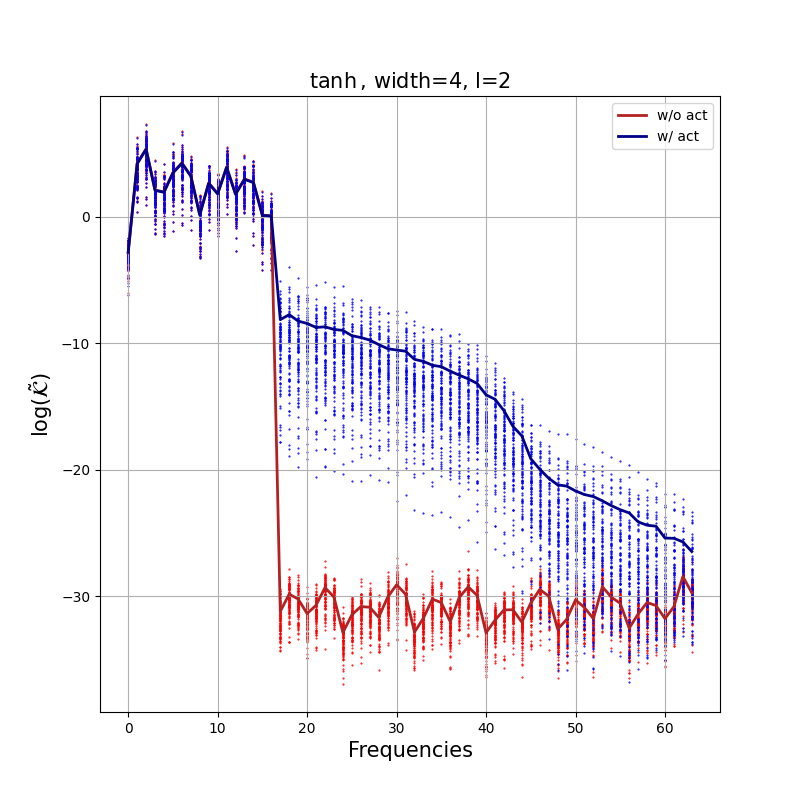}
        \includegraphics[height = 6.9cm]{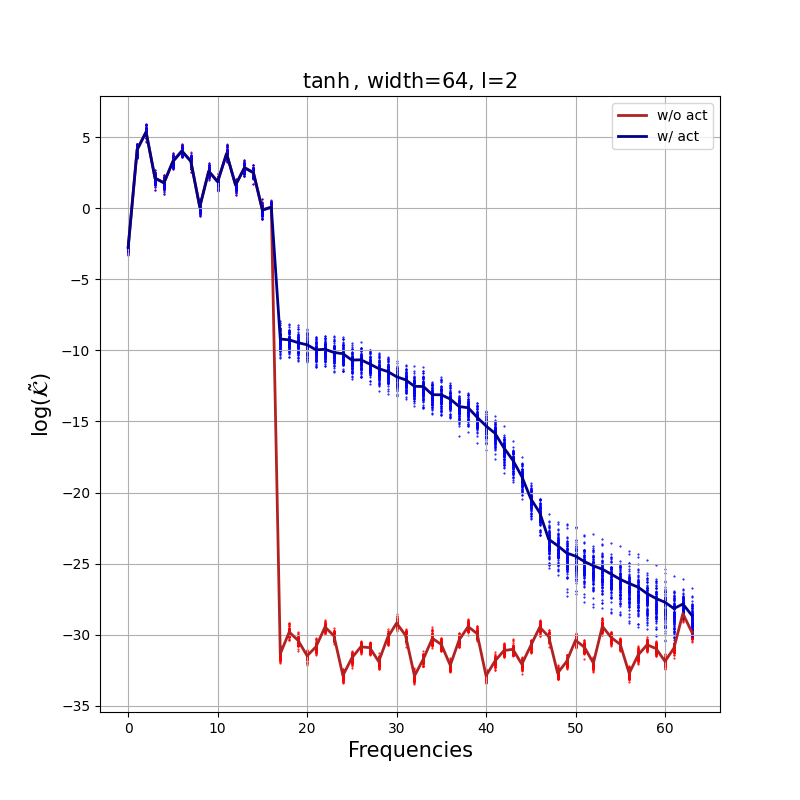}

   \end{center}\caption{Reduced kernel (log scale) tanh, no residuals, widths $n\in\{4,64\}$ at depths $l=2$ (top) and $l=4$ (bottom).
}\label{fig3}
\end{figure}

\begin{figure}[htp]

     \begin{center}

        \includegraphics[height = 6.9cm]{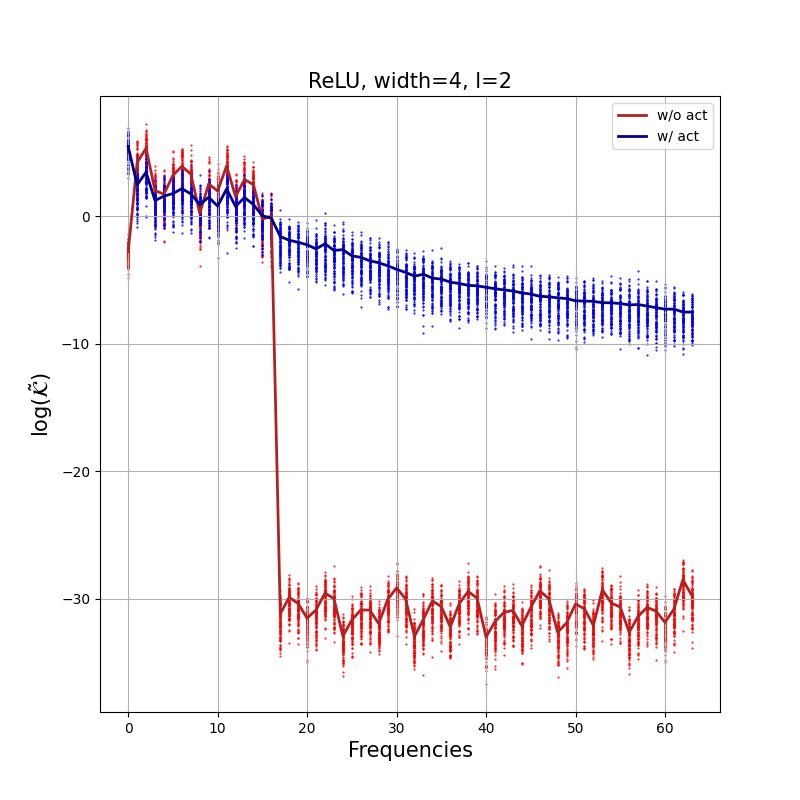}
        \includegraphics[height = 6.9cm]{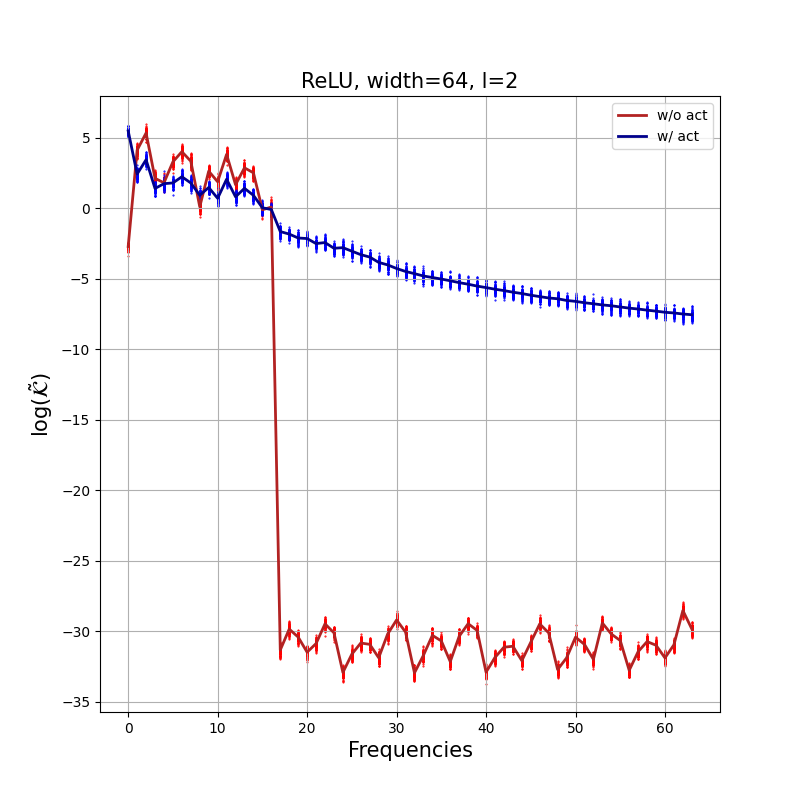}
        \includegraphics[height = 6.9cm]{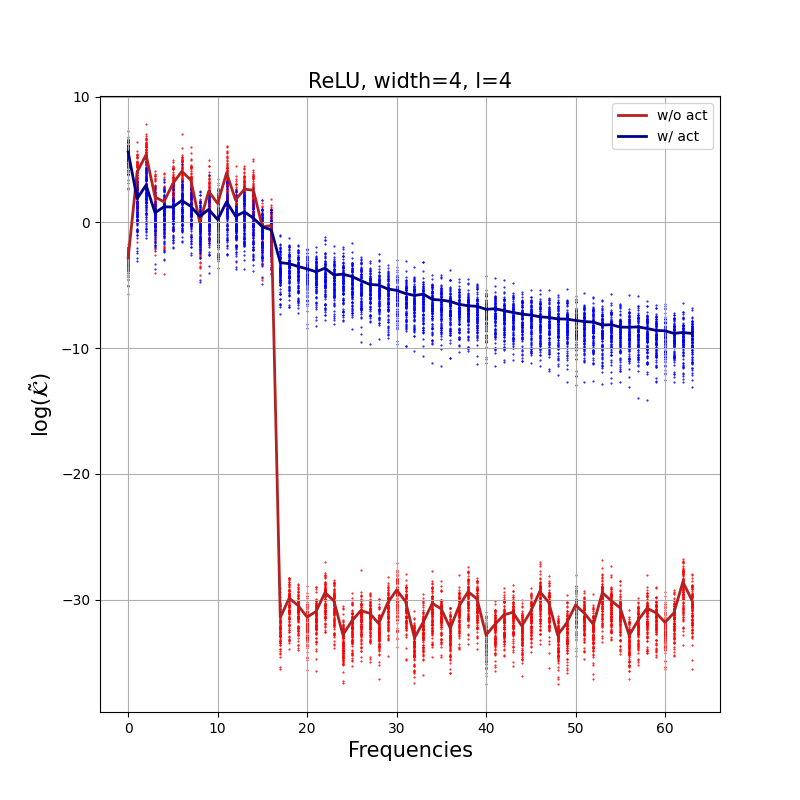}
        \includegraphics[height = 6.9cm]{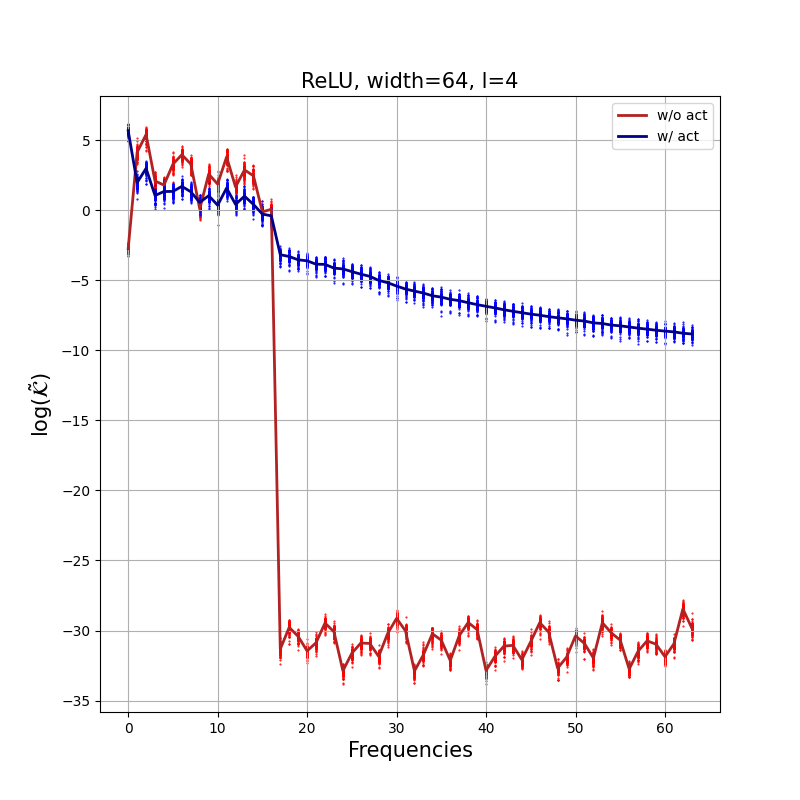}

   \end{center}\caption{Reduced kernel (log scale) ReLU, no residuals, widths $n\in\{4,64\}$ at depths $l=2$ (top) and $l=4$ (bottom).
}\label{fig4}
\end{figure}

\begin{figure}[htp]

     \begin{center}

        \includegraphics[height = 6.9cm]{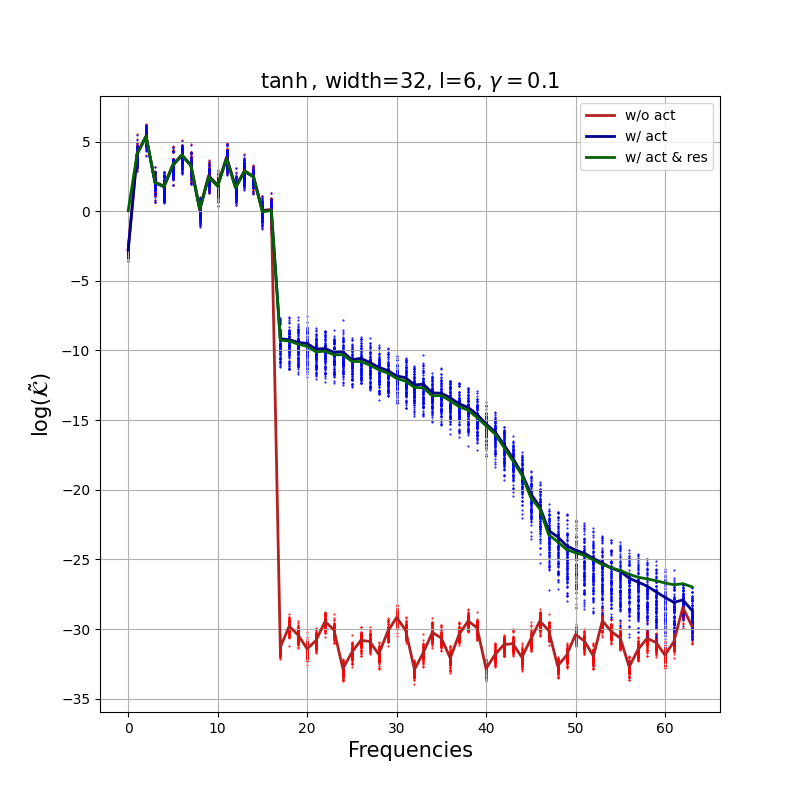}
        \includegraphics[height = 6.9cm]{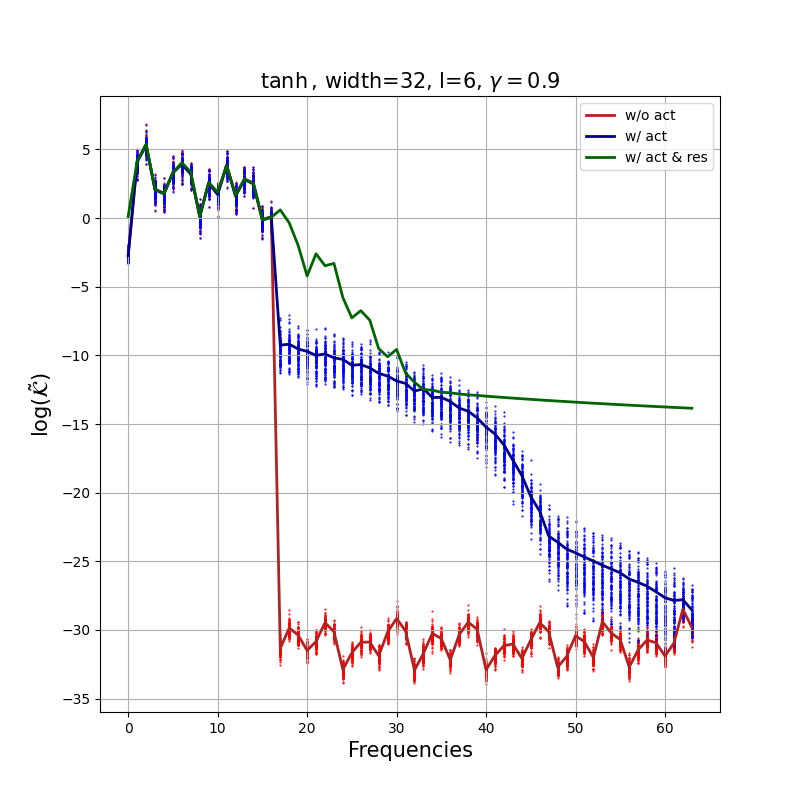}
        \includegraphics[height = 6.9cm]{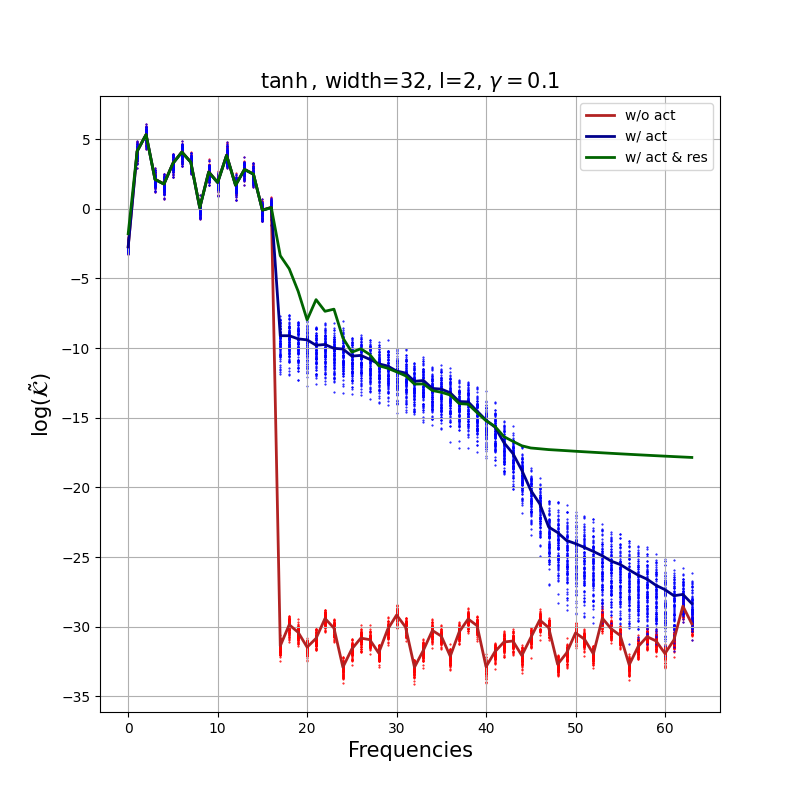}
        \includegraphics[height = 6.9cm]{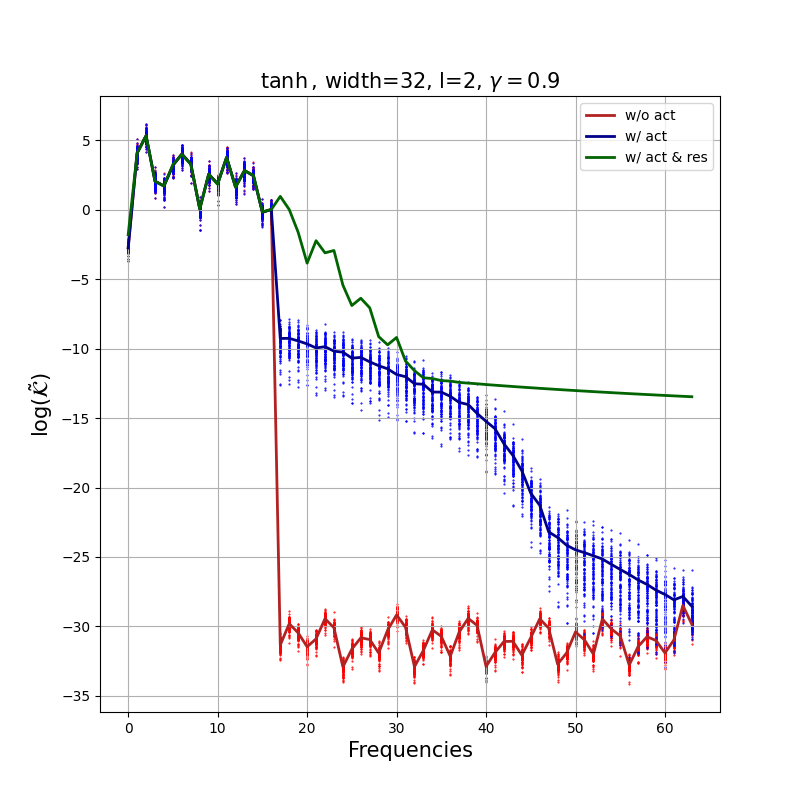}
      
   \end{center}\caption{Reduced kernel (log scale) tanh, with residuals, width $n=32$ with $\gamma=0.1,0.9$ at $l=2$ (top) and $l=6$ (bottom).
}\label{fig5}
\end{figure}

\begin{figure}[htp]

     \begin{center}

        \includegraphics[height = 6.9cm]{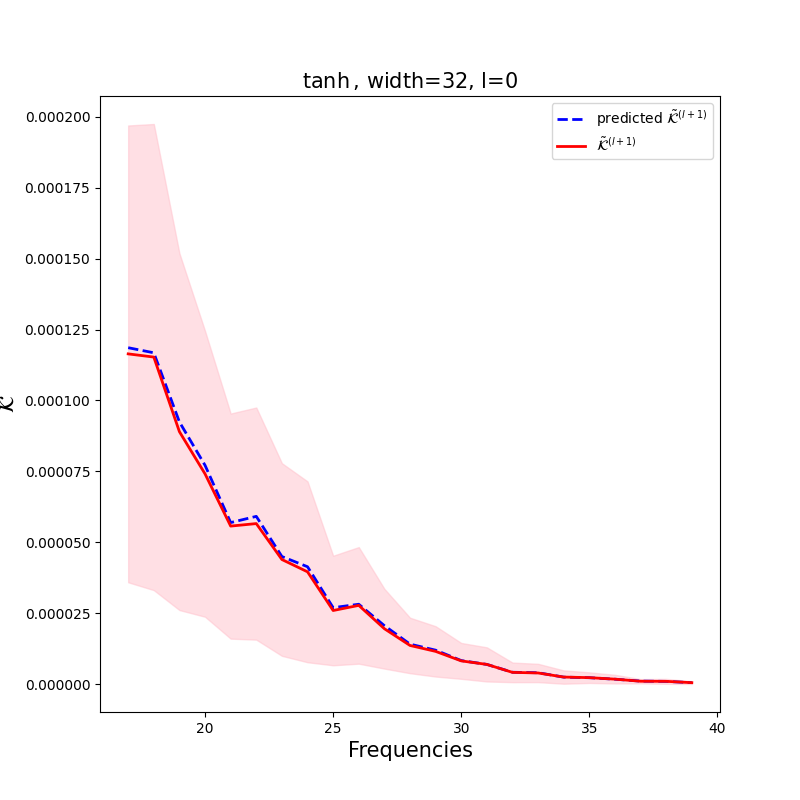}
        \includegraphics[height = 6.9cm]{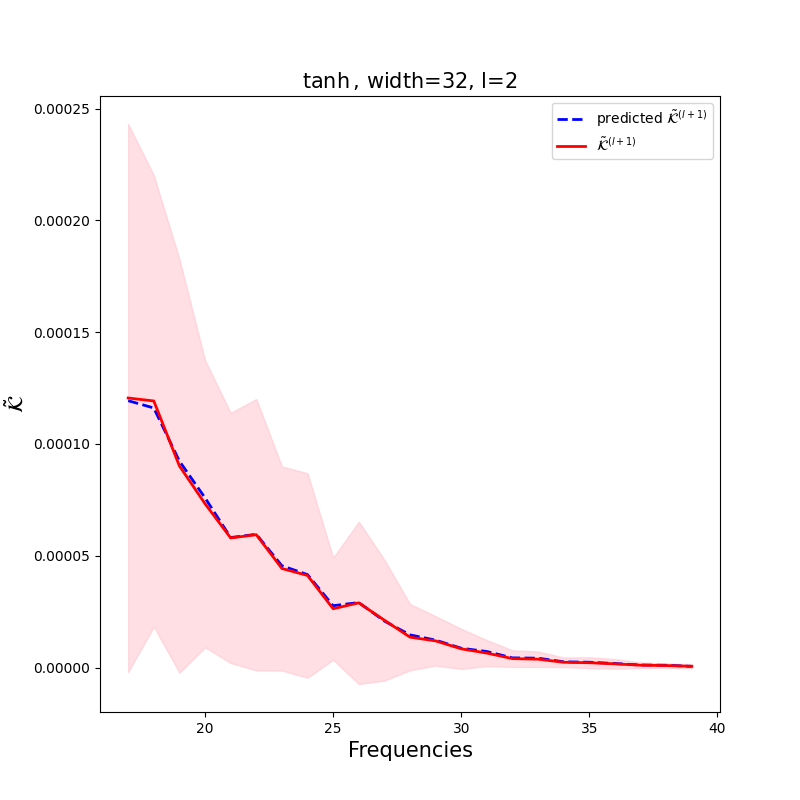}
        \includegraphics[height = 6.9cm]{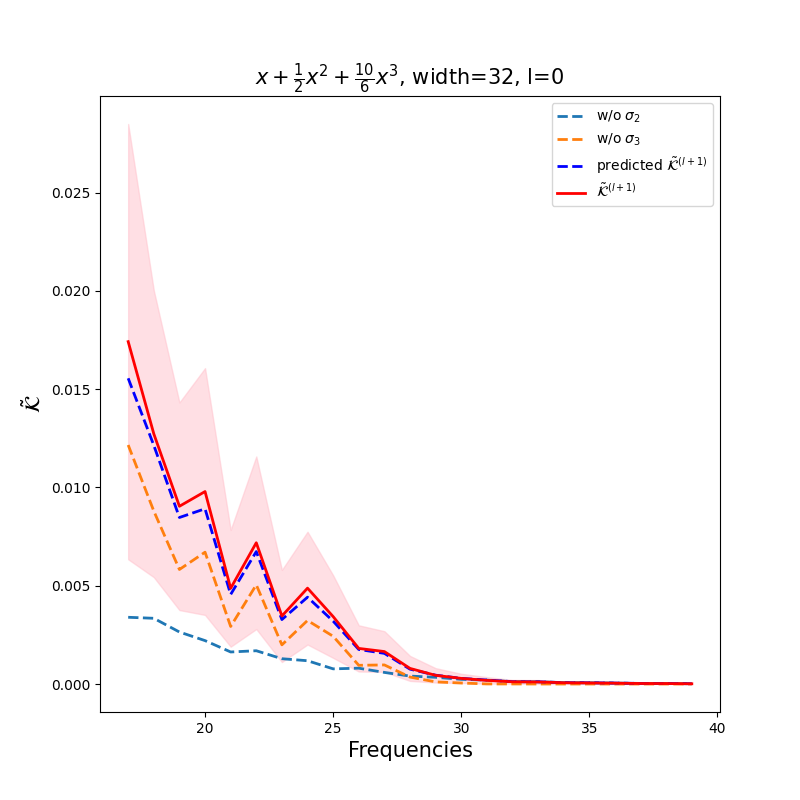}
        \includegraphics[height = 6.9cm]{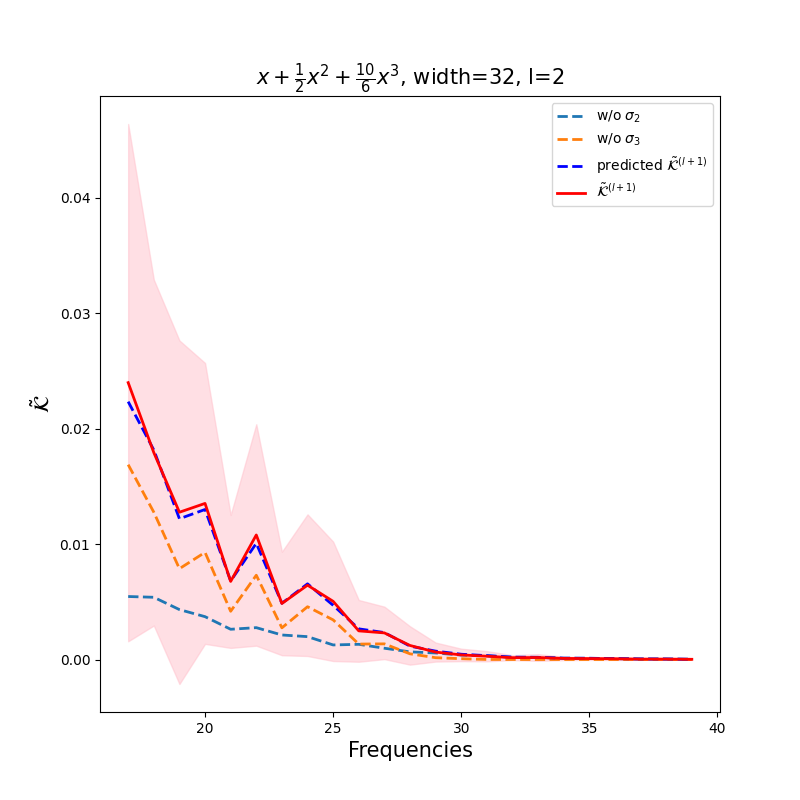}

   \end{center}\caption{Empirical measurement vs. theoretical prediction of reduced kernels (log scale) (width $n=32$, truncation $k_{\max}=16$). $\tanh$ (top), cubic (bottom) with step profile $C_{R}(f)=\mathbf{1}_{\{|f|\le k_{\max}\}}$ $l=0$ (left), $l=2$ (right). Pink shading: mean $\pm 1$ std over $N=100$ initializations; solid curve: theory; dots: measurements.}\label{fig6}
\end{figure}

\begin{figure}[htp]

     \begin{center}

        \includegraphics[height = 6.9cm]{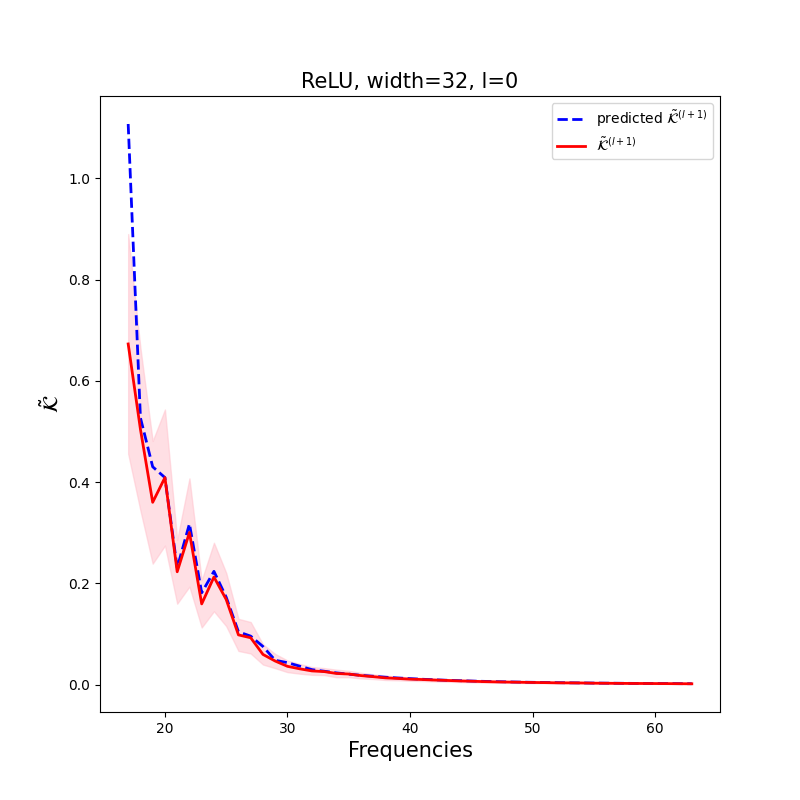}
        \includegraphics[height = 6.9cm]{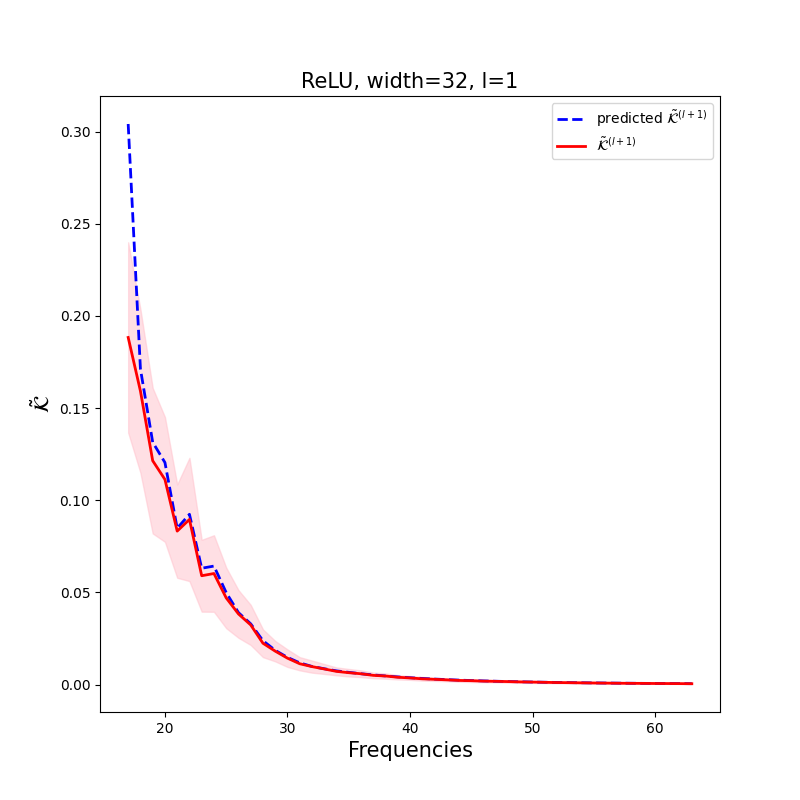}
        \includegraphics[height = 6.9cm]{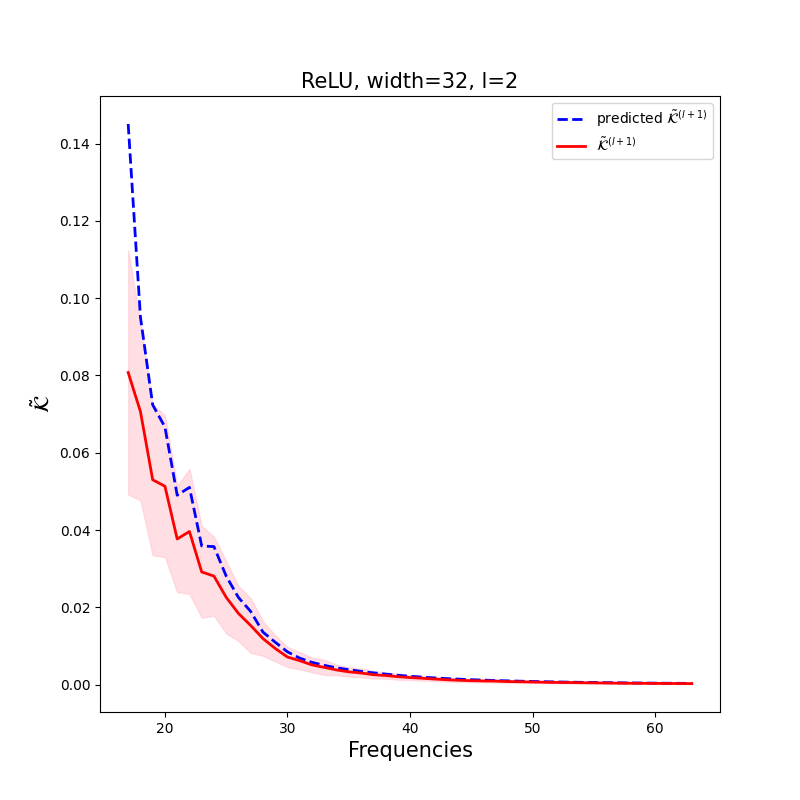}
        \includegraphics[height = 6.9cm]{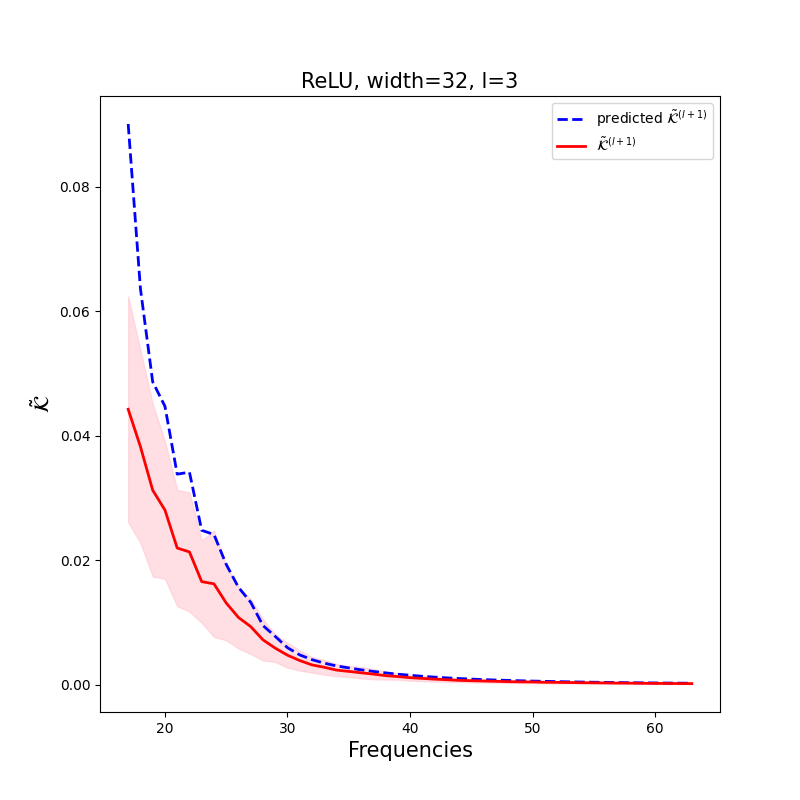}

   \end{center}\caption{Empirical measurement vs. theoretical prediction of reduced kernels (log scale) (width $n=32$, truncation $k_{\max}=16$). ReLU with step profile $C_{R}(f)=\mathbf{1}_{\{|f|\le k_{\max}\}}$ from $l=0$ (top left) to $l=3$ (bottom right). Pink shading: mean $\pm 1$ std over $N=100$ initializations; solid curve: theory; dots: measurements.}\label{fig7}
\end{figure}

\begin{figure}[htp]

     \begin{center}

        \includegraphics[height = 6.9cm]{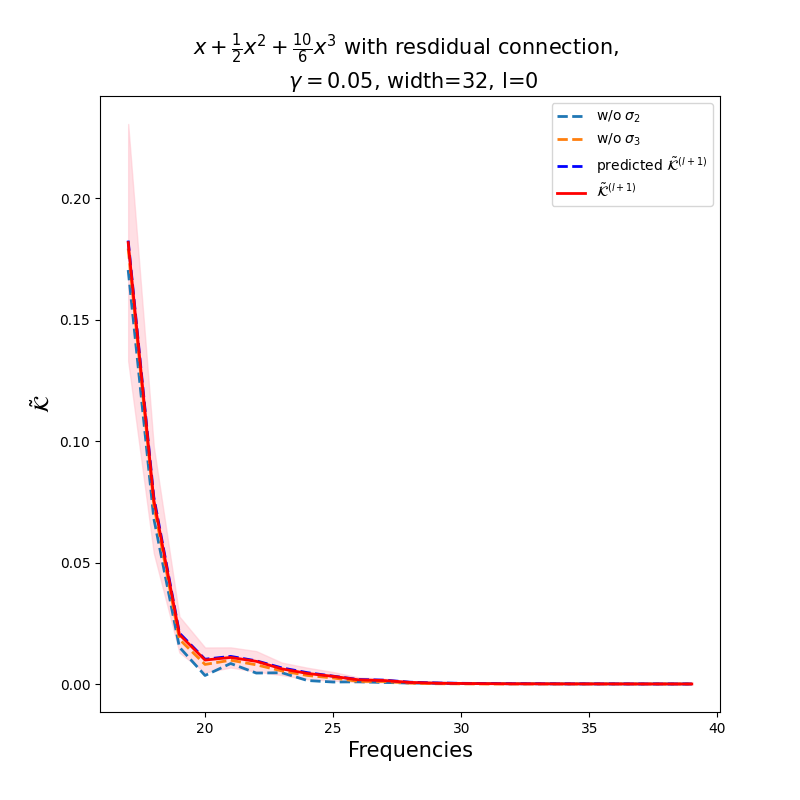}
        \includegraphics[height = 6.9cm]{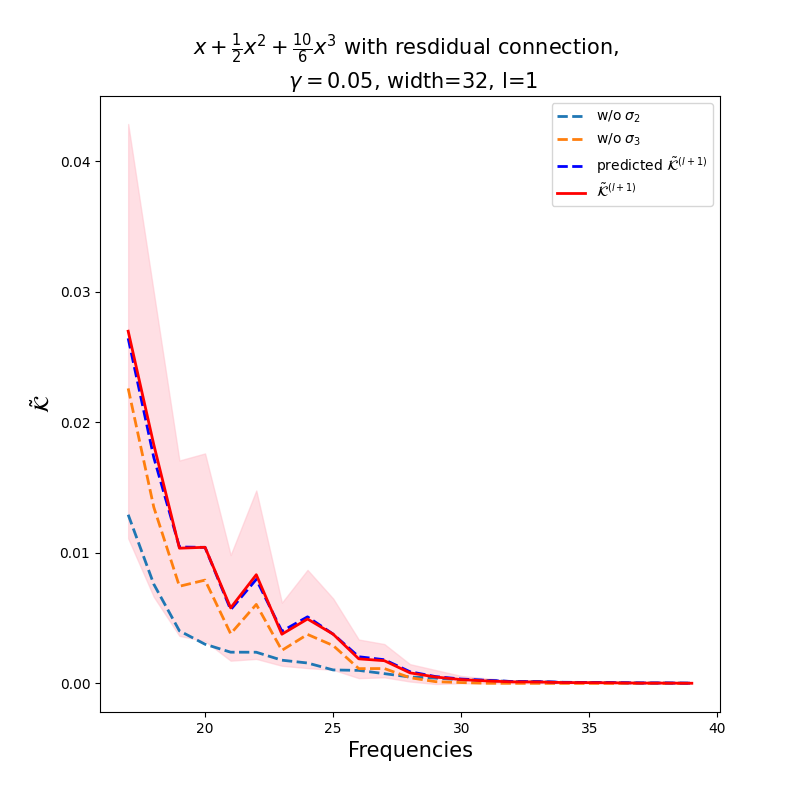}
        \includegraphics[height = 6.9cm]{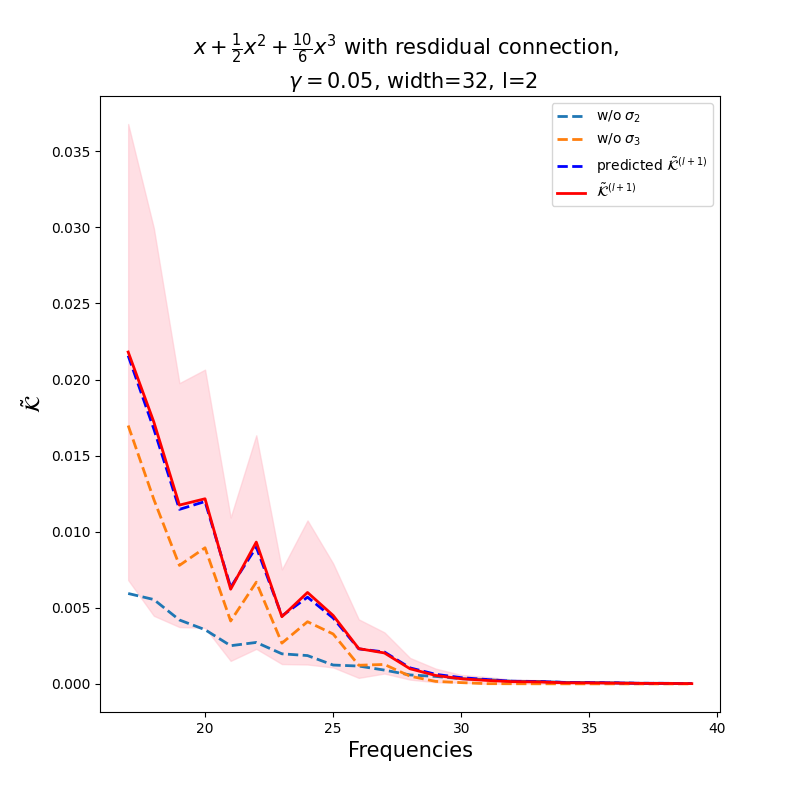}
        \includegraphics[height = 6.9cm]{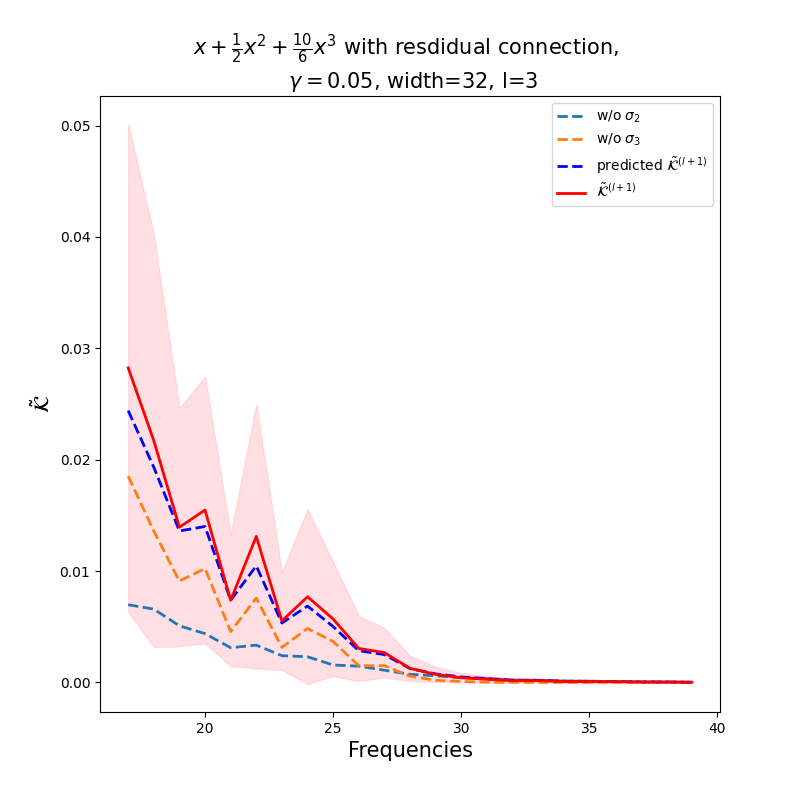}

   \end{center}\caption{Empirical measurement vs. theoretical prediction of reduced kernels (log scale) (width $n=32$, truncation $k_{\max}=16$). Cubic activation with residual connection ($\gamma=0.05$) and step profile $C_{R}(f)=\mathbf{1}_{\{|f|\le k_{\max}\}}$ from $l=0$ (top left) to $l=3$ (bottom right). Pink shading: mean $\pm 1$ std over $N=100$ initializations; solid curve: theory; dots: measurements.}\label{fig8}
\end{figure}

\begin{figure}[htp]

     \begin{center}

        \includegraphics[height = 6.9cm]{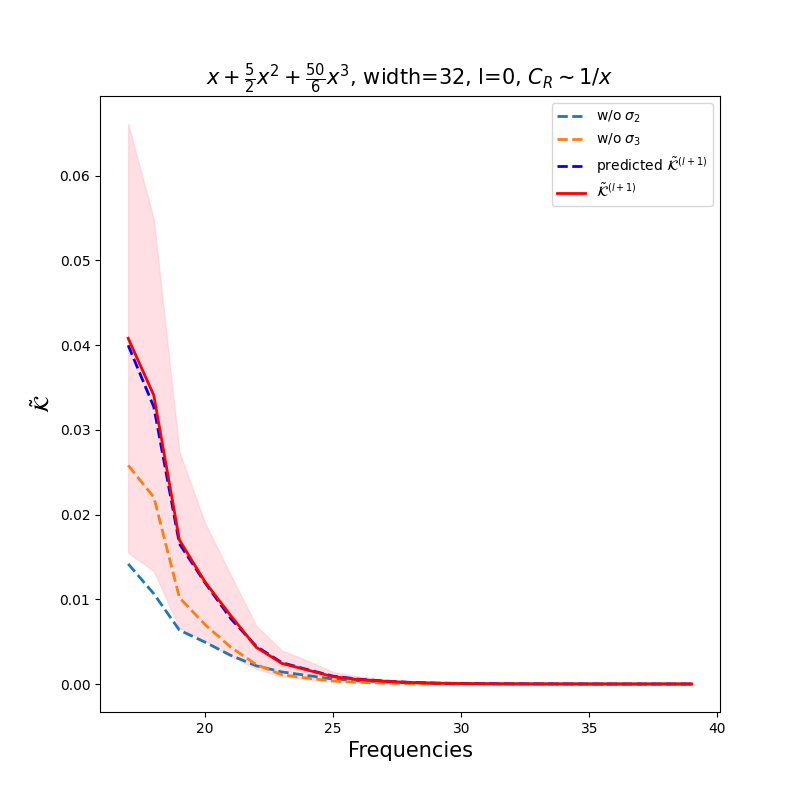}
        \includegraphics[height = 6.9cm]{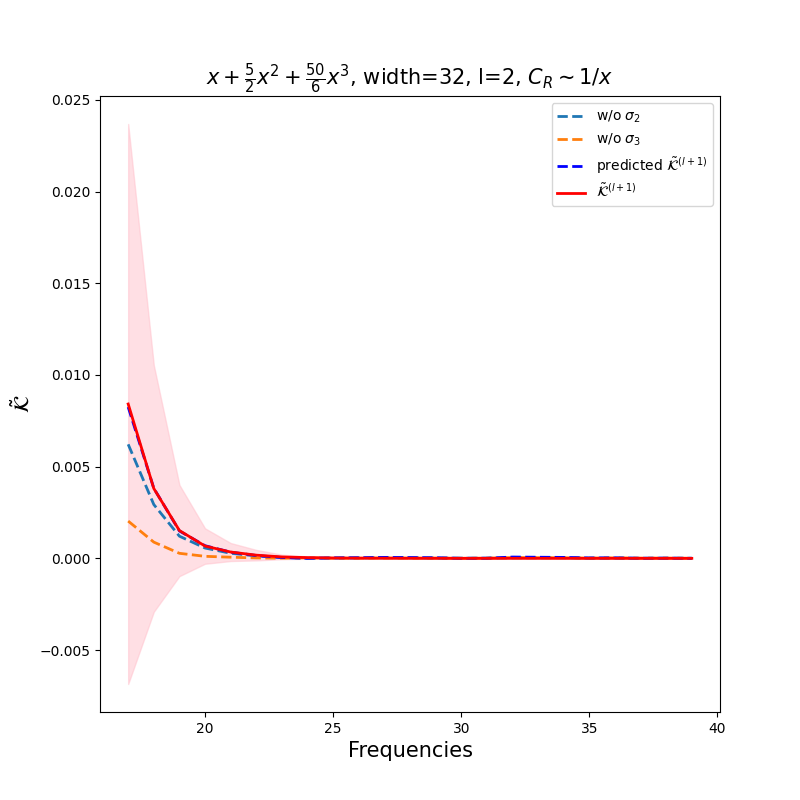}
        \includegraphics[height = 6.9cm]{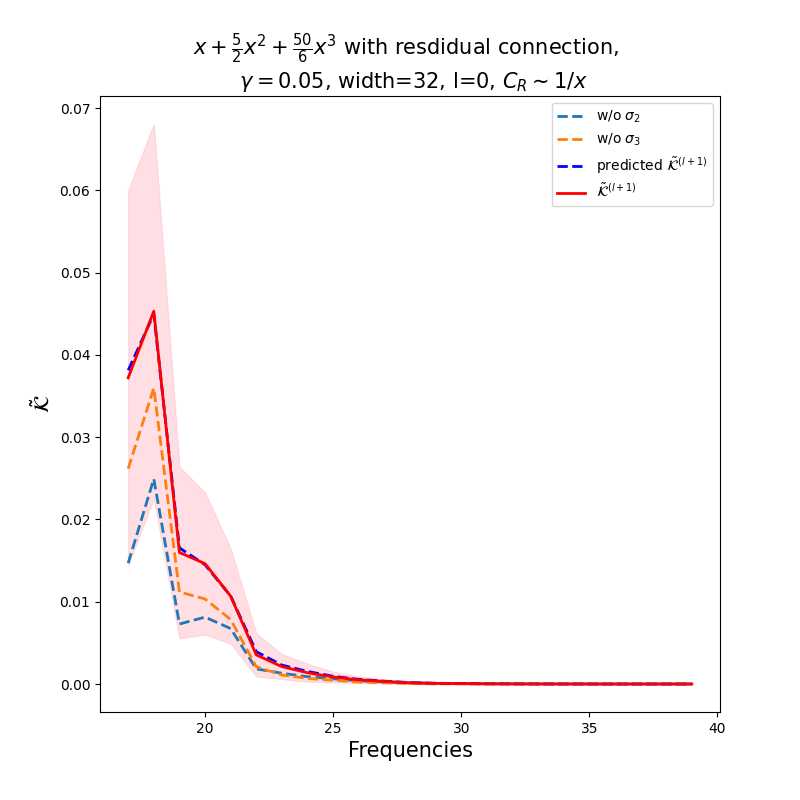}
        \includegraphics[height = 6.9cm]{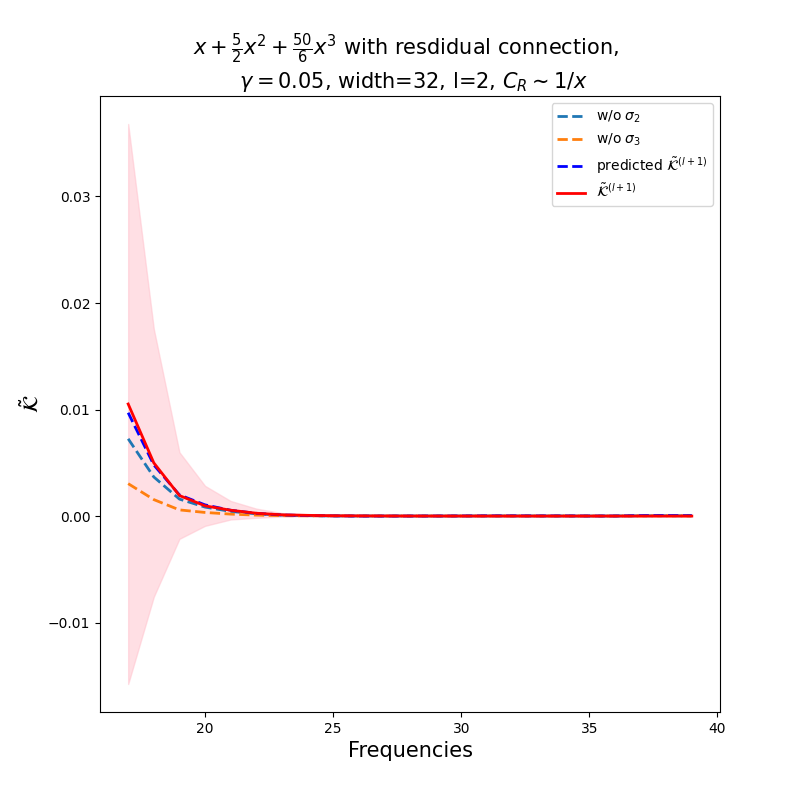}
        
   \end{center}\caption{Empirical measurement vs. theoretical prediction of reduced kernels (log scale) (width $n=32$, truncation $k_{\max}=16$). Cubic activation without (top) and with (bottom) residual connection ($\gamma=0.05$) and step profile $C_{R}(f)\propto 1/|f|$ (truncated at $k_{\max}$) $l=0$ (left), $l=2$ (right). Pink shading: mean $\pm 1$ std over $N=100$ initializations; solid curve: theory; dots: measurements.}\label{fig9}
\end{figure}

\begin{figure}[htp]

     \begin{center}

        \includegraphics[height = 6.9cm]{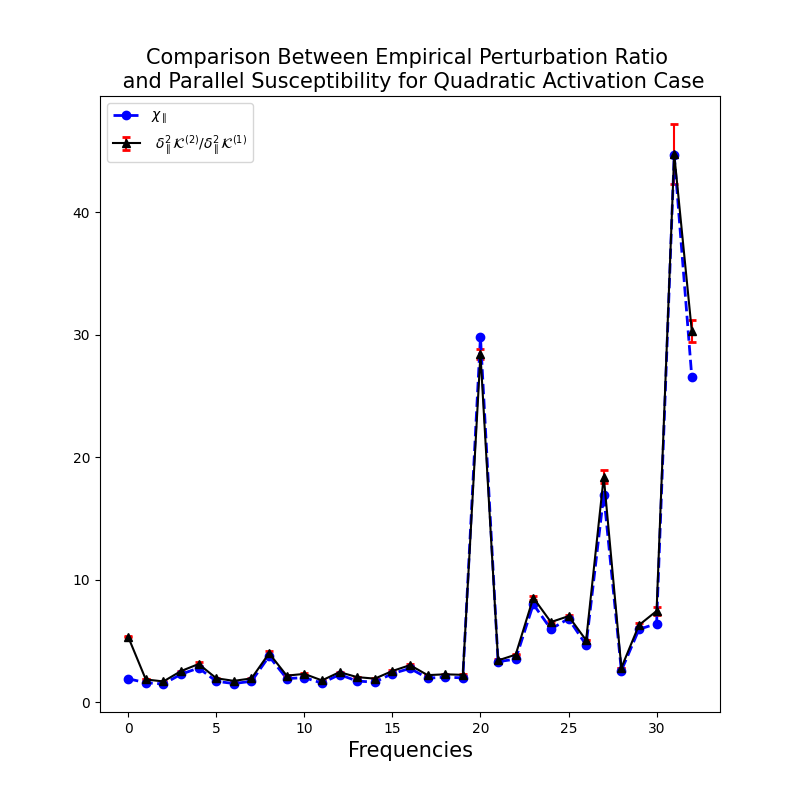}
        \includegraphics[height = 6.9cm]{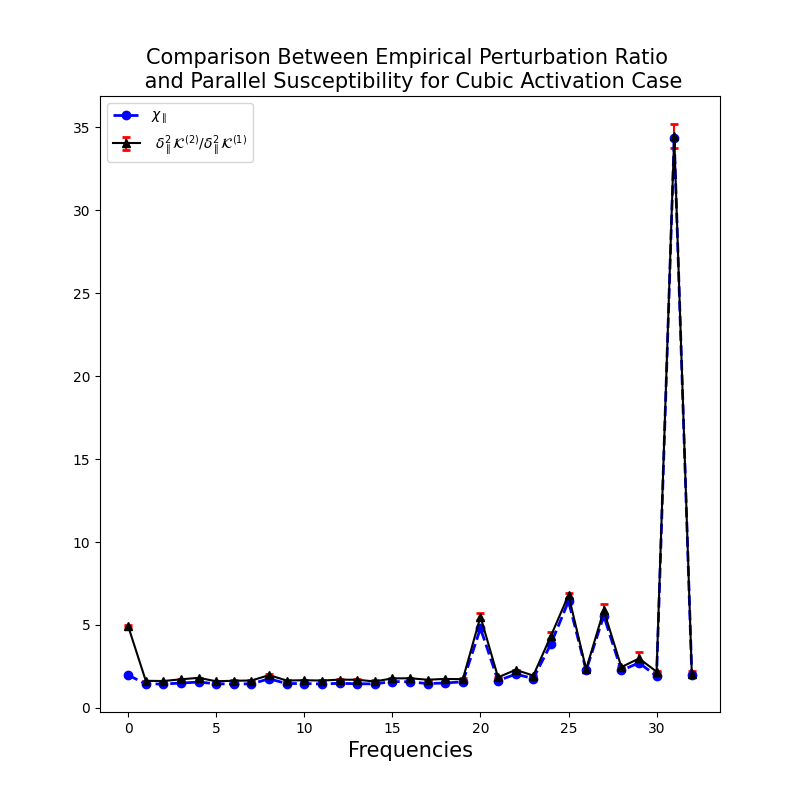}
        
        \includegraphics[height = 6.9cm]{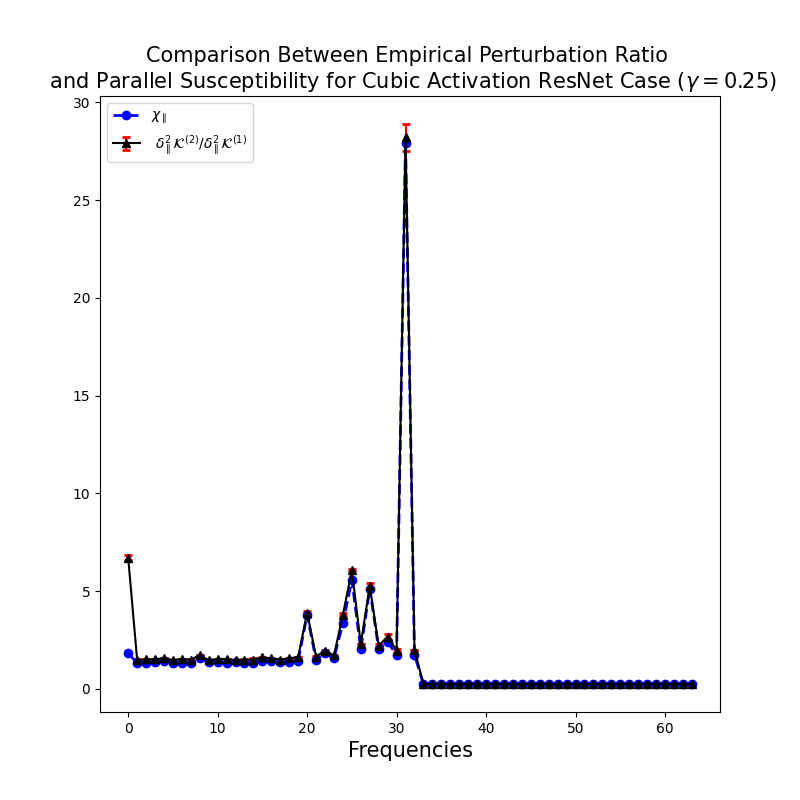}
        \includegraphics[height = 6.9cm]{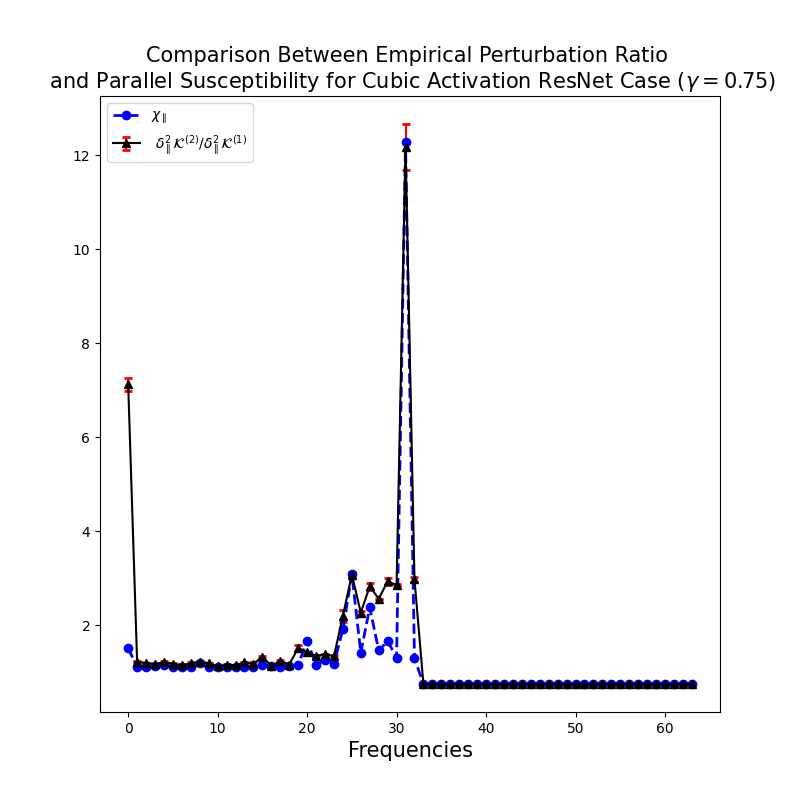}
        
        \includegraphics[height = 6.9cm]{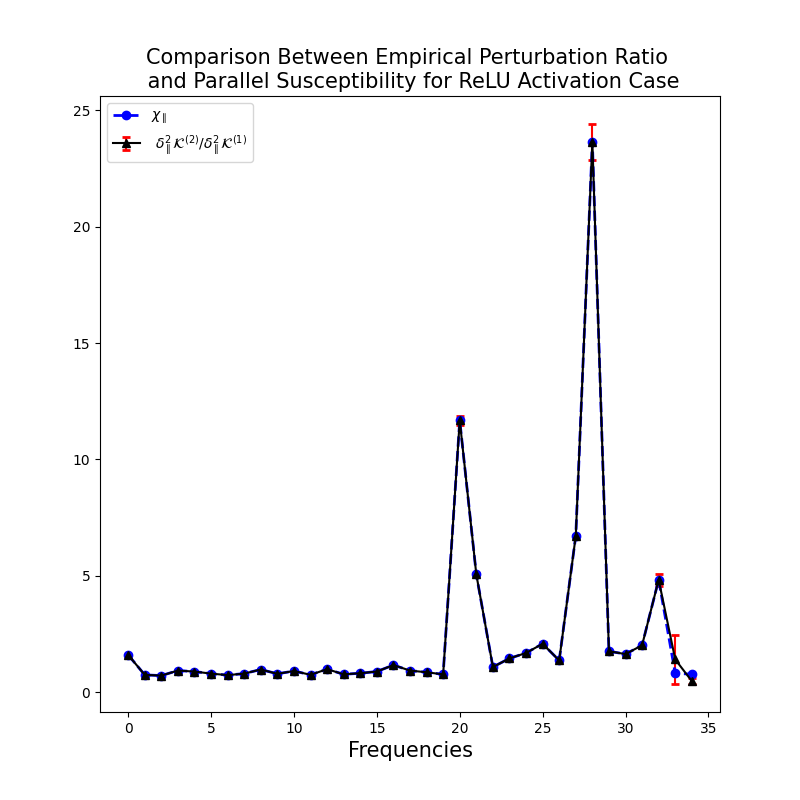}
        \includegraphics[height = 6.9cm]{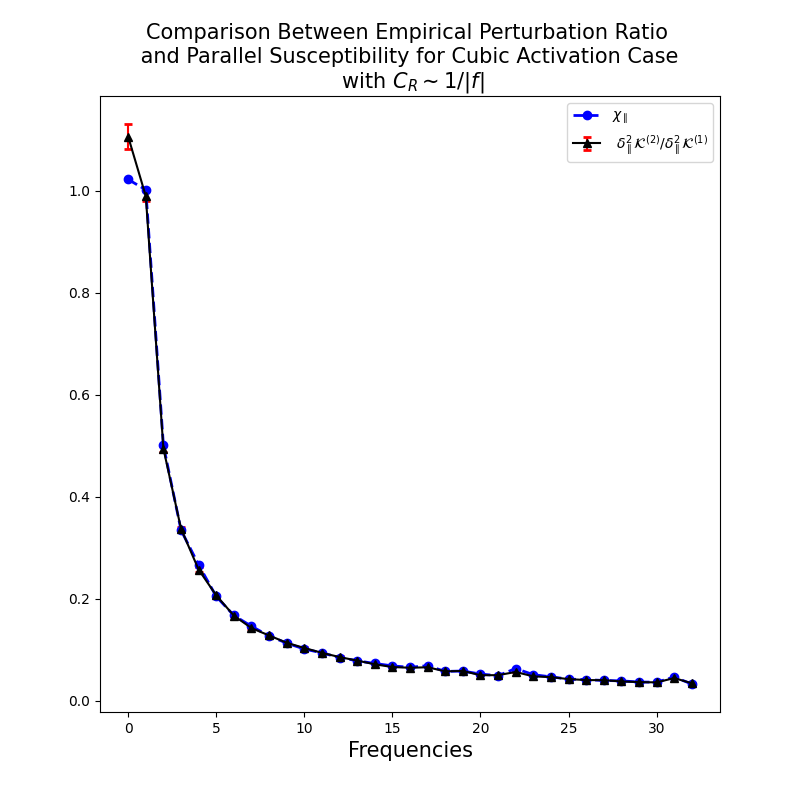}

   \end{center}\caption{Empirical vs. theoretical parallel susceptibility $\chi_{\parallel}$ (log scale; red error bars = $\pm 1$ std, $N=100$) for width $n=32$ and truncation $k_{\max}=32$. Row 1:  quadratic and cubic activations with step $C_{R}$. Row 2: cubic + residual with $\gamma=0.25$ and $\gamma=0.75$. Row 3: ReLU with step $C_{R}$ and cubic with $C_{R}(f)\propto 1/|f|$ (truncated). Used $\varepsilon_{\parallel}=0.01$; Markers = measurements; solid curves = theory.} \label{fig10}
\end{figure}

\begin{figure}[htp]

     \begin{center}

        \includegraphics[height = 6.9cm]{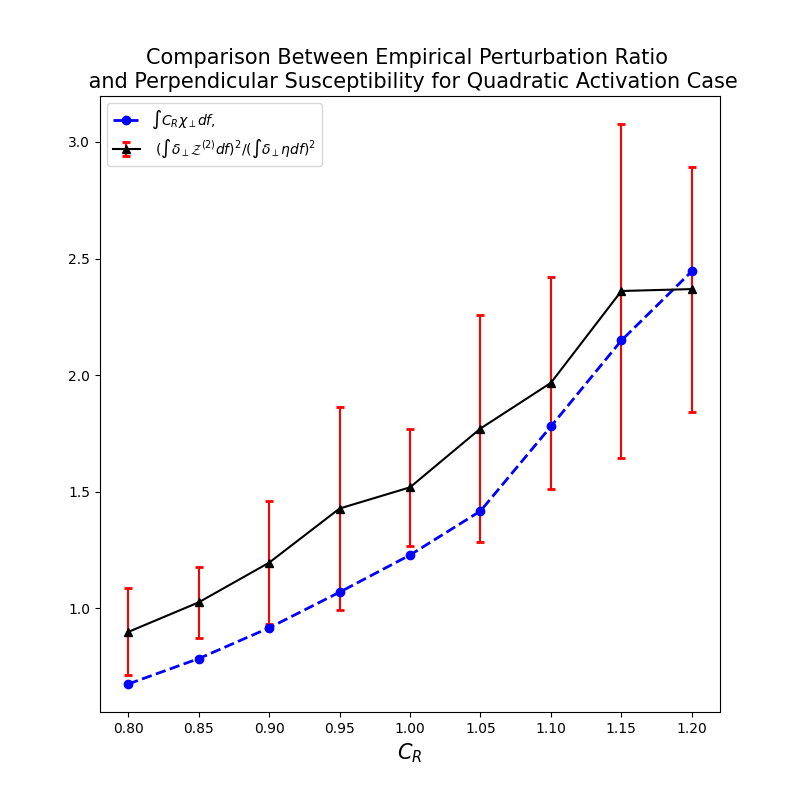}
        \includegraphics[height = 6.9cm]{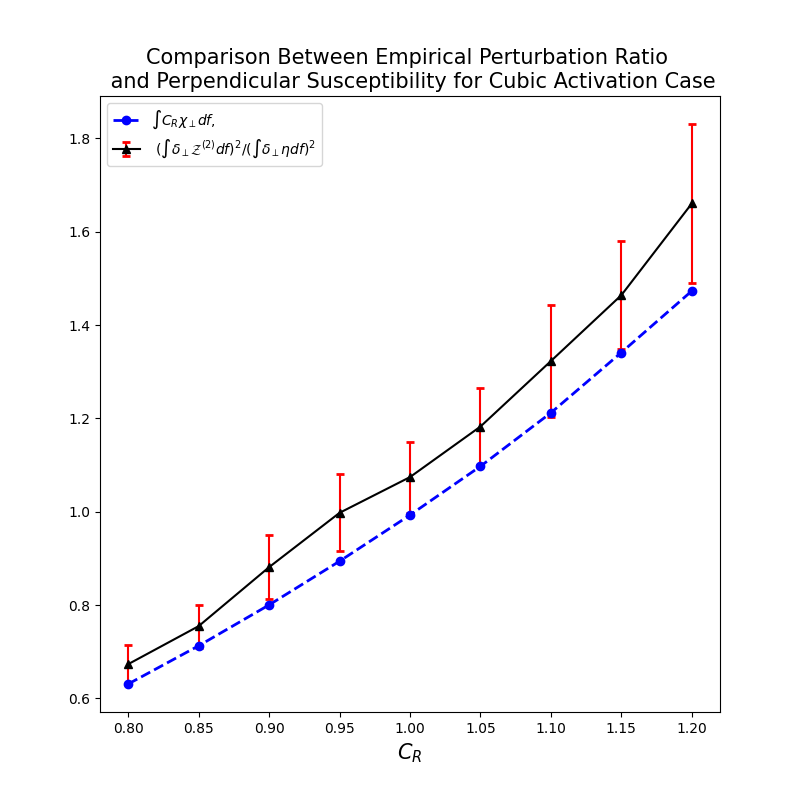}

        \includegraphics[height = 6.9cm]{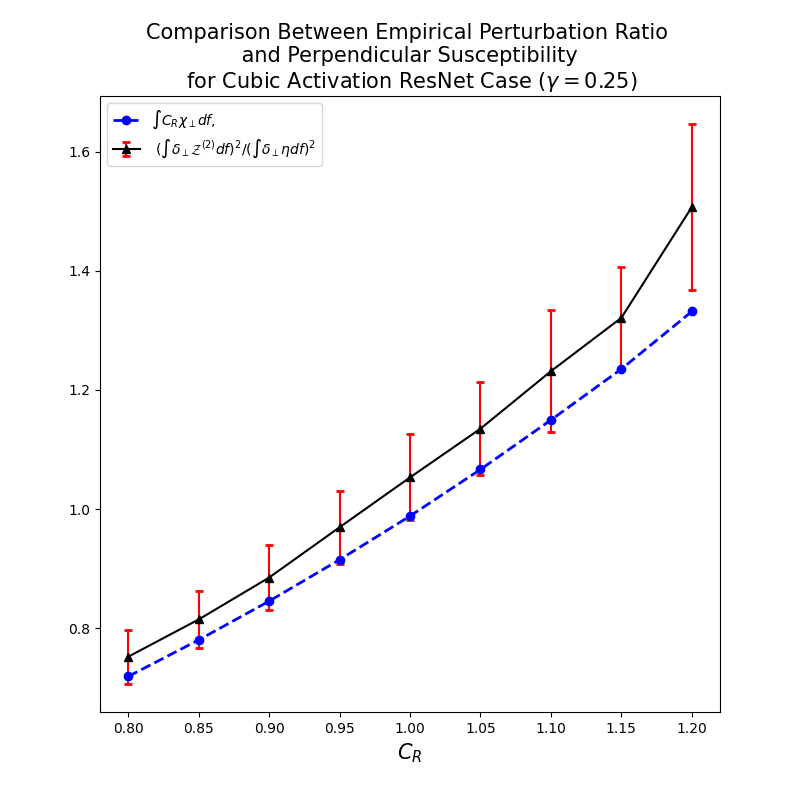}
        \includegraphics[height = 6.9cm]{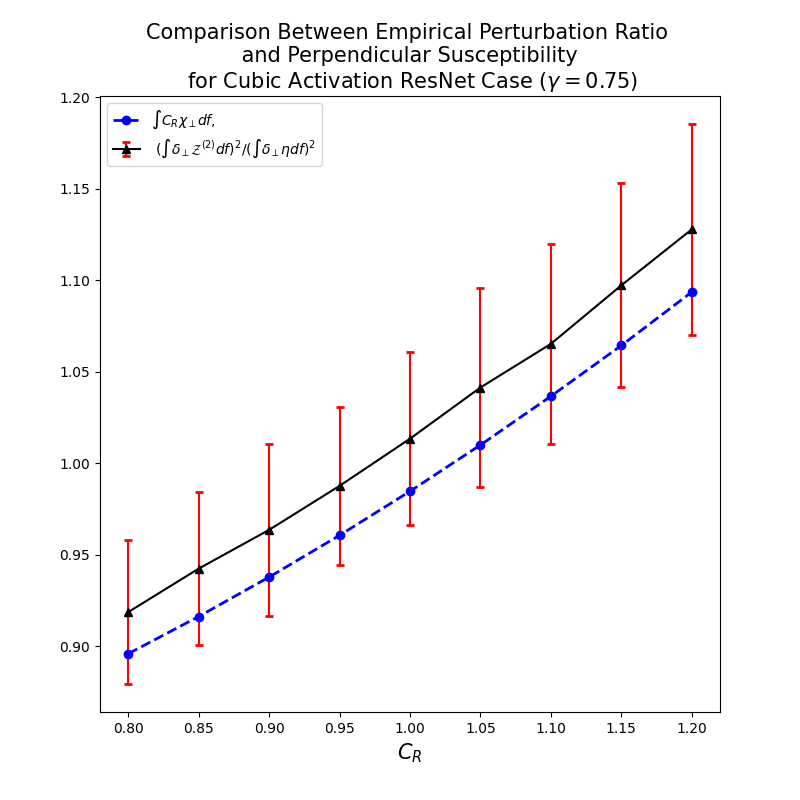}
        
        \includegraphics[height = 6.9cm]{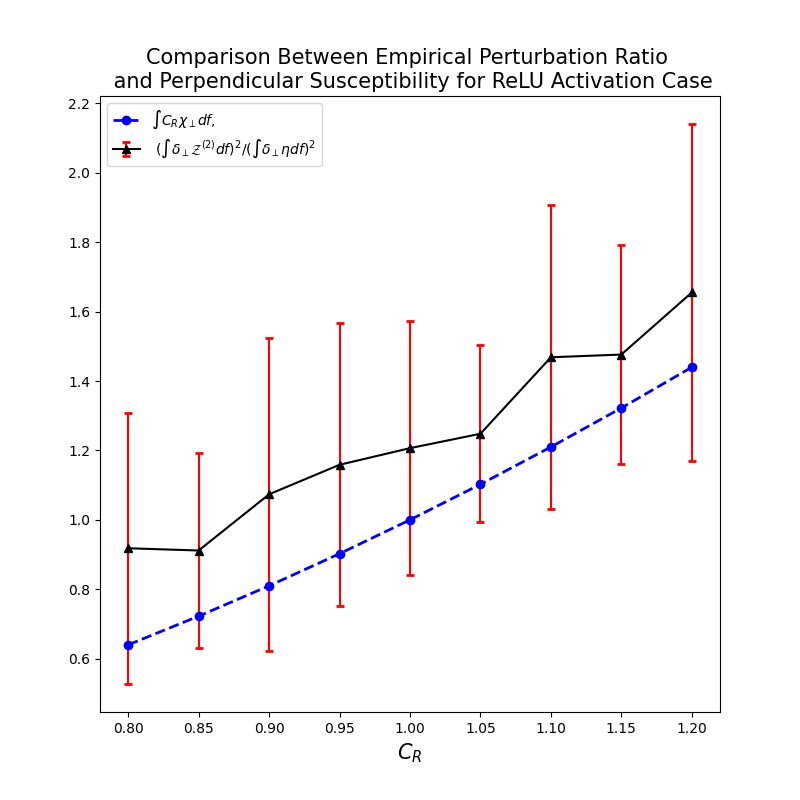}
        \includegraphics[height = 6.9cm]{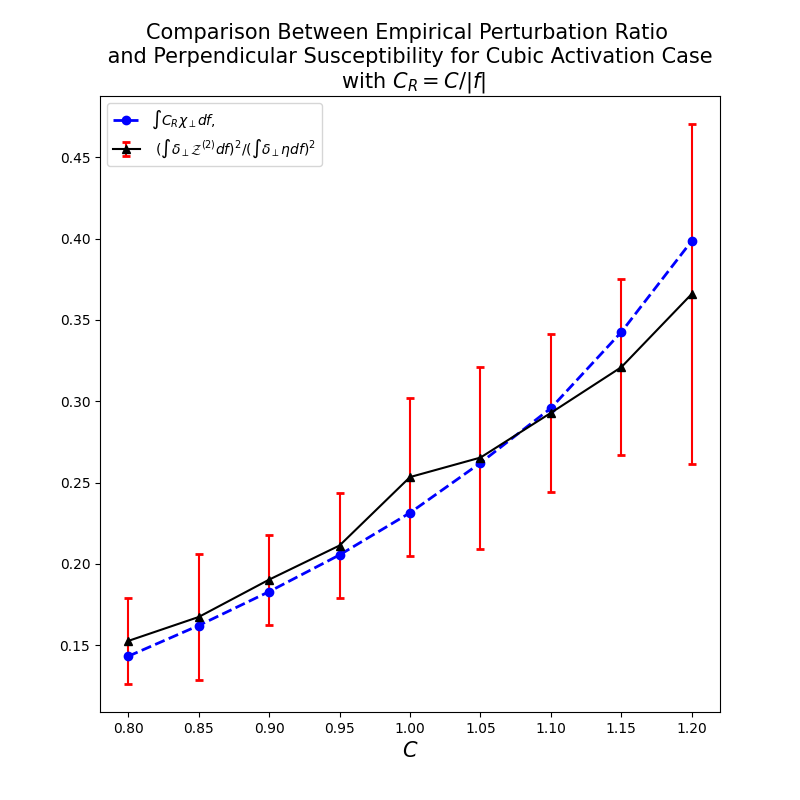}
      
   \end{center}\caption{Empirical vs. theoretical perpendicular susceptibility $\chi_{\perp}$ (log scale; red error bars = $\pm 1$ std, $N=100$) for width $n=32$ and truncation $k_{\max}=32$. Row 1:  quadratic and cubic activations with step $C_{R}$. Row 2: cubic + residual with $\gamma=0.25$ and $\gamma=0.75$. Row 3: ReLU with step $C_{R}$ and cubic with $C_{R}(f)\propto 1/|f|$ (truncated). Used $\varepsilon_{\perp}=10^{-5}$; Markers = measurements; solid curves = theory.} \label{fig11}
\end{figure}

\newpage

\subsection{Susceptibility Prediction}

We validate the parallel ($\chi_{\parallel}$) and perpendicular ($\chi_{\perp}$) susceptibilities predicted by our theory by comparing the theoretical curves with empirical measurements under random initialization. We fix the architecture to truncation $k_{\max}=32$ and width $n=32$, draw inputs from the same Gaussian random field as in Section~5.2, and read out the first Fourier layer.
For the parallel experiment, we add a perturbation of amplitude $\epsilon_{\parallel}=10^{-2}$ along the theory-specified data direction and compute $\chi_{\parallel}(f)$ using \cref{parare2,pararelu,parares}, averaging over $N=100$ seeds.
For the perpendicular experiment, we inject an orthogonal perturbation of amplitude $\epsilon_{\perp}=10^{-5}$ and compute $\chi_{\perp}$ by integrating the resulting power over frequencies as in \cref{perpre,perprelu,perpres}, again averaging over $N=100$ runs.

\paragraph{Activation settings and spectral profiles.}
We test four settings: quadratic, cubic, ReLU, and a ResNet variant with cubic activation and residual gain $\gamma\in\{0.25,0.75\}$.
Except for the decaying-spectrum ablation ($C_R(f)\propto 1/|f|$ with truncation), we use a step profile $C_R(f)=\mathbf{1}_{\{|f|\le k_{\max}\}}$.
For the parallel tests we fix the in-band scale to $C_R=1$, while for the perpendicular tests we sweep $C_R\in[0.80,1.20]$ to probe stability around the criticality condition.

\paragraph{Results.}
Because non-residual models have negligible energy beyond $k_{\max}$, we display their curves only up to $k_{\max}$, whereas residual models retain post-cutoff energy and are therefore plotted up to $2k_{\max}$.
In Fig.~\ref{fig10} and \ref{fig11}, markers denote empirical measurements and solid curves denote theory; error bars indicate $\pm 1$ standard deviation over $N=100$ seeds.
Across all activations, perturbation types, and tested depths, the empirical susceptibilities closely track the theoretical predictions: measurements remain within one standard deviation of the predicted curves throughout the retained band, and, where applicable, the residual model continues to match accurately beyond $k_{\max}$.

As shown in Fig.~\ref{fig10}, under the step profile $C_R(f)=\mathbf{1}_{\{|f|\le k_{\max}\}}$ we observe a pronounced bump in the spectrum near the truncation boundary, in contrast to the smoother behavior under decaying profiles.
We attribute this to an artifact induced by the discontinuity of the step profile: the sharp cutoff introduces oscillatory components and amplifies boundary effects when propagated through the convolutional structure of the kernel recursion.
This observation suggests a practical refinement for initialization: choosing a smoothly decaying $C_R$ near $k_{\max}$ may yield a more stable neural operator by mitigating truncation-edge artifacts.

In Fig.~\ref{fig11}, the empirical susceptibilities are systematically slightly larger than the theoretical predictions.
A plausible explanation is that our prediction retains only the leading two-point statistics (kernel-level theory), while finite-width corrections—controlled by higher-order connected correlators such as the four-point vertex—are neglected.
Incorporating these higher-order terms should improve quantitative accuracy and may account for the observed upward bias.

\subsection{Computational Complexity and Scalability}\label{sec:complexity}

\paragraph{Cost and memory under the baseline (A3).}
Under Assumption~(A3) (frequency-diagonal spectral covariance), the layer kernel remains diagonal in frequency,
so it is represented by a single spectrum $\mathcal{K}^{(l)}(f)$ rather than a dense $\mathcal{K}^{(l)}(f,f')$.
On a $d$-dimensional periodic grid with $N_x^d$ points and a truncation radius $k_{\max}$,
the number of retained modes is $M \asymp (2k_{\max}+1)^d$; storing $\mathcal{K}^{(l)}$ costs $\mathcal{O}(M)$ memory.
This diagonal structure is the main reason the recursions are tractable in practical FNO settings.

\paragraph{Analytic activations (convolution powers).}
For analytic $\sigma$, the recursion in Theorem~1 involves convolution powers of $\mathcal{H}^{(l)}$ up to some order $K$.
Each $k$-fold convolution can be computed via FFTs, since convolution in the frequency domain corresponds to pointwise multiplication in the spatial domain.
Moreover, because there are a total of $2^{(K-1)}$ combinations at each order, summing all of them yields a per-layer cost
\[
\mathcal{O}\!\big(K2^KM \log M\big)
\]
for fixed $K$.
In practice, the effective order can be truncated: the $k$-th term is suppressed by $(n^{(l)})^{-(k-1)}$ (Theorem~1),
so for wide layers only low orders contribute above numerical noise, and we truncate the series once the remaining tail falls below a target tolerance.

\paragraph{Scale-invariant activations (ReLU-type).}
For Theorem~2, the update can be implemented by computing the spatial correlation $\rho^{(l)}(x-x')$ from $\mathcal{H}^{(l)}$
via an inverse FFT, applying the closed-form nonlinearity map in the spatial domain, and transforming back by FFT.
This yields a per-layer complexity $\mathcal{O}(M\log M)$ in any dimension $d$ (up to constant factors).

\paragraph{Four-point vertex.}
Exact evaluation of the four-point vertex is indeed expensive.
However, in the wide-network regime the vertex is $\mathcal{O}(1/n)$ and the kernel recursion closes at leading order,
which is the regime emphasized in this work.
When finite-width corrections are desired, a practical option is Monte-Carlo estimation across initialization seeds:
compute the sample covariance of the metric fluctuation $\Delta\widetilde{\mathcal{G}}^{(l)}$ at selected frequencies,
or estimate only the contracted quantities that enter susceptibilities using stochastic trace estimators.
Both approaches avoid forming a dense four-frequency tensor and keep the cost proportional to the number of probed modes.

\subsection{Training Dynamics}
\paragraph{Stability-criterion calibration.}

In this section, based on the theoretical results derived above, we propose an algorithm for selecting hyperparameters that yield a stable network configuration (i.e. satisfying our criticality criterion). Unlike the FCN case, the spectral profile is a function, which admits infinitely many possible candidate algorithms. Here, we present one concrete choice among them.

From the empirical results in Section 5.3, we observe that when the profile contains an abrupt discontinuity, the parallel susceptibility can exhibit spiky behavior. To avoid this issue, we assume a profile that decays smoothly up to the truncation frequency $k_{\max}$. As a representative example, we consider the following class of profiles:
\begin{equation*}
    \Big\{\frac{s}{1+\exp(-c(\tilde{k}_{\text{decaying}}-|f|))}1_{\{|f|<k_{\max}\}},s>0,c>0\Big\}.
\end{equation*}

This function class is parameterized by a scale factor $s$, a steepness parameter $c$ controlling how rapidly the profile decays, and a fixed decay onset $k_{\text{decaying}}$ ($\tilde{k}_{\text{decaying}}=\frac{k_{\text{decaying}}+k_{\max}}{2}$). In our algorithm, $k_{\text{decaying}}$ effectively replaces the role of the previous $k_{\max}$: the model is treated as handling frequencies up to $k_{\text{decaying}}$, and the parallel susceptibility is enforced to be 1 only up to $k_{\text{decaying}}$. In contrast, the perpendicular susceptibility is considered up to $k_{\max}$.

Algorithm 1 describes the procedure for finding hyperparameters $(s^\star, c^\star)$ that satisfy the susceptibility criteria, given the data distribution, $k_{\text{decaying}}$, $k_{\max}$, and architectural parameters such as width. Based on this algorithm, we conduct the experiment presented in the following paragraph.

\begin{algorithm}[htp]
\caption{Calibration of logistic spectral profile for parallel/perpendicular criticality}
\label{alg:joint_calib}
\begin{algorithmic}[1]
\Require Dataset (or input distribution) $\mathcal{D}$ over functions $u$, cutoffs $k_{\text{decaying}},k_{\max}$, width $n$, depth $L$,
activation class (analytic / ReLU-type / residual), series truncation $K_{\mathrm{trunc}}$ (if analytic),
weights $w(f)\ge 0$ for parallel fitting.
\Require Profile $C_R(f;s,c)=s\,b_c(f)$, $b_c(f)=\big(1+\exp(-c\,g(f))\big)^{-1}$, $g(f)=\tilde{k}_{\text{decaying}}-|f|$.

\Comment{$\tilde{k}_{\text{decaying}}=\frac{k_{\text{decaying}}+k_{\max}}{2}$}
\Ensure Calibrated $(s^\star,c^\star)$.

\State Sample $\{u^{(m)}\}_{m=1}^M\sim\mathcal{D}$ and compute $\hat u^{(m)}(f)$ for $f\in\mathcal{F}$.
\State Compute input-spectrum statistics (mean included):
\[
\mu(f)\leftarrow \frac{1}{M}\sum_{m=1}^M \hat u^{(m)}(f),\qquad
S(f)\leftarrow \frac{1}{M}\sum_{m=1}^M |\hat u^{(m)}(f)|^2.
\]
\Comment{$S(f)=\mathrm{Var}(\hat u(f))+|\mu(f)|^2$}

\State Define the target layer index $\ell$ at which the criterion is enforced (e.g., first Fourier layer).
\State Initialize search range $c\in[c_{\min},c_{\max}]$ and $s\in[s_{\min},s_{\max}]$.

\For{candidate $c$ in a 1D search scheme (grid / golden-section)}
    \State Set $b_c(f)\leftarrow (1+\exp(-c\,g(f)))^{-1}$.

    \Comment{(A) Inner solve: enforce the perpendicular criticality as a scalar root-finding in $s$}
    \State Define the function
    \[
    \Phi_{\perp}(s;c)\;:=\;\sum_{f\in[0,f_{\text{max}}]} C_R(f;s,c)\,\tilde{\chi}_{\perp}^{(\ell)}(f;s,c)\,\Delta f\;-\;1,
    \]
    where $\tilde{\chi}_{\perp}^{(\ell)}$ is computed from the theory (Cor.~1/2/3) using the kernel recursion up to layer $\ell_0$,
    with $C_R(\cdot)=s\,b_c(\cdot)$ and (if analytic) truncation $K_{\mathrm{trunc}}$. $\Delta f$ is the spacing of frequency domain.
    
    \State Solve $\Phi_{\perp}(s;c)=0$ for $s$ via bisection (or safeguarded Newton), yielding $s(c)$.

    \Comment{(B) Evaluate parallel deviation under the perpendicular-calibrated $s(c)$}
    \State Compute $\chi_{\parallel}^{(\ell)}(f; s(c),c)$ for all $f\in[0,f_{\text{decaying}}]$ from the corresponding susceptibility formula
    (e.g. \cref{parare2,pararelu,parares}).

    \State Compute a parallel mismatch score, e.g.,
    \[
    E_{\parallel}(c)\;:=\;\sum_{f\in[0,f_{\text{decaying}}]} w(f)\big(\chi_{\parallel}^{(\ell)}(f; s(c),c)-1\big)^2
    \quad \text{or} \quad
    E_{\parallel}^{\infty}(c)\;:=\;\max_{f\in[0,f_{\text{decaying}}]} \big|\chi_{\parallel}^{(\ell)}(f; s(c),c)-1\big|.
    \]
\EndFor

\State Choose $c^\star\leftarrow \arg\min_c E_{\parallel}(c)$ (or $\arg\min_c E_{\parallel}^{\infty}(c)$), and set $s^\star\leftarrow s(c^\star)$.
\State \Return $(s^\star,c^\star)$.
\end{algorithmic}
\end{algorithm}

\paragraph{Scope of the theory (early-training regime).}
Our kernel and susceptibility-based analysis is primarily valid at initialization and during the early stage of training, as discussed in our previous work~(\cite{Kim:242}).
As optimization proceeds, higher-order correlation functions (e.g., the 4-point vertex) can grow, and the effective kernel may deviate from its initial form, leading to more complex spectral dynamics beyond the present theory.
Accordingly, Algorithm~\ref{alg:joint_calib} should be interpreted as an initialization (or early-training) calibration that enforces criticality at the target layer $\ell$, rather than a guarantee that the criterion remains exactly matched throughout training.

\paragraph{Benchmark results under criterion-matched initialization.}

We compared the training loss and test error of a vanilla FNO and an FNO equipped with our proposed algorithm on the 1D Burgers equation. We used the PDEBench (\cite{Takamoto:22}) 1D Burgers dataset with viscosity $\nu=0.001$ for training. The input corresponds to $T=0$ and the target to $T=0.1$, and we used the full positional grid of 1024 points provided in the original dataset. The training set consists of 1000 samples.

For the model, we used a 1D FNO with $k_{\max}=128$, width $=32$, and 5 Fourier layers, with $\tanh$ as the activation function. For the model with our algorithm, we considered the analytic-activation setting with $K_{\mathrm{trunc}}=3$ and $k_{\mathrm{decaying}}=64$, and searched over $s\in[0,2]$ and $c\in[0.01,3]$, using the $\ell^{\infty}$ parallel mismatch score.

For training, we used the Adam optimizer with learning rate $10^{-3}$ and weight decay $3\cdot10^{-3}$. The batch size was set to 50, and we trained for 1000 epochs. The training objective was the relative $L^2$ error. The training-loss curves are shown in Fig.~\ref{fig:trainloss}, and the test performance is reported in Table~\ref{tab:burgers_testloss}.

As shown in Fig.~\ref{fig:trainloss}, the FNO initialized via our criterion-matched calibration exhibits markedly more stable training dynamics: the shaded variability band across seeds is tightly concentrated around the mean throughout training, indicating low sensitivity to random initialization. In addition, the calibrated model converges faster and achieves a lower final training loss than the vanilla baseline.

Table~\ref{tab:burgers_testloss} reports the mean and standard deviation of the relative $L^2$ error evaluated on an independent test set of 1000 samples drawn from PDEBench. The calibrated model attains a lower test loss than the vanilla FNO, demonstrating that the proposed initialization calibration improves not only early-training stability but also downstream generalization performance.

\begin{figure}[t]
  \centering
  \includegraphics[width=0.78\linewidth]{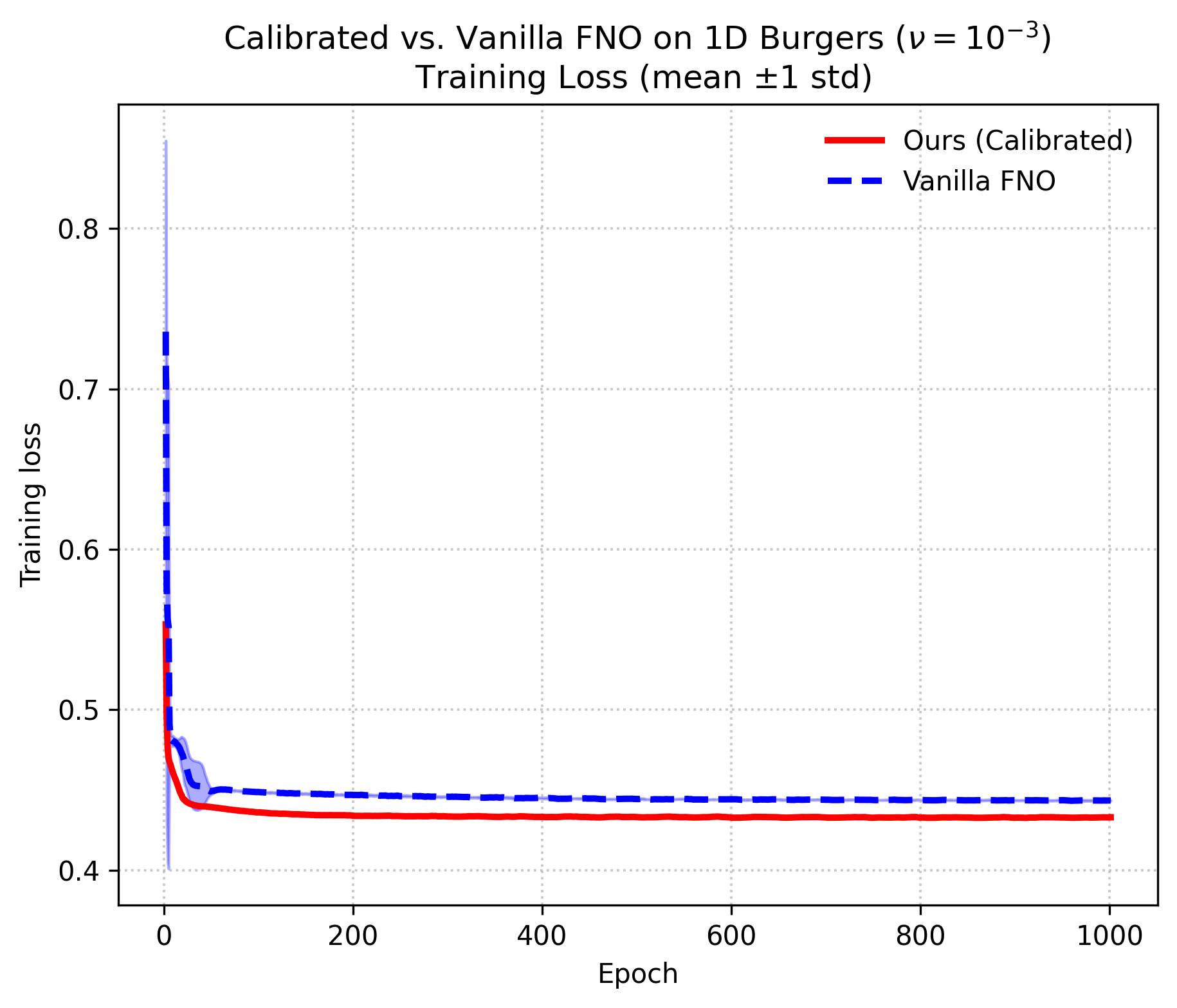}
  \caption{
    \textbf{Training loss dynamics (mean $\pm 1$ std).}
    Curves show the mean training loss across 5 repeated random initializations,
    with shaded regions indicating $\pm 1$ standard deviations.
    The calibrated model (ours) exhibits more stable optimization and faster reduction in training loss compared to the vanilla FNO baseline.
  }
  \label{fig:trainloss}
\end{figure}

\begin{table}[t]
\centering
\begin{tabular}{lcc}
\hline
Model & Mean test loss & Std. dev. \\
\hline
Vanilla FNO       & 0.451 & $3.32\times 10^{-5}$ \\
Calibrated (Ours) & 0.432 & $1.24\times 10^{-4}$ \\
\hline
\end{tabular}
\caption{Test set loss after training (mean and standard deviation over 5 random initializations).}\label{tab:burgers_testloss}
\end{table}

\section{Conclusion}
We analyze neural operators directly in function space, characterizing their frequency-domain kernels and susceptibilities without resorting to discretization. The theory reveals that convolution in Fourier Neural Operators induces frequency coupling: even after spectral truncation, a reduced kernel retains energy at higher modes. The resulting behavior depends systematically on activation class—analytic versus non-analytic (with singular features)—and the predictions are borne out experimentally: the kernel evolution recursion matches measured kernels across depth, and both parallel and perpendicular susceptibilities agree with theory within empirical one-sigma bands.

Beyond early-training predictions, we also show that the criticality theory leads to actionable improvements in practical training. Concretely, our criterion-matched initialization (calibration) produces markedly more stable optimization across random seeds, faster convergence, and lower achieved loss; on a standard PDEBench Burgers benchmark it also improves test performance relative to a vanilla FNO baseline, indicating better downstream generalization.

These findings provide practical criteria for selecting initialization hyperparameters and for anticipating spectral transfer as architectures deepen or widen. Promising next steps are to further develop these criteria into automated initialization/tuning procedures, broaden the analysis beyond the analytic, scale-invariant, and residual cases, track the ensemble distribution through training, and investigate higher-order correlations.

\subsubsection*{Acknowledgments}
Supported by a KIAS Individual Grant (CG102201) at Korea Institute for Advanced Study. This work is supported by the Center for Advanced Computation at Korea Institute for Advanced Study.

\bibliography{iclr2026_conference}
\bibliographystyle{iclr2026_conference}

\appendix

\section{FCN Effective-Theory Background}\label{app:fcn}
We define the fully connected network (FCN), a basic architecture in deep learning models:

\noindent
{\bf Definition 3}{ The FCN is composed as follows:
\begin{equation*}
    \begin{split}
        z^{(1)}_{i}&:=\sum_{j=1}^{n^{(0)}}W^{(1)}_{ij}x_{j}+b^{(1)}_{i} \\
        z^{(l+1)}_{i}&:=\sum_{j=1}^{n^{(l)}}W^{(l+1)}_{ij}\sigma(z^{(l)}_{j})+b^{(l+1)}_{i}, \quad l=1,\dots,L-1.
    \end{split}
\end{equation*}
where $\sigma:\mathbb{R}\rightarrow\mathbb{R}$ is an activation function and $x$ is input-vector, $W^{(l)}$ and $b^{(l)}$ are weights and biases parameters. $z^{(l)}$ denotes the \textbf{pre-activation at the $l$-th layer}. 
}

In deep learning architectures, the learning process involves optimizing the parameters $\{W^{(l)},b^{(l)}\}_{l=1,\dots,L}$. The initial setting of these parameters, known as \textbf{initialization}, follows a specific probability distribution called the \textbf{initialization distribution}. For the statistical analysis of neural network pre-activations, we assume that the initialization distribution consists of independent and identically distributed (i.i.d.) Gaussian random variables. Specifically, we assume:
\begin{equation*}
    \begin{split}
        W_{ij}^{(l)}\sim N\Big(0,\frac{C_{W}}{n^{(l-1)}}\Big) \\
        b_{i}^{(l)} \sim N(0,C_{b}).
    \end{split}
\end{equation*}
Then, the statistics of first-layer pre-activation is also i.i.d Gaussian distribution with following covariances:
\[
\mathbb{E}[z^{(1)}_{i_{1};\alpha_{1}}z^{(1)}_{i_{2};\alpha_{2}}]=\delta_{i_{1}i_{2}}G^{(1)}_{\alpha_{1}\alpha_{2}}
\]
where the indices $\alpha_{j}$ are labels for input data, and $G^{(1)}_{\alpha_{1}\alpha_{2}}$ is a metric for the first pre-activation, which contains all the information about the statistics. (since first pre-activation is Gaussian) 
\[
G^{(1)}_{\alpha_{1}\alpha_{2}}=C^{(1)}_{b}+\frac{C^{(1)}_{W}}{n^{(0)}}\sum_{j}^{n^{(0)}}x_{j;\alpha_{1}}x_{j;\alpha_{2}}.
\]
Now, the distribution of $l+1$-th layer pre-activation conditioned on $l$-th layer is also following Gaussian distribution:
\[
\mathbb{E}[z^{(l+1)}_{i_{1};\alpha_{1}}z^{(l+1)}_{i_{2};\alpha_{2}}|z^{(l)}]=\delta_{i_{1}i_{2}}\hat{G}^{(l)}_{\alpha_{1}\alpha_{2}}
\]
where $\hat{G}^{(l)}_{\alpha_{1}\alpha_{2}}$ itself is a random variable, and defined as follows:
\[
\hat{G}^{(l)}_{\alpha_{1}\alpha_{2}}=C^{(l)}_{b}+\frac{C^{(l)}_{W}}{n^{(l-1)}}\sum_{j}^{n^{(l)}}z^{(l-1)}_{j;\alpha_{1}}z^{(l-1)}_{j;\alpha_{2}}.
\]
Let the fluctuation of $\hat{G}^{(l)}_{\alpha_{1}\alpha_{2}}$ around its mean $G_{\alpha_{1}\alpha_{2}}^{(l)}:=\mathbb{E}[\hat{G}^{(l)}_{\alpha_{1}\alpha_{2}}]$ be $\Delta \hat{G}^{(l)}_{\alpha_{1}\alpha_{2}}:=\hat{G}^{(l)}_{\alpha_{1}\alpha_{2}}-G_{\alpha_{1}\alpha_{2}}^{(l)}$ then it can be shown that the variance of this fluctuation is related to four-point connected correlator and is $\mathcal{O}\Big(\frac{1}{n^{(l-1)}}\Big)$. Specifically, we define four-point vertex as follows:
\[V^{(l)}_{(\alpha_{1}\alpha_{2})(\alpha_{3}\alpha_{4})}:=n^{(l-1)}\mathbb{E}[\Delta \hat{G}^{(l)}_{\alpha_{1}\alpha_{2}}\Delta \hat{G}^{(l)}_{\alpha_{3}\alpha_{4}}].\]
It can be easily checked that $2k$-point connected correlators is $\mathcal{O}\Big(\frac{1}{{n^{(l)}}^{k-1}}\Big)$. So, for large enough widths, we can consider the neural network ensembles as a Gaussian processes. As widths go to infinity, the $2$-point correlations go to some fixed kernel let denote this kernel as $K^{(l)}_{\alpha_{1}\alpha_{2}}$. According to \cite{Roberts22}, for infinite width neural networks, the running of kernels and four-points vertices are described as follows:
\begin{equation}
    \begin{split}
        K_{\alpha_{1}\alpha_{2}}^{(l+1)}&=C_{b}^{(l+1)}+C_{W}^{(l+1)}\langle \sigma_{\alpha_{1}}\sigma_{\alpha_{2}}\rangle_{K^{(l)}}, \\
        V_{(\alpha_{1}\alpha_{2})(\alpha_{3}\alpha_{4})}^{(l+1)}&=(C_{W}^{(l+1)})^{2}[\langle \sigma_{\alpha_{1}}\sigma_{\alpha_{2}}\sigma_{\alpha_{3}}\sigma_{\alpha_{4}}\rangle_{K^{(l)}}-\langle \sigma_{\alpha_{1}}\sigma_{\alpha_{2}}\rangle_{K^{(l)}}-\langle \sigma_{\alpha_{3}}\sigma_{\alpha_{4}}\rangle_{K^{(l)}}]\\
        &+\frac{n^{(l)}}{4n^{(l-1)}}(C_{W}^{(l+1)})^{2}\sum_{\beta_{1},\dots,\beta_{4}\in\mathcal{D}}V_{(l)}^{(\beta_{1}\beta_{2})(\beta_{3}\beta_{4})}\langle \sigma_{\alpha_{1}}\sigma_{\alpha_{2}}(z_{\beta_{1}}z_{\beta_{2}}-K_{\beta_{1}\beta_{2}}^{(l)})\rangle_{K^{(l)}} \\
        &\langle \sigma_{\alpha_{3}}\sigma_{\alpha_{4}}(z_{\beta_{3}}z_{\beta_{4}}-K_{\beta_{3}\beta_{4}}^{(l)})\rangle_{K^{(l)}}+O\Big(\frac{1}{n^{(l)}}\Big)
    \end{split} \label{fcnfour}
\end{equation}
Let’s now examine how the kernel changes under a small perturbation around a reference input $x_{0}$. Define two perturbed inputs $x_{\pm}=x_{0}\pm\frac{1}{2}\delta x$, and let $z_{+}^{(l)}$ and $z_{-}^{(l)}$ denote their pre-activations at layer $l$. For this two-point dataset, the kernel admits an expansion of the form:
\[
K_{\alpha\beta}^{(l)}=\begin{pmatrix}
    K_{++}^{(l)} & K_{+-}^{(l)} \\
    K_{-+}^{(l)} & K_{--}^{(l)}
\end{pmatrix}=K_{00}^{(l)}\begin{pmatrix}
    1 & 1 \\
    1 & 1
\end{pmatrix}+K_{\parallel}^{(l)}\begin{pmatrix}
    1 & 0 \\
    0 & -1
\end{pmatrix}+K_{\perp}^{(l)}\begin{pmatrix}
    1 & -1 \\
    -1 & 1
\end{pmatrix}.
\]
where the coefficients are as follows:
\begin{equation*}
    \begin{split}
        K_{00}^{(l)}&=\mathbb{E}\Big[\frac{1}{n^{(l)}}\sum_{i}^{n^{(l)}}x_{i;0}^{2}\Big], \\
        K_{\parallel}^{(l)}&=\frac{1}{2}\bigg(\mathbb{E}\Big[\frac{1}{n^{(l)}}\sum_{i}^{n^{(l)}}x_{+;0}^{2}\Big]-\mathbb{E}\Big[\frac{1}{n^{(l)}}\sum_{i}^{n^{(l)}}x_{-;0}^{2}\Big]\bigg), \\
        K_{\perp}^{(l)}&=\frac{1}{4}\bigg(\mathbb{E}\Big[\frac{1}{n^{(l)}}\sum_{i}^{n^{(l)}}x_{+;0}^{2}\Big]+\mathbb{E}\Big[\frac{1}{n^{(l)}}\sum_{i}^{n^{(l)}}x_{-;0}^{2}\Big]-2\mathbb{E}\Big[\frac{1}{n^{(l)}}\sum_{i}^{n^{(l)}}x_{+;0}x_{-;0}\Big]\bigg).
    \end{split}
\end{equation*}
As $\delta x\rightarrow 0$, the components $K_{\parallel}^{(l)}$ and $K_{\perp}^{(l)}$ each vanish, while the first term collapses to a degenerate matrix. In \cite{Roberts22}, through a careful eigenvalue expansion one finds that the two–point activation correlation under this perturbed kernel is given by:
\begin{equation*}
    \begin{split}
        &\langle \sigma(z_{\alpha})\sigma(z_{\beta})\rangle_{K^{(l)}} \\
        &=\Big[\sigma(z_{0})\sigma(z_{0})\rangle_{K^{(l)}_{00}}\Big]\gamma_{\alpha\beta}^{[0]} \\
        &+\Big[\Big(\frac{\delta K_{\parallel}^{(l)}}{K_{00}^{(l)}}\Big)\langle z_{0}\sigma'(z_{0})\sigma(z_{0})\rangle_{K^{(l)}_{00}}\Big]\gamma_{\alpha\beta}^{[1]} \\
        &+\Big[\delta\delta K_{\perp}^{(l)}\langle \sigma'(z_{0})\sigma'(z_{0})\rangle_{K^{(l)}_{00}}+\Big(\frac{\delta K_{\parallel}^{(l)}}{2K_{00}^{(l)}}\Big)^{2}\langle (z_{0}^{2}-K_{00}^{(l)})\sigma'(z_{0})\sigma'(z_{0})\rangle_{K^{(l)}_{00}}\Big]\gamma_{\alpha\beta}^{[2]}
    \end{split}
\end{equation*}

From that result, one deduces the following recursion relations for the leading terms of $K_{\parallel}^{(l)}$ and $K_{\perp}^{(l)}$:
\begin{equation*}
    \begin{split}
        \delta K_{\parallel}^{(l+1)}&=\frac{C_{W}}{K}\langle z\sigma'(z)\sigma(z)\rangle_{K}\delta K_{\parallel}^{(l)}, \\
        \delta\delta K_{\perp}^{(l+1)}&=C_{W}\langle \sigma'(z)\sigma'(z)\rangle_{K}\delta\delta K_{\perp}^{(l)}+\frac{C_{W}}{4K^{2}}\langle \sigma'(z)\sigma'(z)(z^{2}-K)\rangle_{K}(\delta K_{\parallel}^{(l)})^{2}.
    \end{split}
\end{equation*}
When the perturbation is orthogonal to the data, the second term on the right‑hand side of the equation for $\delta\delta K_{\perp}^{(l)}$ vanishes. Consequently, in both the parallel and perpendicular cases the two formulas above reduce to geometric‑sequence recursions, and if their common ratios $\frac{C_{W}}{K}\langle z\sigma'(z)\sigma(z)\rangle_{K}$ and $C_{W}\langle \sigma'(z)\sigma'(z)\rangle_{K}$ both equal one, the volume occupied by the data distribution remains constant as it propagates through the layers.
\section{Scaling Law and Four-Point Vertex}\label{app:sec3}
\textbf{Large-Width Scaling of Connected Correlators}
\cref{Conncorrs} presents the general form of an arbitrary $2k$-point connected correlator. Under mild regularity conditions, an inductive estimate shows that the correlator in layer $l$ obeys the scaling $\mathcal{O}(\frac{1}{n^{s}})$. If the activation function is analytic and vanishes at the origin, the leading contribution to the connected correlator of activations in layer $l-1$ obeys the same bound. In addition, the statistics of the pre-kernel integration stages match those of a fully connected network, as established in Section 2.1. Hence, when all widths are identical and large, $n=n_{l}=\dots=n_{0}, n\gg1$,the connected correlator in layer $l$ is suppressed to 
$\mathcal{O}(\frac{1}{n^{k-1}})$. In the infinite width limit, only the two-point connected correlator carries appreciable statistical weight; the analysis that follows therefore concentrates on the relationship between this kernel and the four-point vertex that governs its fluctuations. For conciseness the derivation in \cref{Conncorrs} is given for real-valued functions, but the extension to complex-valued fields is straightforward.
\begin{equation}
\begin{split}
    &\mathbb{E}\bigg[\mathcal{F}\Big(\mathcal{Z}^{(l)}\{\mathbf{u}_{1}\}\Big)_{i_{1}}(f_{1})\dots\mathcal{F}\Big(\mathcal{Z}^{(l)}\{u_{2k}\}\Big)_{i_{2k}}(f_{2k}) \bigg]\bigg|_{\text{conn}} \\
&= \Big(\frac{C_{R^{(l)}}}{n^{(l-1)}}\Big)^{k}\delta_{i_{1}i_{2}}\dots\delta_{i_{2k-1}i_{2k}}\delta(f_{1}-f_{2})\dots\delta(f_{2k-1}-f_{2k})  \\
&\sum_{j_{1},\dots,j_{2k}}^{n^{(l-1)}}\delta_{j_{1}j_{2}}\dots\delta_{j_{2k-1}j_{2k}}\bigg\{ \mathbb{E}
\bigg[\mathcal{F}\Big(\mathfrak{S}^{(l-1)}\{\mathbf{u}_{1}\}\Big)_{j_{1}}(f_{1})\dots\mathcal{F}\Big(\mathfrak{S}^{(l-1)}\{u_{2k}\}\Big)_{j_{2k}}(f_{2k}) \bigg]  \\
&-\sum_{\text{all subdivisions of }(1,\dots,2k)}\mathbb{E}\bigg[\mathcal{F}\Big(\mathfrak{S}^{(l-1)}\{u_{\mu_{1,1}}\}\Big)_{i_{\mu_{1,1}}}\dots\mathcal{F}\Big(\mathfrak{S}^{(l-1)}\{u_{\mu_{1,t_{1}}}\}\Big)_{i_{\mu_{1,t_{1}}}}\bigg]|_{\text{conn}}\dots \\
&\mathbb{E}\bigg[\mathcal{F}\Big(\mathfrak{S}^{(l-1)}\{u_{\mu_{\nu,1}}\}\Big)_{i_{\mu_{\nu,1}}}\dots\mathcal{F}\Big(\mathfrak{S}^{(l-1)}\{u_{\mu_{\nu,t_{\nu}}}\}\Big)_{i_{\mu_{\nu,t_{\nu}}}}\bigg]|_{\text{conn}}\bigg\}
\end{split} \label{Conncorrs}
\end{equation}

\textbf{Four-point vertex}
Henceforth, to obtain a well-defined four-point vertex, we factor out the $\delta(f-f’)$ from the metric and work with the reduced metric. We define the fluctuation of metric $\Delta\widetilde{\mathcal{G}^{(l)}}$ as follows equation:
\begin{equation*}
\begin{split}
       &\Delta\widetilde{\mathcal{G}^{(l)}}\{\mathbf{u},\mathbf{v}\}(f,f'):=\widetilde{\mathcal{G}^{(l)}}(f,f')-\mathcal{G}^{(l)}(f,f') \\
    &=\frac{C_{R^{(l)}}}{n^{(l-1)}}\sum_{j}\bigg(\mathcal{F}\Big(\mathfrak{S}^{(l-1)}\{\mathbf{u}\}\Big)_{j}(f)\overline{\mathcal{F}\Big(\mathfrak{S}^{(l-1)}\{\mathbf{v}\}\Big)_{j}}(f') \\&-\mathbb{E}\bigg[\mathcal{F}\Big(\mathfrak{S}^{(l-1)}\{\mathbf{u}\}\Big)(f)\cdot\overline{\mathcal{F}\Big(\mathfrak{S}^{(l-1)}\{\mathbf{v}\}\Big)}(f')\bigg]\bigg)
\end{split}
\end{equation*}
and define four-point vertex $\mathcal{V}^{(l)}\{(\mathbf{u}_{1},\mathbf{u}_{2}),(\mathbf{u}_{3},\mathbf{u}_{4})\}$ which is scaled variance of fluctuation $\Delta\widetilde{\mathcal{G}^{(l)}}$. Using this quantity, we can calculate four-point correlators perturbatively. \cref{fver} shows the expansion of four-point vertex which is composed of four-point connected correlators of $(l-1)$-th layer activations.
\begingroup
\small
\begin{equation*}
    \begin{split}
        &\frac{1}{n^{(l-1)}}\mathcal{V}^{(l)}\{(\mathbf{u}_{1},\mathbf{u}_{2}),(\mathbf{u}_{3},\mathbf{u}_{4})\}(f_{1},f_{2},f_{3},f_{4}):=\\
        &\mathbb{E}\bigg[\Delta\widetilde{\mathcal{G}^{(l)}}\{\mathbf{u}_{1},\mathbf{u}_{2}\}(f_{1},f_{2})\Delta\widetilde{\mathcal{G}^{(l)}}\{\mathbf{u}_{3},\mathbf{u}_{4}\}(f_{3},f_{4})\bigg] \\
        &=C_{R^{(l)}}^{2}\Big(\frac{1}{n^{(l-1)}}\Big)^{2}\mathbb{E}\bigg[\sum_{j,k}\bigg(\mathcal{F}\Big(\mathfrak{S}^{(l-1)}\{\mathbf{u}_{1}\}\Big)_{j}(f_{1})\mathcal{F}\Big(\mathfrak{S}^{(l-1)}\{\mathbf{u}_{2}\}\Big)_{j}(f_{2}) \\
        &-\mathbb{E}\bigg[\mathcal{F}\Big(\mathfrak{S}^{(l-1)}\{\mathbf{u}_{1}\}\Big)(f_{1})\cdot\mathcal{F}\Big(\mathfrak{S}^{(l-1)}\{\mathbf{u}_{2}\}\Big)(f_{2})\bigg]\bigg) \\
        &\bigg(\mathcal{F}\Big(\mathfrak{S}^{(l-1)}\{\mathbf{u}_{3}\}\Big)_{k}(f_{3})\mathcal{F}\Big(\mathfrak{S}^{(l-1)}\{\mathbf{u}_{4}\}\Big)_{k}(f_{4})-\mathbb{E}\bigg[\mathcal{F}\Big(\mathfrak{S}^{(l-1)}\{\mathbf{u}_{3}\}\Big)(f_{3})\cdot\mathcal{F}\Big(\mathfrak{S}^{(l-1)}\{\mathbf{u}_{4}\}\Big)(f_{4})\bigg]\bigg)\bigg]
    \end{split}
\end{equation*}
\begin{equation}
\begin{split}
        &= C_{R^{(l)}}^{2}\Big(\frac{1}{n^{(l-1)}}\Big)^{2}\sum_{j}\bigg(\mathbb{E}\bigg[\mathcal{F}\Big(\mathfrak{S}^{(l-1)}\{\mathbf{u}_{1}\}\Big)_{j}(f_{1})\mathcal{F}\Big(\mathfrak{S}^{(l-1)}\{\mathbf{u}_{2}\}\Big)_{j}(f_{2})\\&
        \mathcal{F}
        \Big(\mathfrak{S}^{(l-1)}\{\mathbf{u}_{3}\}\Big)_{j}(f_{3})\mathcal{F}\Big(\mathfrak{S}^{(l-1)}\{\mathbf{u}_{4}\}\Big)_{j}(f_{4})\bigg]-\mathbb{E}\bigg[\mathcal{F}\Big(\mathfrak{S}^{(l-1)}\{\mathbf{u}_{1}\}\Big)_{j}(f_{1})\mathcal{F}\Big(\mathfrak{S}^{(l-1)}\{\mathbf{u}_{2}\}\Big)_{j}(f_{2})\bigg] \\&\mathbb{E}\bigg[\mathcal{F}\Big(\mathfrak{S}^{(l-1)}\{\mathbf{u}_{3}\}\Big)_{j}(f_{3})\mathcal{F}\Big(\mathfrak{S}^{(l-1)}\{\mathbf{u}_{4}\}\Big)_{j}(f_{4})\bigg]\bigg) \\&+C_{R^{(l)}}^{2}\Big(\frac{1}{n^{(l-1)}}\Big)^{2}\sum_{j\neq k}\bigg(\mathbb{E}\bigg[\mathcal{F}\Big(\mathfrak{S}^{(l-1)}\{\mathbf{u}_{1}\}\Big)_{j}(f_{1})\mathcal{F}\Big(\mathfrak{S}^{(l-1)}\{\mathbf{u}_{2}\}\Big)_{j}
        (f_{2})\\&\mathcal{F}\Big(\mathfrak{S}^{(l-1)}\{\mathbf{u}_{3}\}\Big)_{k}(f_{3})\mathcal{F}\Big(\mathfrak{S}^{(l-1)}\{\mathbf{u}_{4}\}\Big)_{k}(f_{4})\bigg]-\mathbb{E}\bigg[\mathcal{F}\Big(\mathfrak{S}^{(l-1)}\{\mathbf{u}_{1}\}\Big)_{j}(f_{1})\mathcal{F}\Big(\mathfrak{S}^{(l-1)}\{\mathbf{u}_{2}\}\Big)_{j}(f_{2})\bigg]\\&\mathbb{E}\bigg[\mathcal{F}\Big(\mathfrak{S}^{(l-1)}\{\mathbf{u}_{3}\}\Big)_{k}(f_{3})\mathcal{F}\Big(\mathfrak{S}^{(l-1)}\{\mathbf{u}_{4}\}\Big)_{k}(f_{4})\bigg]\bigg)  \label{fver}
    \end{split}
\end{equation}
\endgroup

Next, we describe the recursion for the four-point vertex. In \cref{fver}, when all four indices coincide, the distribution reduces to one over a single scalar. Therefore, by computing each statistic similar to as in the nearly Gaussian action expansion of \cite{Roberts22}, we obtain the following expression as in \cref{fcnfour}:
\begingroup
\small
\begin{equation}
    \begin{split}
        &\frac{C_{R^{(l)}}^{2}}{n^{(l-1)}}\mathcal{V}^{(l)}\{(\mathbf{u}_{\alpha_{1}},\mathbf{u}_{\alpha_{2}}),(\mathbf{u}_{\alpha_{3}},\mathbf{u}_{\alpha_{4}})\}\\
        &=\frac{1}{n^{(l)}}\Big[\Big\langle\mathcal{F}\Big(\mathfrak{S}^{(l-1)}\Big)\{\mathbf{u}_{\alpha_{1}}\} \mathcal{F}\Big(\mathfrak{S}^{(l-1)}\Big)\{\mathbf{u}_{\alpha_{2}}\}\mathcal{F}\Big(\mathfrak{S}^{(l-1)}\Big)\{\mathbf{u}_{\alpha_{3}}\}\mathcal{F}\Big(\mathfrak{S}^{(l-1)}\Big)\{\mathbf{u}_{\alpha_{4}}\}\Big\rangle\Big] \\
        &-\Big\langle\mathcal{F}\Big(\mathfrak{S}^{(l-1)}\Big)\{\mathbf{u}_{\alpha_{1}}\}\mathcal{F}\Big(\mathfrak{S}^{(l-1)}\Big)\{\mathbf{u}_{\alpha_{2}}\}\Big\rangle\Big\langle\mathcal{F}\Big(\mathfrak{S}^{(l-1)}\Big)\{\mathbf{u}_{\alpha_{3}}\}\mathcal{F}\Big(\mathfrak{S}^{(l-1)}\Big)\{\mathbf{u}_{\alpha_{4}}\}\Big\rangle\Big] \\
        &+\frac{1}{4n^{(l-1)}}\sum_{\beta_{1},\dots,\beta_{4}
        \in\mathcal{D}}{\mathcal{V}_{(l-1)}}^{-1}\{(\mathbf{u}_{\beta_{1}},\mathbf{u}_{\beta_{2}}),(\mathbf{u}_{\beta_{3}},\mathbf{u}_{\beta_{4}})\}\\&\Big\langle\Big(\mathcal{F}\Big(\mathfrak{S}^{(l-1)}\Big)\{\mathbf{u}_{\alpha_{1}}\}\mathcal{F}\Big(\mathfrak{S}^{(l-1)}\Big)\{\mathbf{u}_{\alpha_{2}}\}\Big(\mathcal{F}(\mathcal{Z}^{(l-1)}\{\mathbf{u}_{\beta_{1}}\})\mathcal{F}(\mathcal{Z}^{(l-1)}\{\mathbf{u}_{\beta_{2}}\})-\mathcal{G}^{(l-1)}\{\mathbf{u}_{\beta_{1}},\mathbf{u}_{\beta_{2}}\}\Big)\Big)\Big\rangle_{\mathcal{G}^{(l-1)}} \\
        &\Big\langle\Big(\mathcal{F}\Big(\mathfrak{S}^{(l-1)}\Big)\{\mathbf{u}_{\alpha_{3}}\}\mathcal{F}\Big(\mathfrak{S}^{(l-1)}\Big)\{\mathbf{u}_{\alpha_{4}}\}\Big(\mathcal{F}(\mathcal{Z}^{(l-1)}\{\mathbf{u}_{\beta_{1}}\})\mathcal{F}(\mathcal{Z}^{(l-1)}\{\mathbf{u}_{\beta_{2}}\})-\mathcal{G}^{(l-1)}\{\mathbf{u}_{\beta_{1}},\mathbf{u}_{\beta_{2}}\}\Big)\Big)\Big\rangle_{\mathcal{G}^{(l-1)}} \\
        &+\mathcal{O}\Big(\frac{1}{n^{2}}\Big)
    \end{split} \label{four}
\end{equation}
\endgroup
We use angle brackets $\langle\cdot\rangle$ to indicate averaging with respect to the kernel (i.e., expectation under the kernel induced measure) in \cref{four}; this notation should not be confused with an inner product.

\section{Susceptibility Recursions and Criticality Conditions} \label{app:susc}

As in Appendix~\ref{app:fcn} for FCNs, we derive the criticality condition by writing layerwise recursions for perturbations taken parallel and perpendicular to the data manifold. 
For the parallel case, fix a reference pre-activation $\mathbf{Z}_0$ and consider two nearby samples 
$\mathbf{Z}^{(l)}_{\pm}=(1\pm\epsilon)\,\mathbf{Z}_0$, where $\epsilon>0$ is the perturbation amplitude at layer $l$. 
Let $\mathbf{u}_0:=\mathcal{F}(\mathbf{Z}_0)$. 
Then the difference between 
$\big\|\mathcal{F}(\mathbf{Z}^{(l)}_{+})\big\|_{\mathcal{K}^{(l)}\{\mathbf{u}_0\}}$ and 
$\big\|\mathcal{F}(\mathbf{Z}^{(l)}_{-})\big\|_{\mathcal{K}^{(l)}\{\mathbf{u}_0\}}$ 
is $4\,\mathcal{K}^{(l)}\{\mathbf{u}_0\}\,\epsilon$. 
Performing a first–order Taylor expansion to evaluate the difference at layer $l\!+\!1$ yields:
\begin{equation*}
\begin{split}
        &\|\mathcal{F}(\mathcal{Z}^{(l+1)}|_{\mathbf{\mathcal{Z}}_{0}(1+\epsilon)})\|_{\mathcal{K}^{(l)}\{u_{0}\}}=C_{R^{(l+1)}}\|\mathcal{F}(\mathfrak{S}^{(l)}|_{\mathbf{\mathcal{Z}}_{0}(1+\epsilon)}) \|_{\mathcal{K}^{(l)}\{u_{0}\}} \\
    &=C_{R^{(l+1)}}\|\mathcal{F}(\mathfrak{S}^{(l)}|_{\mathbf{\mathcal{Z}}_{0}}) +\mathcal{F}(\mathbf{\mathbf{\mathcal{Z}}_{0}}\mathfrak{S}'^{(l)}|_{\mathbf{\mathcal{Z}}_{0}} )\epsilon+\mathcal{O}(\epsilon^{2})\|_{\mathcal{K}^{(l)}\{u_{0}\}} \\
    &\Rightarrow \big\|\mathcal{F}(\mathbf{Z}^{(l+1)}_{+})\big\|_{\mathcal{K}^{(l)}\{\mathbf{u}_0\}}-\big\|\mathcal{F}(\mathbf{Z}^{(l+1)}_{-})\big\|_{\mathcal{K}^{(l)}\{\mathbf{u}_0\}} \\
    &=2C_{R^{(l+1)}}\Big(\langle \mathcal{F}(\mathbf{\mathbf{\mathcal{Z}}_{0}}\mathfrak{S}'^{(l)}|_{\mathbf{\mathcal{Z}}_{0}} ) ,\mathcal{F}(\mathfrak{S}^{(l)}|_{\mathbf{\mathcal{Z}}_{0}}) \{\mathbf{u_{0}}\}\rangle_{\mathcal{K}^{(l)}\{u_{0}\}}+\langle\mathcal{F}(\mathfrak{S}^{(l)}|_{\mathbf{\mathcal{Z}}_{0}}) ,\mathcal{F}(\mathbf{\mathbf{\mathcal{Z}}_{0}}\mathfrak{S}'^{(l)}|_{\mathbf{\mathcal{Z}}_{0}} )\rangle_{\mathcal{K}^{(l)}\{u_{0}\}}\Big)\epsilon.
\end{split}
\end{equation*}

\noindent
Therefore, for the difference in variances under parallel perturbations between the $l$-th and $(l+1)$-th layers to remain unchanged, the following equation must be satisfied:
\begingroup
\small
\begin{equation*}
\begin{split}
        &\chi_{\parallel}(f,f'):=\frac{ \big\|\mathcal{F}(\mathbf{Z}^{(l+1)}_{+})\big\|_{\mathcal{K}^{(l)}\{\mathbf{u}_0\}}-\big\|\mathcal{F}(\mathbf{Z}^{(l+1)}_{-})\big\|_{\mathcal{K}^{(l)}\{\mathbf{u}_0\}}}{\big\|\mathcal{F}(\mathbf{Z}^{(l)}_{+})\big\|_{\mathcal{K}^{(l)}\{\mathbf{u}_0\}}-\big\|\mathcal{F}(\mathbf{Z}^{(l)}_{-})\big\|_{\mathcal{K}^{(l)}\{\mathbf{u}_0\}}} \\
        &=\frac{2C_{R^{(l+1)}}\Big(\langle \mathcal{F}(\mathbf{\mathbf{\mathcal{Z}}_{0}}\mathfrak{S}'^{(l)}|_{\mathbf{\mathcal{Z}}_{0}} ) ,\mathcal{F}(\mathfrak{S}^{(l)}|_{\mathbf{\mathcal{Z}}_{0}}) \{\mathbf{u_{0}}\}\rangle_{\mathcal{K}^{(l)}\{u_{0}\}}+\langle\mathcal{F}(\mathfrak{S}^{(l)}|_{\mathbf{\mathcal{Z}}_{0}}) ,\mathcal{F}(\mathbf{\mathbf{\mathcal{Z}}_{0}}\mathfrak{S}'^{(l)}|_{\mathbf{\mathcal{Z}}_{0}} )\rangle_{\mathcal{K}^{(l)}\{u_{0}\}}\Big)\epsilon}{4\mathcal{K}^{(l)}\{\mathbf{\mathbf{u_0}}\}\epsilon} \\
        &=\frac{C_{R^{(l+1)}}\Big(\langle \mathcal{F}(\mathbf{\mathbf{\mathcal{Z}}_{0}}\mathfrak{S}'^{(l)}|_{\mathbf{\mathcal{Z}}_{0}} ) ,\mathcal{F}(\mathfrak{S}^{(l)}|_{\mathbf{\mathcal{Z}}_{0}}) \{\mathbf{u_{0}}\}\rangle_{\mathcal{K}^{(l)}\{u_{0}\}}+\langle\mathcal{F}(\mathfrak{S}^{(l)}|_{\mathbf{\mathcal{Z}}_{0}}) ,\mathcal{F}(\mathbf{\mathbf{\mathcal{Z}}_{0}}\mathfrak{S}'^{(l)}|_{\mathbf{\mathcal{Z}}_{0}} )\rangle_{\mathcal{K}^{(l)}\{u_{0}\}}\Big)(f,f')}{2\mathcal{K}^{(l)}\{\mathbf{\mathbf{u_0}}\}(f,f')} \\&= 1.
\end{split}
\end{equation*}
\endgroup
\noindent
Next, we consider perturbations in the perpendicular direction. In principle, the kernel on the two–point set $\{\mathbf{\mathcal{Z}}_{+},\mathbf{\mathcal{Z}}_{-}\}$, with $\mathbf{\mathcal{Z}}_{+}=\mathbf{\mathcal{Z}}_{0}+\delta\boldsymbol{\eta}$ and $\mathbf{\mathcal{Z}}_{-}=\mathbf{\mathcal{Z}}_{0}-\delta\boldsymbol{\eta}$, should be treated as a $2\times2$ matrix. 
However, under the orthogonality assumption $\delta\boldsymbol{\eta}\perp \mathbf{\mathcal{Z}}_{0}$, all mixed terms proportional to $\delta\boldsymbol{\eta}\cdot\mathbf{\mathcal{Z}}_{0}$ vanish, and the joint law of the norms factorizes as
\begin{equation*}
    \mathcal{P}(f)\;\propto\;\exp\!\left(
    -\frac{\|\mathbf{\mathcal{Z}}_{0}\|^{2}(f)}{2\,\mathcal{K}_{0}(f)}
    -\frac{\|\delta\boldsymbol{\eta}\|^{2}(f)}{2\,\mathcal{K}_{\eta}(f)}
    \right).
    \label{perpdis}
\end{equation*}
Therefore, the variance of the difference $\mathbf{\mathfrak{S}}^{(l)}\{u_{+}\}-\mathbf{\mathfrak{S}}^{(l)}\{u_{-}\}$ admits the simple form below, where $\mathbf{u}_{\pm}:=\mathcal{F}(\mathbf{\mathcal{Z}}_{\pm})$.

\begin{equation}
    \begin{split}
        &\frac{\|\mathfrak{S}^{(l)}\{u_{+}\}-\mathfrak{S}^{(l)}\{u_{-}\}\|_{\mathcal{K}^{(l)}}}{4} \\
        &=\frac{\mathcal{K}^{(l+1)}\{u_{+}\}(f)+\mathcal{K}^{(l+1)}\{u_{-}\}(f)-\mathcal{K}^{(l+1)}\{u_{+},u_{-}\}(f)-\mathcal{K}^{(l+1)}\{u_{-},u_{+}\}(f)}{4} \\
        &=\frac{\langle\mathcal{F}(\mathfrak{S}^{(l)}\{u_{0}\})+ \mathcal{F}(\mathfrak{S}'^{(l)}\{u_{0}\}\delta\boldsymbol{\eta})+\mathcal{O}(\delta\boldsymbol{\eta}^{2}),\mathcal{F}(\mathfrak{S}^{(l)}\{u_{0}\})+ \mathcal{F}(\mathfrak{S}'^{(l)}\{u_{0}\}\delta\boldsymbol{\eta})+\mathcal{O}(\delta\boldsymbol{\eta}^{2})\rangle_{\mathcal{K}^{(l)}}}{4} \\
        &+\frac{\langle\mathcal{F}(\mathfrak{S}^{(l)}\{u_{0}\})-\mathcal{F}(\mathfrak{S}'^{(l)}\{u_{0}\}\delta\boldsymbol{\eta})+\mathcal{O}(\delta\boldsymbol{\eta}^{2}),\mathcal{F}(\mathfrak{S}^{(l)}\{u_{0}\})-\mathcal{F}(\mathfrak{S}'^{(l)}\{u_{0}\}\delta\boldsymbol{\eta})+\mathcal{O}
        (\delta\boldsymbol{\eta}^{2})\rangle_{\mathcal{K}^{(l)}}}{4} \\
        &-\frac{\langle\mathcal{F}(\mathfrak{S}^{(l)}\{u_{0}\})-\mathcal{F}(\mathfrak{S}'^{(l)}\{u_{0}\}\delta\boldsymbol{\eta})+\mathcal{O}(\delta\boldsymbol{\eta}^{2}),\mathcal{F}(\mathfrak{S}^{(l)}\{u_{0}\})+ \mathcal{F}(\mathfrak{S}'^{(l)}\{u_{0}\}\delta\boldsymbol{\eta})+\mathcal{O}(\delta\boldsymbol{\eta}^{2})\rangle_{\mathcal{K}^{(l)}}}{4} \\
        &-\frac{\langle\mathcal{F}(\mathfrak{S}^{(l)}\{u_{0}\})+\mathcal{F}(\mathfrak{S}'^{(l)}\{u_{0}\}\delta\boldsymbol{\eta})+\mathcal{O}(\delta\boldsymbol{\eta}^{2}),\mathcal{F}(\mathfrak{S}^{(l)}\{u_{0}\})- \mathcal{F}(\mathfrak{S}'^{(l)}\{u_{0}\}\delta\boldsymbol{\eta})+\mathcal{O}(\delta\boldsymbol{\eta}^{2})\rangle_{\mathcal{K}^{(l)}}}{4}\\
        &=\Big\|\mathcal{F}\Big(\mathfrak{S}'^{(l)}\{u_{0}\}\Big)*\widehat{\delta\boldsymbol{\eta}}\Big\|_{\mathcal{K}^{(l)}}+\mathcal{O}(|\delta\boldsymbol{\eta}|^{3}).\label{perpsus}
    \end{split} 
\end{equation}

Here, the kernel $\mathcal{K}^{(l)}$ in \cref{perpsus} is defined on the joint space that includes the distribution of $\delta\boldsymbol{\eta}$; Since the variance of $\mathbf{u_{+}}-\mathbf{u_{-}}$ at the $l$-th layer is $4\mathcal{K}_{\eta}$ and the variance of the pre-activation at the $l+1$-th layer is derived in \cref{perpsus} the variance of the difference between the two data points will remain constant across layers only if the corresponding condition is satisfied:
\begin{equation*}
\begin{split}
        \chi_{\perp}(f,f')&:=\frac{ \|\mathcal{Z}^{(l+1)}|_{\mathbf{\mathcal{Z}_0}+\mathbf{\delta\boldsymbol{\eta}}}-\mathcal{Z}^{(l+1)}|_{\mathbf{\mathcal{Z}_0}-\mathbf{\delta\boldsymbol{\eta}}}\|_{\mathcal{K}^{(l)}}}{\|\mathcal{F}(\mathbf{\mathcal{Z}_0}+\mathbf{\delta\boldsymbol{\eta}})-\mathcal{F}(\mathbf{\mathcal{Z}_0}-\mathbf{\delta\boldsymbol{\eta}})\|_{\mathcal{K}_{\eta}}} \\
        &=\delta(f-f')C_{R^{(l+1)}}\frac{\Big\|\mathcal{F}\Big(\mathfrak{S}'^{(l)}\{u_{0}\}\Big)*\widehat{\delta\boldsymbol{\eta}}\Big\|_{\mathcal{K}^{(l)}}(f,f')}{\mathcal{K}_{\eta}(f,f')}=1.
\end{split}
\end{equation*}

\section{Proofs for Theorem 1}\label{app:thm1}

 To handle the non-linearity introduced by an analytic activation in Fourier space, we first state the following lemma.
 
\noindent
{\bf Lemma 1} { Let $\sigma:\mathbb{R}\rightarrow\mathbb{R}$ be an analytic function passing through the origin and $g\in L^{1}$ such that $\sigma(g)\in L^{1}$ then, 
\begin{equation*}
\begin{split}
        \text{For } \sigma(x)=\sum_{n=1}^{\infty}\frac{\sigma_{n}}{n!}x^{n},  \\
        \mathcal{F}(\sigma(g))(f)=\sum_{n=1}^{\infty}\frac{\sigma_{n}}{n!}\hat{g}^{*n}(f).
\end{split}
\end{equation*}
where $\hat{g}^{*n}$ means $n$-times self-convolution of $\hat{g}$.
}

Then, the kernel can be written as follows:
\begin{equation*}
    \begin{split}
        &\mathcal{K}^{(l+1)}\{\mathbf{u}\}(f)=C_{R^{(l+1)}}(f)\int\Big\|\mathcal{F}\Big(\sigma(k^{(l)}*u)\Big)\Big\|^{2}(f)e^{-\int\frac{1}{2\mathcal{K}^{(l)}\{\mathbf{u}\}(f)}\hat{u}(f)^{2}df}\mathcal{D}u \\
        &=\int\sum_{i}\frac{1}{n^{(l)}}\bigg\|\sigma_{1}R^{(l)}_{ij}(f)\hat{u}_{j}(f)+\frac{\sigma_{2}}{2!}\Big((R^{(l)}_{ij}\hat{u}_{j})*(R^{(l)}_{ij'}\hat{u}_{j'})\Big)(f)
    +\cdots\bigg\|^{2}e^{-\int\frac{1}{2\mathcal{K}^{(l)}\{\mathbf{u}\}(f)}\hat{u}(f)^{2}df}\mathcal{D}u
    \end{split}
\end{equation*}

When the network output is expanded analytically and its Fourier transform is taken, the result can be written as a sum of terms in which a Gaussian random field is composed with itself multiple times. Evaluating those terms requires Lemma 2. The lemma is stated for complex, scalar-valued fields, but the vector-valued case follows immediately: simply apply the lemma componentwise, using the distributive property to handle each vector entry separately.

\noindent
{\bf Lemma 2} { Suppose independent, mean-zero Gaussian random fields $\{R(f)\}_{f\in \mathbb{R}}, \{U(f)\}_{f\in \mathbb{R}}$ have following two-point correlators:
\begin{equation*}
    \begin{split}
        \mathbb{E}[R(f)\overline{R(f')}]=\delta(f-f')C(f), \\
        \mathbb{E}[\text{Re}(R(f))\text{Im}(R(f'))]=0, \\
        \mathbb{E}[U(f)U(f')]=K(f,f').
    \end{split}
\end{equation*}
then we have the following relations:

\begin{equation*}
    \begin{split}
        &\mathbb{E}[(RU)^{*n}(f)(\overline{RU})^{*m}(f')] \\
        &=\delta_{n,m}(n!)^{2}\delta(f-f')H^{*n}(f').
    \end{split}
\end{equation*}

where $H(f)=C(f)K(f,f)$.
} 

\begin{proof}
\begin{equation*}
    \begin{split}
        &\mathbb{E}[(RU)^{*n}(f)(\overline{RU})^{*m}(f')]  \\
        &=\mathbb{E}[\int\delta(f-\sum_{k=1}^{n}f^{(k)})\prod_{k=1}^{n}(RU)(f^{(k)})df^{(k)}\int\delta(f'-\sum_{k=1}^{m}f'^{(k)})\prod_{k=1}^{m}(\overline{RU})(f'^{(k)})df^{(m)}] \\
        &=\int\delta(f-\sum_{k=1}^{n}f^{(k)})\delta(f'-\sum_{k=1}^{m}f'^{(k)})\\&\mathbb{E}[\prod_{k=1}^{n}\prod_{l=1}^{m}R(f^{(k)})\overline{R}(f^{(l)})]\mathbb{E}[\prod_{k=1}^{n}\prod_{l=1}^{m}U(f^{(k)})\overline{U}(f^{(l)})]\prod_{k=1}^{n}\prod_{l=1}^{n}df^{(k)}df'^{(l)} 
    \end{split}
\end{equation*}

if $n\neq m$, the expectation of $\prod_{l=1}^{m}R(f^{(k)})\overline{R}(f^{(l)})$ is zero since the wick contraction of each term contains expectation of $R^{2}$ or $\overline{R}^{2}$. So assuming $n=m$ then,

\begin{equation*}
    \begin{split}
        &\mathbb{E}[(RU)^{*n}(f)(\overline{RU})^{*m}(f')]  \\
        &=\delta_{n,m}\int\delta(f-\sum_{k=1}^{n}f^{(k)})\delta(f'-\sum_{k=1}^{n}f'^{(k)})\\&\mathbb{E}[\prod_{k=1}^{n}\prod_{l=1}^{n}R(f^{(k)})\overline{R}(f'^{(l)})]\mathbb{E}[\prod_{k=1}^{n}\prod_{l=1}^{n}U(f^{(k)})\overline{U}(f'^{(l)})]\prod_{k=1}^{n}\prod_{l=1}^{n}df^{(k)}df'^{(l)} \\
        &=\delta_{n,m}\int\delta(f-\sum_{k=1}^{n}f^{(k)})\delta(f'-\sum_{k=1}^{n}f'^{(k)})(n!)^{2}\\&\Big(\prod_{k=1}^{n} \delta(f^{(k)}-f'^{(k)})C(f^{(k)})\Big)\Big(\prod_{k=1}^{n}K(f^{(k)},f'^{(k)})\Big) \prod_{k=1}^{n}df^{(k)}df'^{(k)} \\
        &=\delta_{n,m}(n!)^{2}\int\delta(f-\sum_{k=1}^{n}f^{(k)})\delta(f'-\sum_{k=1}^{n}f'^{(k)})\\&\Big(\prod_{k=1}^{n} \delta(f^{(k)}-f'^{(k)})C(f^{(k)})\Big)\Big(\prod_{k=1}^{n}K(f^{(k)},f'^{(k)})\Big) \prod_{k=1}^{n}df^{(k)}df'^{(k)} \\
        &=\delta_{n,m}(n!)^{2}\int\delta(f-\sum_{k=1}^{n}f^{(k)})\delta(f'-\sum_{k=1}^{n}f^{(k)})\Big(\prod_{k=1}^{n} H(f^{(k)})\Big) \prod_{k=1}^{n}df^{(k)} \\
        &=\delta_{n,m}(n!)^{2}\delta(f-f')H^{*n}(f').
    \end{split}
\end{equation*}    
\end{proof}

Using Lemma 2, we can now express the kernel of the next layer in terms of the current layer’s kernel through the following expansion:
\begin{equation*}
    \begin{split}
        &\mathcal{K}^{(l+1)}(f,f')=\sum_{i}\frac{C_{R^{(l+1)}}(f)}{n^{(l+1)}}\bigg\|\bigg(\sigma_{1}R^{(l)}_{ij}(f)\hat{u}_{j}(f)+\frac{\sigma_{2}}{2!}\Big((R^{(l)}_{ij}\hat{u}_{j})*(R^{(l)}_{ij'}\hat{u}_{j'})\Big)(f)
    +\cdots\bigg)\bigg\|_{\mathcal{K}^{(l)}} \\
    &=C_{R^{(l+1)}}(f)\sum_{i}\frac{1}{n^{(l)}}\sum_{k=1}^{\infty}\frac{\sigma_{k}^{2}}{(k!)^{2}}\|(R^{(l)}_{ij}(f)\hat{u}_{j}(f))^{*k}\|_{\mathcal{K}^{(l)}}\\
    &=\delta(f-f')C_{R^{(l+1)}}(f)\sum_{k=1}^{\infty}\frac{\sigma_{k}^{2}}{(n^{(l)})^{k-1}}\sum_{n_{k,1}+\dots+n_{k,l}=k}\frac{(n_{k,1})!\dots(n_{k,l})!}{k!}(\mathcal{H}^{(l)})^{*n_{k,1}}\cdots(\mathcal{H}^{(l)})^{*n_{k,l}}(f)
    \end{split}
\end{equation*}
In the second line, every term that cannot be paired with its complex conjugate drops out by Lemma 2. Combinatorially, at order $k$, the number of surviving monomials associated with a multiplicity vector $(n_{k,1},\dots,n_{k,l})$ is $\frac{k!}{n_{k,1}!\dots n_{k,l}!}$. Applying Lemma 2 once more, each pair of identical factors contributes an additional factor $(n_{k,j}!)^{2}$. Collecting these factors for all indices produces the coefficients $\frac{(n_{k,1})!\dots(n_{k,l})!}{k!}$ which yields the expression shown in the third line.
And for parallel susceptibility, we get the following:

\begin{equation*}
    \begin{split}
        &\chi_{\parallel}(f,f')=\\
        &\frac{C_{R^{(l+1)}}\Big(\langle \mathcal{F}(\mathbf{\mathbf{\mathcal{Z}}_{0}}\mathfrak{S}'^{(l)}|_{\mathbf{\mathcal{Z}}_{0}} ) ,\mathcal{F}(\mathfrak{S}^{(l)}|_{\mathbf{\mathcal{Z}}_{0}}) \{\mathbf{u_{0}}\}\rangle_{\mathcal{K}^{(l)}\{u_{0}\}}+\langle\mathcal{F}(\mathfrak{S}^{(l)}|_{\mathbf{\mathcal{Z}}_{0}}) ,\mathcal{F}(\mathbf{\mathbf{\mathcal{Z}}_{0}}\mathfrak{S}'^{(l)}|_{\mathbf{\mathcal{Z}}_{0}} )\rangle_{\mathcal{K}^{(l)}\{u_{0}\}}\Big)}{2\mathcal{K}^{(l)}\{\mathbf{u_{0}}\}}  \\
        &=C_{R^{(l+1)}}\frac{\sum_{i}\frac{1}{n^{(l)}}\bigg\langle \sum_{k=1}\frac{\sigma_{k}}{(k-1)!}\Big(R^{(l)}_{ij}(f)\hat{u}_{i}(f)\Big)^{*k}, \sum_{k=1}\frac{\sigma_{k}}{(k)!}\Big(R^{(l)}_{ij}(f)\hat{u}_{i}(f)\Big)^{*k}\bigg\rangle_{\mathcal{K}^{(l)}}}{\mathcal{K}^{(l)}\{\mathbf{u_{0}}\}}  
    \end{split}
\end{equation*}
\begin{equation*}
\begin{split}
        &=C_{R^{(l+1)}}\sum_{i}\frac{1}{n^{(l)}}\sum_{k=1}^{\infty} \frac{\sigma_{k}^2}{(k-1)!k!}\frac{\langle (R^{(l)}_{ij}(f)\hat{u}_{j}(f))^{*k},(R^{(l)}_{ij}(f)\hat{u}_{j}(f))^{*k}\rangle_{\mathcal{K}^{(l)}}}{\mathcal{K}^{(l)}\{\mathbf{u_{0}}\}}. \\
        &=\delta(f-f')C_{R^{(l+1)}}\sum_{k=1}^{\infty} \frac{\sigma_{k}^{2}}{(n^{(l)})^{k-1}}\sum_{n_{k,1}+\dots+n_{k,l}=k}\frac{(n_{k,1})!\dots(n_{k,l})!}{(k-1)!}\frac{(\mathcal{H}^{(l)})^{*n_{k,1}}\cdots(\mathcal{H}^{(l)})^{*n_{k,l}}(f)}{\mathcal{K}^{(l)}\{\mathbf{u_0}\}(f)}.
    \end{split}
\end{equation*}

and for perpendicular susceptibility, we get the following:

\begin{equation*}
    \begin{split}
        &\tilde{\chi}_{\perp}(f,f')=\| \mathcal{F}(\mathfrak{S}'^{(l)}\{\mathbf{u_{0}}\})\|_{\mathcal{K}^{(l)}\{u_{0}\}}(f,f') \\
        &=\sum_{i}\frac{1}{n^{(l)}}\|\sigma_{1}+\sigma_{2}R^{(l)}_{ij}(f)\hat{u}_{j}(f)+\frac{\sigma_{3}}{2!}\Big((R^{(l)}_{ij}\hat{u}_{j})*(R^{(l)}_{ij'}\hat{u}_{j'})\Big)(f)+\cdots\|_{\mathcal{K}^{(l)}\{u_{0}\}}\\
        &=\delta(f-f')\sum_{k=1}^{\infty}\frac{\sigma_{k}^{2}}{(n^{(l)})^{k-1}}\sum_{n_{k,1}+\dots+n_{k,l}=k-1}\frac{(n_{k,1})!\dots(n_{k,l})!}{(k-1)!}(\mathcal{H}^{(l)})^{*n_{k,1}}\cdots(\mathcal{H}^{(l)})^{*n_{k,l}}(f).
    \end{split}
\end{equation*}

\section{Proofs for Theorem 2}\label{app:thm2}

Since the sequence of functions $\sigma_{n}(x)=\frac{x}{2}(1+\text{erf}(nx))$ converges to $\max(x,0)$, we have the following approximation:

\noindent
{\bf Lemma 3} { Let $\text{ReLU}(x):=\max(x,0)$ then we have the following convergent function sequence.
\begin{equation*}
    \text{ReLU}(x)\simeq \sigma_{n}(x)=\frac{1}{2}x+\frac{n}{\sqrt{\pi}}x^{2}-\frac{n^3}{3\sqrt{\pi}}x^{4}+\cdots
\end{equation*}

} 

And to approximate the correlation of $\text{ReLU}$ via $\sigma_{n}$'s, we need following lemma:

\noindent
{\bf Lemma 4} { Suppose $g_{n}$ converges pointwise to $g$ and $h\in L^{1}\cap L^{\infty}$, satisfying following conditions:
\begin{itemize}
    \item $g_{n}(x)\leq C|x|$ for some constant $C>0$.
\end{itemize}
then we have the following:
\begin{equation*}
    \mathcal{F}(g_{n}\circ h) \rightarrow \mathcal{F}(g\circ h) \quad \text{ (uniformly)}
\end{equation*}
}

\begin{proof}
\begin{equation*}
    |g_{n}\circ h(x)|\leq C|h(x)|.
\end{equation*}
So, $\{g_{n}(h)\}$ is dominated by $C|h(x)|$. By dominated convergence theorem we get the following:
\begin{equation*}
    \lim_{n\rightarrow \infty}\int |g_{n}\circ h(x)-g\circ h(x)|dx =0.
\end{equation*}
then, from the following inequality we get the uniform convergence of $g_{n}\circ h$ to $g(h)$:
\begin{equation*}
    |\mathcal{F}(g_{n}\circ h)-\mathcal{F}(g\circ h)|\leq\int |(g_{n}\circ h-g\circ h)e^{-ifx}|dx =\int |g_{n}\circ h-g
    \circ h|dx  = \|g_{n}\circ h-g\circ h\|_{L^{1}}.
\end{equation*}
\end{proof}

\noindent
{\bf Lemma 5} { Suppose mean zero Gaussian random fields $\{R(f)\}_{f\in \mathbb{R}}, \{U(f)\}_{f\in \mathbb{R}}$ have following two-point correlators:
\begin{equation*}
    \begin{split}
        \mathbb{E}[R(f)R(f')]&=\delta(f-f')C(f), \\
        \mathbb{E}[U(f)U(f')]&=K(f,f').
    \end{split}
\end{equation*}
then we have the following relations:
\begin{equation*}
\begin{split}
    \mathbb{E}\Big[\Big\|\text{ReLU}\Big(\int UdR\Big)\Big\|^{2}\Big]&=\frac{1}{2}\int C(f)K(f)df. 
\end{split}
\end{equation*}

}
\begin{proof}

By Ito isometry we have the following identity:

\begin{equation}
\mathbb{E}\Big[\Big\|\Big(\int UdR\Big)\Big\|^{2}\Big]=\int C(f)K(f)df. \label{ito}
\end{equation}
And since the expectation of $\int UdR$ is zero, only half of samples affects to the expectation of first equation. So by halving \cref{ito}, we get the result. 
\end{proof}

To compute the covariance value of absolute of random variables, we need following lemmas from \cite{Heydenreich:09}, \cite{Vleck:43}.

\noindent
{\bf Lemma 6} { For two Gaussian random variables $X,Y$ with means zero and correlation $\rho$, the expectation of $|X||Y|$ is formulated as follows:
\begin{equation*}
\begin{split}
    &\mathbb{E}[|X||Y|]=\frac{2}{\pi}(\sqrt{1-\rho^{2}}+\rho\arcsin{\rho}), \\
    &\mathbb{E}[1_{\{X>0
    \}}1_{\{Y>0
    \}}]=\frac{1}{4}+\frac{1}{2\pi}\arcsin{\rho}.
\end{split}
\end{equation*}
}

\noindent
{\bf Lemma 7} { Suppose mean zero complex-valued Gaussian random fields $\{R(f)\}_{f\in \mathbb{R}}, \{U(f)\}_{f\in \mathbb{R}}$ have following two-point correlators:
\begin{equation*}
    \begin{split}
        \mathbb{E}[R(f)\overline{R}(f')]=\delta(f-f')C(f), \\
        \mathbb{E}[U(f)]=m(f), \\
        \mathbb{E}[U(f)\overline{U}(f')]=K(f,f').
    \end{split}
\end{equation*}
let us define random fields $I(x),H(x)$ as follows:
\begin{equation*}
\begin{split}
    I(x)&:=\text{ReLU}(\int R(f)U(f)e^{ifx}df), \\
    H(x)&:=1_{\{\int R(f)U(f)e^{ifx}df>0\}}.
\end{split}
\end{equation*}
then the mean and two-point correlators are calculated as follows:
\begin{equation*}
    \begin{split}
        \mathbb{E}[I(x)]&=\frac{1}{\sqrt{2\pi}}V^{\frac{1}{2}} \quad \text{for all }x\in\mathbb{R}, \\
        \mathbb{E}[H(x)]&=\frac{1}{2} \quad \text{for all }x\in\mathbb{R}, \\
        \mathbb{E}[I(x)\overline{I}(x')]&=\frac{V}{4\pi}\Big(2\sqrt{1-\rho^{2}}+\rho(\pi+2\arcsin{\rho})\Big), \\
        \mathbb{E}[H(x)\overline{H}(x')]&=\frac{1}{4}+\frac{1}{2\pi}\arcsin{\rho}, \\
        \mathbb{E}[\overline{R}(f')\overline{U}(f')\mathcal{F}(I(x))]&=\frac{1}{2}\delta(f-f')K(f,f')C(f)
    \end{split}
\end{equation*}
where 
\begin{equation*}
    \begin{split}
        V&=\int C(f)K(f,f)df,  \\
        V\rho(x,x')&=\int C(f)K(f,f)\cos{(f(x-x'))}df.
    \end{split}
\end{equation*}
}

\begin{proof}
Firstly, since for fixed $x\in \mathbb{R}$, $\int R(f)U(f)e^{ifx}df$ is a mean-zero Gaussian distribution with variance $V:=\int C(f)K(f,f)df$. The mean of $I(x)$ which can be seen as a mean over Truncated Gaussian distribution is as follows:
\begin{equation*}
    \mathbb{E}[I(x)]=\sqrt{\frac{V}{2\pi}}.
\end{equation*}
and we can decompose ReLU activation into sum of $\frac{x}{2}$ and $\frac{|x|}{2}$. So, we get the following:
\begingroup
\small
\begin{equation*}
    \begin{split}
        &\mathbb{E}[I(x)\overline{I}(x')]\\
        &=\mathbb{E}\bigg[\frac{\Big(\int R(f)U(f)e^{ifx}df+\Big|\int R(f)U(f)e^{ifx}df\Big|\Big)\Big(\int \overline{R}(f)\overline{U}(f)e^{-ifx'}df+\Big|\int \overline{R}(f)\overline{U}(f)e^{-ifx'}df\Big|\Big)}{4} \bigg] \\
        &=\frac{1}{4}\bigg(\mathcal{F}^{-1}_{x}\mathcal{F}^{-1}_{x'}\Big(\mathbb{E}[R(f)\overline{R}(f')]\mathbb{E}[U(f)U(f')])+\mathbb{E}[R(f)\overline{R}(f')U(f)\overline{U}(f')1_{\int R(f)U(f)e^{ifx}df>0}] \\
        &+\mathbb{E}[R(f)\overline{R}(f')U(f)\overline{U}(f')1_{\int R(f)U(f)e^{ifx'}df>0}]\Big)+\mathbb{E}[\Big|\int R(f)U(f)e^{ifx}df\Big|\Big|\int \overline{R}(f)\overline{U}(f)e^{-ifx'}df\Big|]\bigg) \\
        &=\frac{1}{4}\bigg(\mathcal{F}^{-1}_{x}\mathcal{F}^{-1}_{x'}\Big(\delta(f-f')C(f)K(f,f')\Big)+V\frac{2}{\pi}\Big(\sqrt{1-\rho^{2}}+\rho\arcsin{\rho}\Big)\bigg) \\
        &=\frac{V}{4\pi}\Big(2\sqrt{1-\rho^{2}}+\rho(\pi+2\arcsin{\rho})\Big).
    \end{split}
\end{equation*}
\endgroup
It is clear that $\mathbb{E}[H(x)]=\frac{1}{2}$ and the result for $ \mathbb{E}[H(x)\overline{H}(x')]$ follows directly from the second equation of Lemma 6. Finally for the last equation, we have the following equation:
\begin{equation*}
\begin{split}
	   &\mathbb{E}[\overline{R}(f')\overline{U}(f')\mathcal{F}(I(x))] \\
	   &=\mathcal{F}(\mathbb{E}[\overline{R}(f')\overline(U)(f')I(x)]).
\end{split}
\end{equation*}
since $\int R(f)U(f)e^{ifx}df$ is mean-zero the second line of following equations holds, 
\begin{equation*}
\begin{split}
	&\mathcal{F}(\mathbb{E}[\overline{R}(f')\overline{U}(f')I(x)] )=\mathcal{F}( \mathbb{E}[\int \overline{R}(f')R(f)\overline{U}(f')U(f) e^{ifx}1_{I(x)>0}]) \\
	&=\mathcal{F}\frac{1}{2}\mathbb{E}[\mathcal{F}^{-1}(\overline{R}(f')R(f)\overline{U}(f')U(f))] \\
	&=\frac{1}{2}\mathcal{F}\mathcal{F}^{-1}\mathbb{E}[\overline{R}(f')R(f)\overline{U}(f')U(f))] \\
	&=\frac{1}{2}\delta(f-f')K(f,f')C(f).
\end{split}
\end{equation*}

\end{proof}

Now, consider the Fourier transform $\mathcal{F}(\mathfrak{S}^{(l)}\{\mathbf{u}\}) $ in the case of scale-invariant activation. We have 

\begin{equation*}
    \mathcal{F}(\mathfrak{S}^{(l)}\{\mathbf{u}\}) =(\alpha-\beta)\int \text{ReLU}(\int R^{(l)}_{ij}\hat{u}_{j}e^{ifx}df)e^{-ifx}dx + \beta (R^{(l)}_{ij}\hat{u}_{j}).
\end{equation*}
then

\begin{equation}
\begin{split}
    &\mathbb{E}\Big[\mathcal{F}(\mathfrak{S}^{(l)}\{\mathbf{u}\}) \overline{\mathcal{F}(\mathfrak{S}^{(l)}\{\mathbf{u}\}) }\Big](f,f')\\
&\Rightarrow \mathbb{E}\Big[(\alpha-\beta)^{2}\int \text{ReLU}(\int R^{(l)}_{ij}\hat{u}_{j}e^{ifx}df)e^{-ifx}dx \int \overline{\text{ReLU}(\int R^{(l)}_{ij}\hat{u}_{j}e^{ifx}df)}e^{if'x}dx \\
&+(\alpha-\beta)\beta \overline{(R^{(l)}_{ij}\hat{u}_{j})}\int \text{ReLU}(\int R^{(l)}_{ij}\hat{u}_{j}e^{ifx}df)e^{-ifx}dx\\
&+(\alpha-\beta)\beta (R^{(l)}_{ij}\hat{u}_{j})\int\overline{\text{ReLU}(\int R^{(l)}_{ij}\hat{u}_{j}e^{ifx}df)}e^{-if'x'}dx'+\beta^{2}(R^{(l)}_{ij}\hat{u}_{j})\overline{(R^{(l)}_{ij}\hat{u}_{j})}\Big]\\
&=(\alpha-\beta)^{2}\int\int \Big(\frac{V}{4\pi}(2\sqrt{1-\rho^{2}}+\rho(\pi+2\arcsin{\rho})\Big)e^{-ifx+if'x'}dxdx'\\
&(\alpha-\beta)\beta \delta(f-f')C(f)\mathcal{K}(f,f')+\beta^{2}\delta(f-f')C(f)\mathcal{K}(f,f').
\end{split} \label{relurecur}
\end{equation}

And from the fact that $\int (f*g)(x)dx = (\int f(x)dx)(\int g(x)dx)$ we can calculate the total integration of kernel over the two-point frequency domain by transforming the convoluted terms into geometric terms:
\begingroup
\small
\begin{equation*}
\begin{split}
&\int\int\mathbb{E}\Big[\mathcal{F}(\mathfrak{S}^{(l)}\{\mathbf{u}\}) \overline{\mathcal{F}(\mathfrak{S}^{(l)}\{\mathbf{u}\}) }\Big](f,f')dfdf'\\
    &\simeq \mathbb{E}\Big[(\alpha-\beta)^{2}\sigma_{n}\Big(\int\hat{u}_{j}R^{(l)}_{ij}(df)\Big)\overline{\sigma_{n}\Big(\int\hat{u}_{j}R^{(l)}_{ij}(df')\Big)} +(\alpha-\beta)\beta\overline{\int\hat{u}_{j}R^{(l)}_{ij}(df')}\sigma_{n}\Big(\int\hat{u}_{j}R^{(l)}_{ij}(df)\Big) \\
    &+(\alpha-\beta)\beta \int\hat{u}_{j}R^{(l)}_{ij}(df)\overline{\sigma_{n}\Big(\int\hat{u}_{j}R^{(l)}_{ij}(df')\Big)}+\beta^{2}\int R^{(l)}_{ij}(f)\overline{R^{(l)}_{ij'}}(f')\hat{u}_{j}(f)\hat{u}_{j'}(f')dfdf'\Big]\\
    & \Rightarrow \int\langle\mathcal{F}\mathfrak{S}^{(l)}\{\mathbf{u_{0}}\} ,\mathcal{F}\mathfrak{S}^{(l)}\{\mathbf{u_{0}}\} \rangle_{\mathcal{K}^{(l)}\{u_{0},u_{0}\}} (f,f')dfdf' \\
    &=\frac{\alpha^{2}+\beta^{2}}{2}\int \mathcal{H}^{(l)}(f)df.
\end{split}
\end{equation*}
\endgroup
From \cref{relurecur}, if $\alpha\neq\beta$, the first term survives which mixes all the components in positional domain. Then this term makes the kernel not to vanish over the truncated frequencies. And now we consider the parallel susceptibility. For that we first consider the following term:

\begin{equation*}
    \begin{split}
        &\langle \mathcal{F}(\mathbf{\mathbf{\mathcal{Z}}_{0}}\mathfrak{S}'^{(l)}|_{\mathbf{\mathcal{Z}}_{0}} ) ,\mathcal{F}(\mathfrak{S}^{(l)}|_{\mathbf{\mathcal{Z}}_{0}}) \{\mathbf{u_{0}}\}\rangle_{\mathcal{K}^{(l)}\{u_{0}\}}(f,f') \\
        &=\mathcal{F}_{x}\mathcal{F}_{x'}\Big(\langle\mathbf{\mathcal{Z}}_{0}\mathfrak{S}'^{(l)}|_{\mathbf{\mathcal{Z}}_{0}} ,\mathfrak{S}^{(l)}|_{\mathbf{\mathcal{Z}}_{0}}\rangle_{\mathcal{K}^{(l)}\{u_{0}\}}\Big)(f,f') \\
        &=\mathcal{F}_{x}\mathcal{F}_{x'}\Big(\langle\mathbf{\mathcal{Z}}_{0}H(\mathbf{\mathcal{Z}}_{0})(x),\mathfrak{S}^{(l)}|_{\mathbf{\mathcal{Z}}_{0}}(x')\rangle\Big)(f,f') \\
        &=\mathcal{F}_{x}\mathcal{F}_{x'}\Big(\langle\mathfrak{S}^{(l)}|_{\mathbf{\mathcal{Z}}_{0}}(x),\mathfrak{S}^{(l)}|_{\mathbf{\mathcal{Z}}_{0}}(x')\rangle\Big)(f,f').
    \end{split}
\end{equation*}
So, for parallel susceptibility we get the following simple result:
\begin{equation*}
    \begin{split}
        &\chi_{\parallel}(f,f')\\&=\frac{C_{R^{(l+1)}}\Big(\langle \mathcal{F}(\mathbf{\mathbf{\mathcal{Z}}_{0}}\mathfrak{S}'^{(l)}|_{\mathbf{\mathcal{Z}}_{0}} ) ,\mathcal{F}(\mathfrak{S}^{(l)}|_{\mathbf{\mathcal{Z}}_{0}}) \{\mathbf{u_{0}}\}\rangle_{\mathcal{K}^{(l)}\{u_{0}\}}+\langle\mathcal{F}(\mathfrak{S}^{(l)}|_{\mathbf{\mathcal{Z}}_{0}}) ,\mathcal{F}(\mathbf{\mathbf{\mathcal{Z}}_{0}}\mathfrak{S}'^{(l)}|_{\mathbf{\mathcal{Z}}_{0}} )\rangle_{\mathcal{K}^{(l)}\{u_{0}\}}\Big)}{2\mathcal{K}^{(l)}\{\mathbf{u_{0}}\}} \\
        &=\frac{\mathcal{K}^{(l+1)}(f,f')}{\mathcal{K}^{(l)}(f,f')}.
    \end{split}
\end{equation*}
And for the reduced perpendicular susceptibility:
\begin{equation*}
    \begin{split}
        &\tilde{\chi}_{\perp}(f,f')=\langle\mathcal{F}(\mathfrak{S}'^{(l)}\{\mathbf{u_{0}}\} ),\mathcal{F}(\mathfrak{S}'^{(l)}\{\mathbf{u_{0}}\} )\rangle_{\mathcal{K}^{(l)}\{u_{0},u_{0}\}}\\
        &=\mathcal{F}_{x}\mathcal{F}_{x'}\Big(\Big\|\Big((\alpha-\beta)H(x)+\beta\Big) \Big\|_{\mathcal{K}^{(l)}\{u_{0},u_{0}\}}\Big)\\
        &=\mathcal{F}_{x}\mathcal{F}_{x'}\Big(\frac{(\alpha-\beta)^{2}}{4}\Big(1+\frac{2}{\pi}\arcsin{\rho}\Big)+\alpha\beta\Big).
    \end{split}
\end{equation*}

\section{Derivation of Theorem 3} \label{app:thm3}
The recursive kernel for residual connection architecture is modified as follows:
\begin{equation*}
    \begin{split}
        \mathcal{K}^{(l+1)}\{\mathbf{u},\mathbf{v}\}(f):=\langle\mathcal{F}(\mathcal{R}^{(l+1)}( \mathfrak{S}^{(l)}\{\mathbf{u}\})+\tilde{\gamma}\mathcal{Z}^{(l)}\{\mathbf{u}\}),\mathcal{F}(\mathcal{R}^{(l+1)}(\mathfrak{S}^{(l)}\{\mathbf{v}\})+\tilde{\gamma}\mathcal{Z}^{(l)}\{\mathbf{v}\})\rangle_{\mathcal{K}^{(l)}}
    \end{split}
\end{equation*}
Then, for single input data, we get the following recursive formula:
\begin{equation*}
    \begin{split}
        &\mathcal{K}^{(l+1)}\{\mathbf{u}\}(f):=C_{R^{(l+1)}}\| \mathcal{F}(\mathfrak{S}^{(l)}\{\mathbf{u}\})\|_{\mathcal{K}^{(l)}}+\tilde{\gamma}\langle \mathcal{F}(\mathcal{Z}^{(l)}\{\mathbf{u}\}),\mathcal{F}(\mathcal{R}^{(l+1)}(\mathfrak{S}^{(l)}\{\mathbf{u}\}))\rangle_{\mathcal{K}^{(l)}}\\
        &+\tilde{\gamma}\langle \mathcal{F}(\mathcal{R}^{(l+1)}(\mathfrak{S}^{(l)}\{\mathbf{u}\})),\mathcal{F}(\mathcal{Z}^{(l)}\{\mathbf{u}\})\rangle_{\mathcal{K}^{(l)}}+\tilde{\gamma}^{2}\|\mathcal{F}(\mathcal{Z}^{(l)}\{\mathbf{u}\})\|_{\mathcal{K}^{(l)}}
    \end{split}
\end{equation*}
then for analytic activation case, we have the following kernel recursive formula:

\begingroup
\small
\begin{equation*}
    \begin{split}
        &C_{R^{(l+1)}}\| \mathcal{F}(\mathfrak{S}^{(l)}\{\mathbf{u}\})\|_{\mathcal{K}^{(l)}}+\tilde{\gamma}\langle \mathcal{F}(\mathcal{Z}^{(l)}\{\mathbf{u}\}),\mathcal{F}(\mathcal{R}^{(l+1)}(\mathfrak{S}^{(l)}\{\mathbf{u}\}))\rangle_{\mathcal{K}^{(l)}}\\
        &+\tilde{\gamma}\langle \mathcal{F}(\mathcal{R}^{(l+1)}(\mathfrak{S}^{(l)}\{\mathbf{u}\})),\mathcal{F}(\mathcal{Z}^{(l)}\{\mathbf{u}\})\rangle_{\mathcal{K}^{(l)}}+\tilde{\gamma}^{2}\|\mathcal{F}(\mathcal{Z}^{(l)}\{\mathbf{u}\})\|_{\mathcal{K}^{(l)}}\\
        &=\delta(f-f')\Big(\gamma\mathcal{K}^{(l)}\{\mathbf{u}\}(f)\\&+C_{R^{(l+1)}}(f)\sum_{k=1}^{\infty} \frac{\sigma_{k}^{2}}{(n^{(l)})^{k-1}}\sum_{n_{k,1}+\dots+n_{k,l}=k}\frac{k(n_{k,1})!\dots(n_{k,l})!}{(k-1)!}(\mathcal{H}^{(l)})^{*n_{k,1}}\cdots(\mathcal{H}^{(l)})^{*n_{k,l}}(f)\Big).
    \end{split}
\end{equation*}
\endgroup

\begin{equation*}
    \begin{split}
        &\Rightarrow \mathcal{K}^{(l+1)}\{\mathbf{u}\}(f,f')=\delta(f-f')\Big((\sigma_{1}^{2}C_{R^{(l+1)}}(f)+\gamma)\mathcal{K}^{(l)}\{\mathbf{u}\}(f)\\&+\sum_{k=2}^{\infty}\frac{\sigma_{k}^{2}}{(n^{(l)})^{k-1}}\sum_{n_{k,1}+\dots+n_{k,l}=k}\frac{(n_{k,1})!\dots(n_{k,l})!}{k!}(\mathcal{H}^{(l)})^{*n_{k,1}}\cdots(\mathcal{H}^{(l)})^{*n_{k,l}}(f)\Big).
    \end{split}
\end{equation*}

where we set $\gamma=\tilde{\gamma}^{2}$. For the susceptibilities, we proceed analogously; the parallel case is omitted for brevity:
\begin{equation*}
    \begin{split}
        &\langle \mathcal{F}(\mathfrak{S}'^{(l)}\{\mathbf{u_{0}}\}+\tilde{\gamma}\mathbf{1}),\mathcal{F}(\mathfrak{S}'^{(l)}\{\mathbf{u_{0}}\}+\tilde{\gamma}\mathbf{1})\rangle_{\mathcal{K}^{(l)}\{u_{0},u_{0}\}} \\
        &=\langle \mathcal{F}(\mathfrak{S}'^{(l)}\{\mathbf{u_{0}}\}),\mathcal{F}(\mathfrak{S}'^{(l)}\{\mathbf{u_{0}}\})\rangle_{\mathcal{K}^{(l)}\{u_{0},u_{0}\}} + 2\langle \mathcal{F}(\mathfrak{S}'^{(l)}\{\mathbf{u_{0}}\}),\tilde{\gamma}\mathbf{1}\rangle_{\mathcal{K}^{(l)}\{u_{0},u_{0}\}}+\gamma\delta(f)\delta(f')\\
        &=\delta(f-f')\Big(\sum_{k=1}^{\infty}\frac{\sigma_{k}^{2}}{(n^{(l)})^{k-1}}\sum_{n_{k,1}+\dots+n_{k,l}=k-1}\frac{(n_{k,1})!\dots(n_{k,l})!}{(k-1)!}(\mathcal{H}^{(l)})^{*n_{k,1}}\cdots(\mathcal{H}^{(l)})^{*n_{k,l}}(f)\Big)\\
        &+2\tilde{\gamma}\sigma_{1}\delta(f)\delta(f')+\gamma\delta(f)\delta(f').
    \end{split}
\end{equation*}
Here $\mathbf{1}=(1,\dots,1)^{T}\in \mathbb{R}^{n^{(l)}}$.
\section{On the Compact Periodic Domain} \label{app:compact}
As noted above, when the domain is the entire real line, the formula in Lemma 2 yields a $\delta(0)$ term—i.e., a divergent value. In practice, however, we implement the FNO on a finite periodic domain, and in this setting the $\delta(0)$ in equations in previous sections equals to $\frac{1}{L}$ where $L$ is size of the domain, so the divergence disappears. Consequently, the kernel evolution equations and criticality conditions in Sections 4.1, 4.2, and 4.3 can be rewritten as follows:

\noindent
{\bf Corollary 1} { With the analytic, origin passing activation specified in \cref{analy} and a Fourier Neural Operator defined by \cref{defno} under the initialization ensemble \cref{initen}, the kernel and the susceptibilities are given by the following recursion relations:
\begingroup
\small
\begin{equation}
\begin{split}
&\mathcal{K}^{(l+1)}(m,n)=\frac{\delta_{m,n}C_{R^{(l+1)}}(n)}{L}\sum_{k=1}^{\infty}\frac{\sigma_{k}^{2}}{(n^{(l)})^{k-1}}\sum_{n_{k,1}+\dots+n_{k,l}=k}\frac{(n_{k,1})!\dots(n_{k,l})!}{k!}(\mathcal{H}^{(l)})^{*n_{k,1}}\cdots(\mathcal{H}^{(l)})^{*n_{k,l}}(n) 
\end{split} \label{kerre}
\end{equation}
\begin{equation}
\begin{split}
        &\chi_{\parallel}(m,n)=\frac{\delta_{m,n}}{L}C_{R^{(l+1)}}(n)\sum_{k=1}^{\infty} \frac{\sigma_{k}^{2}}{(n^{(l)})^{k-1}}\sum_{n_{k,1}+\dots+n_{k,l}=k}\frac{(n_{k,1})!\dots(n_{k,l})!}{(k-1)!}\frac{(\mathcal{H}^{(l)})^{*n_{k,1}}\cdots(\mathcal{H}^{(l)})^{*n_{k,l}}(n)}{\mathcal{K}^{(l)}\{\mathbf{u_0}\}(n)}, 
        \end{split}\label{parare2}
\end{equation} 
\begin{equation}
\begin{split}
    &\tilde{\chi}_{\perp}(m,n)=\delta_{m,n}\sum_{k=1}^{\infty}\frac{\sigma_{k}^{2}}{(n^{(l)})^{k-1}}\sum_{n_{k,1}+\dots+n_{k,l}=k-1}\frac{(n_{k,1})!\dots(n_{k,l})!}{(k-1)!}(\mathcal{H}^{(l)})^{*n_{k,1}}\cdots(\mathcal{H}^{(l)})^{*n_{k,l}}(n).
\end{split} \label{perpre}
\end{equation}
\endgroup
} 
\noindent
{\bf Corollary 2} { With the scale-invariant activation specified in \cref{scaleac} and a Fourier Neural Operator defined by \cref{defno} under the initialization ensemble \cref{initen}, the kernel and the susceptibilities are given by the following recursion relations:
\begingroup
\small
\begin{equation}
\begin{split}
\mathcal{K}^{(l+1)}(m,n)&=(\alpha-\beta)^{2}\frac{\delta_{m,n}C_{R^{(l+1)}}(n)}{L} \\
&\int \frac{V^{(l)}}{4\pi}\Big(2\sqrt{1-(\rho^{(l)})^{2}}+\rho^{(l)}(\pi+2\arcsin{\rho^{(l)}})\Big)e^{-imx+inx'}dxdx' \\&+\alpha\beta  \frac{\delta_{m,n}C_{R^{(l+1)}}(n)}{L}\mathcal{H}^{(l)}(n),
\end{split} \label{kerrelu}
\end{equation}
\begin{equation}
\begin{split}
\sum_{n}\mathcal{K}&^{(l+1)}(n)=\frac{\alpha^{2}+\beta^{2}}{2}\sum_{n} \mathcal{H}^{(l)}(n) 
\end{split}
\end{equation}
\begin{equation}
\begin{split}
        \chi_{\parallel}(m,n)&=\frac{\mathcal{K}^{(l+1)}(m,n)}{\mathcal{K}^{(l)}(m,n)},
        \end{split} \label{pararelu}
\end{equation}
\begin{equation}
\begin{split}
    \tilde{\chi}_{\perp}(m,n)&=\int\Big(\frac{(\alpha-\beta)^{2}}{4}\Big(1+\frac{2}{\pi}\arcsin{\rho^{(l)}}\Big)+\alpha\beta\Big)e^{-imx+inx'}dxdx'. \label{perprelu}
\end{split}
\end{equation}
 where $V^{(l)}=\sum_{n}{\mathcal{K}}^{(l)}(n)$, $V^{(l)}\rho^{(l)}(x,x')=\sum_{n} \mathcal{H}^{(l)}(n)\cos(n(x-x'))$.
\endgroup

} 

\noindent
{\bf Corollary 3} { With the analytic, origin passing activation specified in \cref{analy} and a Fourier Neural Operator defined based on \cref{defno} and modification of Fourier layers by \cref{resfno} under the initialization ensemble \cref{initen}, the kernel and the susceptibilities are given by the following recursion relations:
\begingroup
\small
\begin{equation}
\begin{split}
&\mathcal{K}^{(l+1)}\{\mathbf{u}\}(m,n)=\frac{\delta_{m,n}}{L}\Big((\sigma_{1}^{2}C_{R^{(l+1)}}(n)+\gamma)\mathcal{K}^{(l)}\{\mathbf{u}\}(n)\\&+C_{R^{(l+1)}}(n)\sum_{k=2}^{\infty}\frac{\sigma_{k}^{2}}{(n^{(l)})^{k-1}}\sum_{n_{k,1}+\dots+n_{k,l}=k}\frac{(n_{k,1})!\dots(n_{k,l})!}{k!}(\mathcal{H}^{(l)})^{*n_{k,1}}\cdots(\mathcal{H}^{(l)})^{*n_{k,l}}(n)\Big). 
\end{split} \label{kerres}
\end{equation}
\begin{equation}
\begin{split}
        &\chi_{\parallel}(m,n)\\
        &=\frac{\delta_{m,n}}{L}\Big(\gamma+C_{R^{(l+1)}}(n)\\&\sum_{k=1}^{\infty} \frac{\sigma_{k}^{2}}{(n^{(l)})^{k-1}}\sum_{n_{k,1}+\dots+n_{k,l}=k}\frac{k(n_{k,1})!\dots(n_{k,l})!}{(k-1)!}\frac{(\mathcal{H}^{(l)})^{*n_{k,1}}\cdots(\mathcal{H}^{(l)})^{*n_{k,l}}(n)}{\mathcal{K}^{(l)}\{\mathbf{u_0}\}(n)}\Big).
\end{split} \label{parares}
\end{equation}
\begin{equation}
\begin{split}
    &\tilde{\chi}_{\perp}(m,n)\\
    &=\delta_{m,n}\Big(\sum_{k=1}^{\infty}\frac{\sigma_{k}^{2}}{(n^{(l)})^{k-1}}\sum_{n_{k,1}+\dots+n_{k,l}=k-1}\frac{(n_{k,1})!\dots(n_{k,l})!}{(k-1)!}(\mathcal{H}^{(l)})^{*n_{k,1}}\cdots(\mathcal{H}^{(l)})^{*n_{k,l}}(n)\Big)\\&+\frac{\gamma+2\tilde{\gamma}\sigma_{1}}{L^{2}}\delta_{n,0}\delta_{m,0}.
\end{split} \label{perpres}
\end{equation}
\endgroup
}

\end{document}